\documentclass{article}

\usepackage{microtype}
\usepackage{graphicx}
\usepackage{subfigure}
\usepackage{booktabs} 

\usepackage{hyperref}

\usepackage[accepted]{icml2024}

\usepackage{amsmath}
\usepackage{amssymb}
\usepackage{mathtools}
\usepackage{amsthm}

\usepackage[capitalize,noabbrev]{cleveref}

\theoremstyle{plain}

\theoremstyle{definition}

\theoremstyle{remark}

\usepackage[textsize=tiny]{todonotes}

\usepackage{enumitem} 
\usepackage{pifont}   
\usepackage{multirow} 
\usepackage{makecell} 
\usepackage{comment}  

\icmltitlerunning{Learning Coverage Paths in Unknown Environments with Deep Reinforcement Learning}

\begin{document}

\twocolumn[
\icmltitle{Learning Coverage Paths in Unknown Environments \\ with Deep Reinforcement Learning}



\icmlsetsymbol{equal}{*}

\begin{icmlauthorlist}
\icmlauthor{Arvi Jonnarth}{liu,husqvarna}
\icmlauthor{Jie Zhao}{dalian}
\icmlauthor{Michael Felsberg}{liu,durban}
\end{icmlauthorlist}

\icmlaffiliation{liu}{Linköping University.}
\icmlaffiliation{husqvarna}{Husqvarna Group.}
\icmlaffiliation{dalian}{Dalian University of Technology.}
\icmlaffiliation{durban}{Co-affiliation: University of KwaZulu-Natal}

\icmlcorrespondingauthor{Arvi Jonnarth}{arvi.jonnarth@liu.se}

\icmlkeywords{Reinforcement Learning, Coverage Path Planning, Online, Robotics}

\vskip 0.3in
]



\printAffiliationsAndNotice{}  

\begin{abstract}
Coverage path planning (CPP) is the problem of finding a path that covers the entire free space of a confined area, with applications ranging from robotic lawn mowing to search-and-rescue. When the environment is unknown, the path needs to be planned online while mapping the environment, which cannot be addressed by offline planning methods that do not allow for a flexible path space. We investigate how suitable reinforcement learning is for this challenging problem, and analyze the involved components required to efficiently learn coverage paths, such as action space, input feature representation, neural network architecture, and reward function. We propose a computationally feasible egocentric map representation based on frontiers, and a novel reward term based on total variation to promote complete coverage. Through extensive experiments, we show that our approach surpasses the performance of both previous RL-based approaches and highly specialized methods across multiple CPP variations.
\end{abstract}

\begin{figure*}[!t]
    \vskip 0.05in
    \centering
    \setlength{\tabcolsep}{5pt}
    \begin{tabular}{cc}
        \includegraphics[height=.3\linewidth]{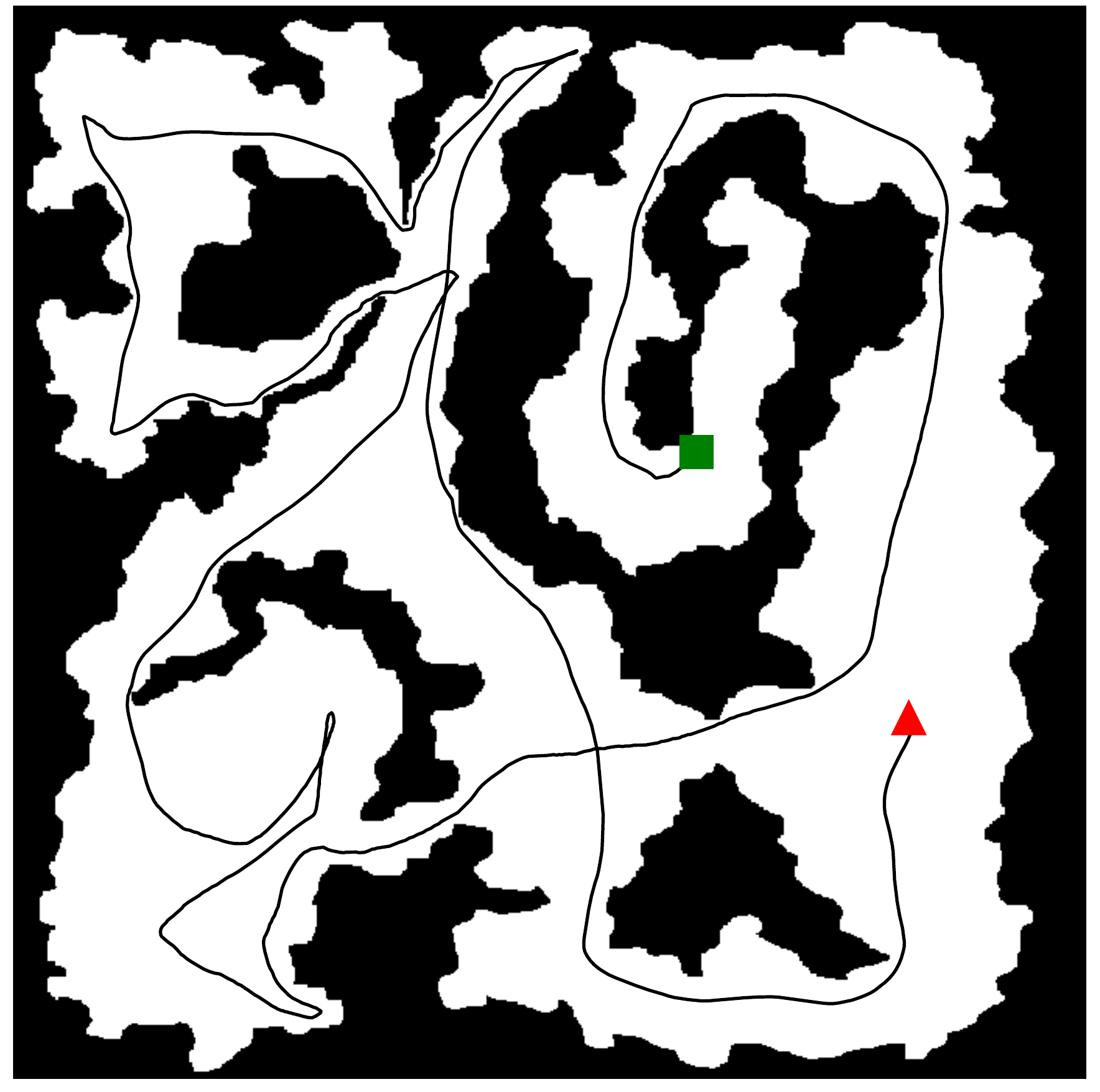} &
        \includegraphics[height=.3\linewidth]{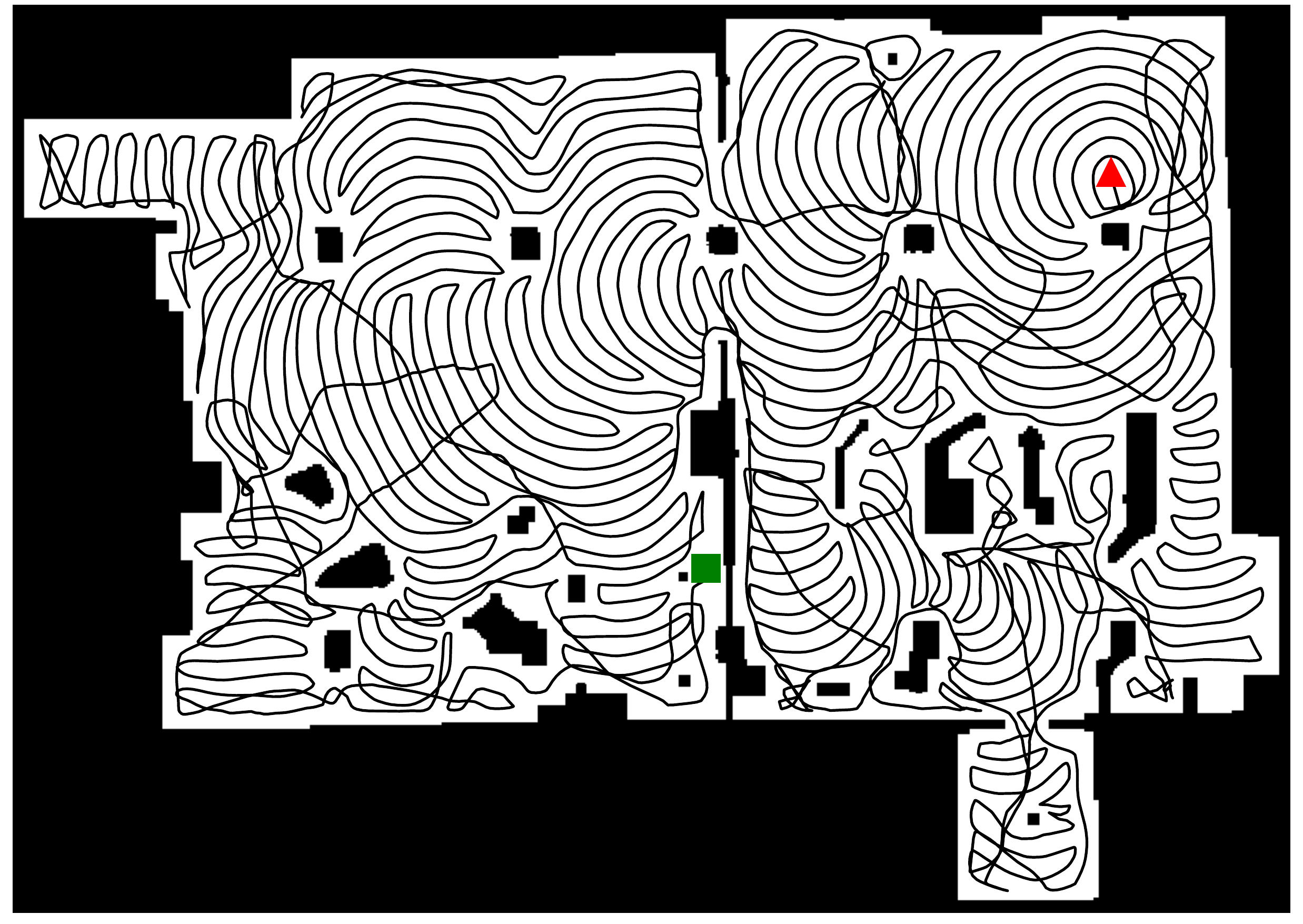} \\
    \end{tabular}
    \vspace{-8pt}
    \caption{Learned paths for exploration (left) and lawn mowing (right), including the start (red triangle) and end position (green square).}
    \label{fig_qualitative_paths}
\end{figure*}

\section{Introduction}

Task automation is an ever growing field, and as automated systems become more intelligent, they are expected to solve increasingly challenging problems. One such problem is coverage path planning (CPP), where a path is sought that covers the entire free space of a confined area. It is related to the traveling salesman and covering salesman problems \citep{galceran2013ras}, and known to be NP-hard \citep{arkin2000cg}. If the environment is known \textit{a priori}, including the extent of the area and the location and geometry of all obstacles, an optimal path can be planned \textit{offline} \citep{huang2001optimal}. However, if the environment is initially unknown, or at least partially unknown, the path has to be planned \textit{online}, \textit{e.g.}\ by a robot, using sensors to simultaneously map the environment. In this case, an optimal path cannot generally be found \citep{galceran2013ras}.

Coverage path planning is an essential part in a wide variety of robotic applications, where the range of settings define different variants of the CPP problem. In applications such as robotic lawn mowing \citep{cao1988jrs}, snow removal \citep{choset2001amai}, and vacuum cleaning \citep{yasutomi1988icra}, the region that is covered is defined by an attached work tool. In contrast, in applications such as search-and-rescue \citep{jia2016ssrr}, autonomous underwater exploration \citep{hert1996ar}, and aerial terrain coverage \citep{xu2011icra}, the covered region is defined by the range of a sensor. Additionally, the space of valid paths is limited, induced by the motion constraints of the robot.

We aim to develop a task- and robot-agnostic method for online CPP in unknown environments. This is a challenging task, as intricacies in certain applications may be difficult to model, or even impossible. In the real world, unforeseen discrepancies between the modeled and true environment may occur, \textit{e.g.}\ due to wheel slip.
A further aspect that needs to be considered is the sensing capability of the agent, \textit{e.g.}\ if it is not omnidirectional, the agent needs to rotate itself to explore a different view. Incompleteness of the present model must be assumed to account for open world scenarios. Thus, we approach the problem from the perspective of learning through embodiment. Reinforcement learning (RL) lends itself as a natural choice, where an agent can interact with the world and adapt to specific environment conditions, without the need for an explicit model. Concretely, we consider the case where an RL model directly predicts the control signals for an agent from sensor data. Besides the benefit of adapting to specific environment characteristics, this approach does not constrain the path space in terms of pre-defined patterns in cellular decomposition methods \citep{choset1998coverage}, or primitive motions in grid-based methods \citep{gabriely2002ras}. In other words, it allows for flexibility in the path space, as shown in \cref{fig_qualitative_paths}.

We analyze the different \textit{components} in RL to efficiently learn coverage paths. To enable learning every \textit{valid} path, a \textit{discrete action space} based on a grid world is insufficient, as it excludes a large set of paths. A \textit{continuous action space} is necessary, where the model predicts control signals for an agent. To decide which action to perform, the agent needs information about its pose and the currently known state of the environment. Following \citet{chen2018learning}, we use egocentric maps. This provides a suitable \textit{observation space}, as the egocentric maps efficiently encode the pose in an easily digestible manner. For a scalable input representation for the RL model, we propose to use multiple maps in different scales, with lower resolutions for the larger scales. With this approach, a square of size $n \times n$ can be represented in $\mathcal{O}(\log n)$ memory complexity. We make this multi-scale approach viable through frontier maps that preserve information about the existence of non-covered space at any scale. A convolutional \textit{network architecture} is a natural choice as it efficiently exploits the spatial correlation in the maps. We propose to group the maps by scale and convolve them separately, as their spatial positions do not correspond to the same location. Besides intuitive positive \textit{reward terms} based on covering new ground \citep{chen2018learning,chaplot2020Learning}, we propose a novel reward term based on total variation, which guides the agent not to leave small holes of non-covered free space. Our code implementation can be found online\footnote{Code: \url{https://github.com/arvijj/rl-cpp}}, and our contributions are summarized as follows:
\begin{itemize}[leftmargin=*,topsep=-4pt]
    \setlength\itemsep{-2pt}
    \item We propose an end-to-end deep reinforcement learning approach for the continuous online CPP problem.
    \item We make multi-scale maps viable through frontier maps, which can represent non-visited space in any scale.
    \item We propose a network architecture that exploits the structure of the multi-scale maps.
    \item We introduce a novel total variation reward term for reducing holes of non-covered space, improving convergence.
    \item Finally, we conduct extensive experiments and ablations to validate the effectiveness of our approach.
\end{itemize}

\section{Related Work}

Coverage path planning has previously been approached from the perspective of classical robotics, as well as reinforcement learning. Both approaches require a suitable map representation of the environment. Below, we summarize the related works within these topics.

\textbf{Planning-based methods.} \textit{Cellular decomposition methods}, such as boustrophedon cellular decomposition (BCD) \citep{choset1998coverage} and Morse decomposition \citep{acar2002morse}, subdivide the area into non-overlapping cells free of obstacles. Each cell is covered with a pre-defined pattern, \textit{e.g.}\ a back-and-forth motion. The cells are then connected using transport paths, which inevitably lead to overlap with the covered cells. \textit{Grid-based methods}, such as Spiral-STC \citep{gabriely2002ras} and the backtracking spiral algorithm (BSA) \citep{gonzalez2005icra}, decompose the environment into uniform grid cells, such as rectangles \citep{zelinsky1993planning} or triangles \citep{oh2004toie}, with a comparable size to the agent. The coverage path is defined by primitive motions between adjacent cells. The simplicity of these methods is attractive, but they heavily constrain the space of available paths. \textit{Frontier-based methods} keep track of the boundary between covered and non-covered free space, \textit{i.e.\ the frontier}. Segments of the frontier are clustered into nodes, where one node is chosen as the next short-term goal. Different criteria for choosing which node to visit next have been explored, such as the distance to the agent \citep{yamauchi1997frontier}, the path in a rapidly exploring random tree (RRT) \citep{umari2017autonomous}, or the gradient in a potential field \citep{yu2021smmr}. These methods are mainly tailored for robot exploration, where the notion of a frontier is suitable. Compared to planning-based methods, RL can adapt to specific environment characteristics.

\textbf{RL-based methods.} Many learning-based methods combine classical components with reinforcement learning. For known environments, \citet{chen2019iros} present Adaptive Deep Path, where they apply RL to find the cell order in BCD. Discrete methods \citep{piardi2019aip,kyaw2020access} utilize the simplicity of grid-based approaches, where an RL model predicts movement primitives, e.g.\ to move up, down, left, or right, based on a discrete action space. However, this constrains the space of possible paths, and may result in paths that are not possible to navigate by a robot in the real world, which is continuous by nature. Thus, we consider a continuous action space, which has been used successfully for the related navigation problem \citep{tai2017iros}. We utilize soft actor-critic (SAC) RL \citep{haarnoja2018soft} due to its sample efficiency for continuous spaces. Other works combine RL with a frontier-based approach, either by using an RL model to predict the cost of each frontier node \citep{niroui2019ral}, or to predict the control signals for navigating to a chosen node \citep{hu2020tovt}. Different from previous approaches, which use RL within a multi-stage framework, we predict continuous control signals end-to-end to fully utilize the flexibility of RL. To the best of our knowledge, we are the first to do so for CPP.

\textbf{Map representation.} To perform coverage path planning, the environment needs to be represented in a suitable manner. Similar to previous work, we discretize the map into a 2D grid with sufficiently high resolution to accurately represent the environment. This map-based approach presents different choices for the input feature representation. \citet{saha2021efficient} observe such maps in full resolution, where the effort for a $d \times d$ grid is $\mathcal{O}(d^2)$, which is infeasible for large environments. \citet{niroui2019ral} resize the maps to keep the input size fixed. However, this means that the information in each grid cell varies between environments, and hinders learning and generalization for differently sized environments. Meanwhile, \citet{shrestha2019learned} propose to learn the map in unexplored regions. Instead of considering the whole environment at once, other works \citep{heydari2021arxiv,saha2021deep} observe a local neighborhood around the agent. While the computational cost is manageable, the long-term planning potential is limited as the space beyond the local neighborhood cannot be observed. For example, if the local neighborhood is fully covered, the agent must pick a direction at random to explore further. To avoid the aforementioned limitations, we use multiple maps in different scales, similar to \citet{klamt2018planning}.

\section{Learning Coverage Paths}

Before presenting our approach, we first define the online CPP problem in \cref{sec_problem_definition}, after which we formulate it as a Markov decision process in \cref{sec_reinforcement_learning}. Then, we present our RL-based approach in terms of observation space in \cref{sec_observation_space}, action space in \cref{sec_action_space}, reward function in \cref{sec_reward_function}, and agent architecture in \cref{sec_agent_architecture}.

\begin{figure*}[!t]
    \vskip 0.1in
    \centering
    \includegraphics[width=0.99\linewidth]{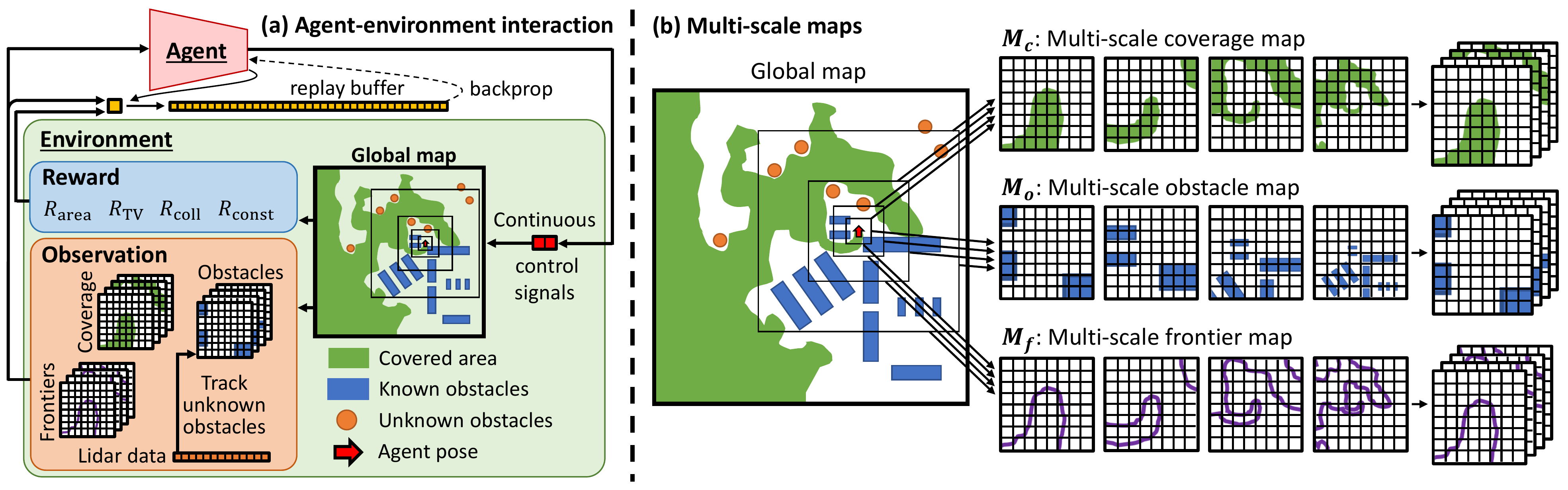}
    \vspace{-6pt}
    \caption{\textbf{(a) Agent-environment interaction:} The observation consists of multi-scale maps from (b) and lidar detections, based on which the model predicts continuous control signals for an agent. \textbf{(b) Illustration of coverage, obstacle, and frontier maps in multiple scales:} This example shows $m=4$ scales with a scale factor of $s=2$. All scales are centered at the agent, and discretized into the same pixel resolution, resulting in the multi-scale maps $M_c$, $M_o$, and $M_f$, of size $8 \times 8 \times 4$ here.}
    \label{fig_multi_scale_maps}
\end{figure*}

\subsection{Problem Definition and Delineations}
\label{sec_problem_definition}

Without prior knowledge about the geometry of a confined area, the goal is to navigate in it with an agent with radius $r$ to visit every point of the free space. The free space includes all points inside the area that are not occupied by obstacles. A point is considered visited when the distance between the point and the agent is less than the coverage radius $d$, and the point is within the field-of-view of the agent. This definition unifies variations where $d \leq r$, which we define as the \textit{lawn mowing problem}, and where $d > r$, which we refer to as \textit{exploration}. To interactively gain knowledge about the environment, and for mapping its geometry, the agent needs some form of sensor. Both ranging sensors \citep{hu2020tovt} and depth cameras \citep{chen2018learning} have been utilized for this purpose. Following \citet{xu2022explore}, we choose a simulated 2D light detection and ranging (lidar) sensor that can detect obstacles in fixed angles relative to the agent, although our proposed framework is not limited to this choice. Based on the pose of the agent, the detections are transformed to global coordinates, and a map of the environment is continuously formed. As the focus of this paper is to learn coverage paths, and not to solve the localization problem, we assume known pose up to a certain noise level. However, our method may be extended to the case with unknown pose with the use of an off-the-shelf SLAM method. Finally, while our method can be further extended to account for moving obstacles and multi-agent coverage, they are beyond the scope of this paper.

\subsection{Coverage Path Planning as a Markov Decision Process}
\label{sec_reinforcement_learning}

We formulate the CPP problem as a partially observable Markov decision process (POMDP) with discrete time steps $t \in [1 .. T]$, continuous actions $a_t \sim \pi (a_t | o_t)$ made by the agent according to policy $\pi$, observations and rewards $o_t, r_t \sim p(o_t, r_t | s_{t}, a_{t-1})$, and states $s_t \sim p(s_t | s_{t-1}, a_{t-1})$ provided by the environment. We use a neural network for the policy, and the goal for the agent is to maximize the expected discounted reward $\mathbb{E} ( \sum_{t=1}^T \gamma^t r_t )$ with discount factor $\gamma$. In our case, the state $s_t$ consists of the geometry of the area, location and shape of all obstacles, visited points, and the pose of the agent. The observation $o_t$ contains the area and obstacle geometries mapped until time step $t$, as well as visited points, agent pose, and sensor data. Based on the observation, the model predicts control signals for the agent. The reinforcement learning loop is depicted in Figure~\ref{fig_multi_scale_maps}(a), where observations, actions, and rewards are collected into a replay buffer, from which training batches are drawn for gradient backpropagation.

\subsection{Observation Space}
\label{sec_observation_space}

To efficiently learn coverage paths, the observation space needs to be mapped into a suitable input feature representation for the neural network policy. To this end, we represent the visited points as a \textit{coverage map}, and the mapped obstacles and boundary of the area as an \textit{obstacle map}. The maps are discretized into a 2D grid with sufficiently high resolution to accurately represent the environment. To represent large regions in a scalable manner, we propose to use multi-scale maps, which was necessary for large environments, see Appendix \ref{supp_sec_multi_scale_maps}. We make this viable through \textit{frontier maps} that preserve information about the existence of non-covered space, even in coarse scales.

\textbf{Multi-scale maps.} Inspired by multi-layered maps with varying scale and coarseness levels for search-and-rescue \citep{klamt2018planning}, we propose to use multi-scale maps for the coverage and obstacles to solve the scalability issue. We extract multiple local neighborhoods with increasing size and decreasing resolution, keeping the grid size fixed. We start with a local square crop $M^{(1)}$ with side length $d_1$ for the smallest and finest scale. The multi-scale map representation $M = \{M^{(i)}\}_{i=1}^m$ with $m$ scales is constructed by cropping increasingly larger areas based on a fixed scale factor $s$. Concretely, the side length of map $M^{(i)}$ is $d_{i} = s d_{i-1}$. The resolution for the finest scale is chosen sufficiently high such that the desired level of detail is attained in the nearest vicinity of the agent, allowing it to perform precise \textit{local navigation}. At the same time, the large-scale maps allow the agent to perform \textit{long-term planning}, where a high resolution is less important. This multi-scale map representation can completely contain an area of size $d \times d$ in $\mathcal{O}(\log d)$ number of scales. The total number of grid cells is $\mathcal{O}(wh \log d)$, where $w$ and $h$ are the fixed width and height of the grids, and do not depend on $d$. This is a significant improvement over a single fixed-resolution map with $\mathcal{O}(d^2)$ grid cells. For the observation space, we use a multi-scale map $M_c$ for the coverage and $M_o$ for the obstacles. These are illustrated in Figure \ref{fig_multi_scale_maps}(b).

\textbf{Frontier maps.} When the closest vicinity is covered, the agent needs to make a decision where to explore next. However, the information in the low-resolution large-scale maps may be insufficient. For example, consider an obstacle-cluttered region where the obstacle boundaries have been mapped. A low coverage could either mean that some parts are non-covered free space, or that they are occupied by obstacles. These cases cannot be distinguished if the resolution is too low. As a solution to this problem, we propose to encode a multi-scale frontier map $M_f$, which we define in the following way. In the finest scale, a grid cell that has not been visited is a frontier point if any of its neighbours have been visited. Thus, a frontier point is non-visited free space that is reachable from the covered area. A region where the entire free space has been visited does not induce any frontier points. In the coarser scales, a grid cell is a frontier cell if and only if it contains a frontier point. In this way, the existence of frontier points persists through scales. Thus, regions with non-covered free space can be deduced in any scale, based on this multi-scale frontier map representation.

\textbf{Egocentric maps.} As the movement is made relative to the agent, its pose needs to be related to the map of the environment. Following \citet{chen2018learning}, we use egocentric maps which encode the pose by representing the maps in the coordinate frame of the agent. Each multi-scale map is aligned such that the agent is in the center of the map, facing upwards. This allows the agent to easily map observations to movement actions, instead of having to learn an additional mapping from a separate feature representation of its position, such as a 2D one-hot map \citep{theile2020iros}.

\textbf{Sensor observations.} To react on short-term obstacle detections, we include the sensor data in the input feature representation. The depth measurements from the lidar sensor are normalized to $[0, 1]$ based on its maximum range, and concatenated into a vector $S$ (see Appendix \ref{supp_sec_normalizing_distance} for a discussion on implications of this choice).

\subsection{Action Space}
\label{sec_action_space}

We let the model directly predict the control signals for the agent. This allows it to adapt to specific environment characteristics, while avoiding a constrained path space, different from a discrete action space. We consider an agent that independently controls the linear velocity $v$ and the angular velocity $\omega$, although the action space may seamlessly be changed to specific vehicle models. To keep a fixed output range, the actions are normalized to $[-1, 1]$ based on the maximum linear and angular velocities, where the sign of the velocities controls the direction of travel and rotation.

A continuous action space is of course not the only choice, but it is more realistic than a discrete one. There are many reasons why we chose continuous actions. (1) This allows us to model a continuous pose that is not constrained by the grid discretization, and thus not constraining the path space. (2) The agent can adapt and optimize its path for a specific kinematic model. In Section \ref{sec_experiments} we find that our approach works well in a continuous setting, which hints that it may also work in other continuous control spaces for specific kinematic models. The same conclusion could not be drawn if a discrete action space was used. (3) Our specific choice of linear and angular velocities applies directly to a wide variety of robots, e.g. differential drive systems and the Ackermann kinematic model (by using a constraint on the angular velocity). Thus, sim-to-real transfer is more likely compared to a discrete action space. (4) Compared to grid-based discrete actions, a continuous action space introduces additional sources for error, both regarding dynamics and localization, which are more realistic.

\subsection{Reward Function}
\label{sec_reward_function}

\begin{figure}[t]
    \vskip 0.05in
    \centering
    \includegraphics[width=0.85\linewidth]{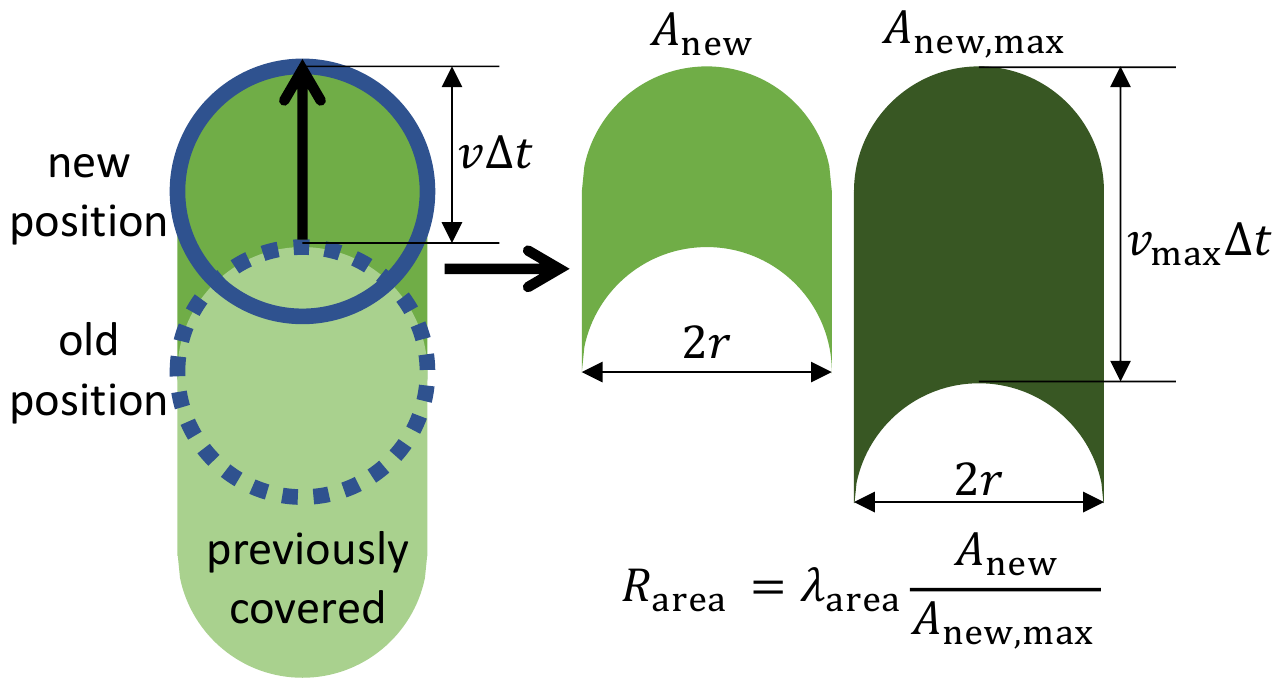}
    \vspace{-6pt}
    \caption{The area reward $R_\mathrm{area}$ is based on the maximum possible area that can be covered in each time step.}
    \label{fig_reward_area}
\end{figure}

As the goal is to cover the free space, a reward based on covering new ground is required. Similar to \citet{chaplot2020Learning} and \citet{chen2018learning}, we define a reward term $R_\mathrm{area}$ based on the newly covered area $A_\mathrm{new}$ in the current time step that was not covered previously. To control the scale of this reward signal, we first normalize it to $[0, 1]$ based on the maximum area that can be covered in each time step, which is related to the maximum speed $v_\mathrm{max}$. We subsequently multiply it with the scale factor $\lambda_\mathrm{area}$, which is the maximum possible reward in each step, resulting in the reward
\begin{equation}
    \label{eq_area_reward}
    R_\mathrm{area} = \lambda_\mathrm{area} \frac{A_{\mathrm{new}}}{2 r v_\mathrm{max} \Delta t},
\end{equation}
where $r$ is the agent radius and $\Delta t$ is the time step size. See Figure \ref{fig_reward_area} for an illustration.

By only maximizing the coverage reward in \eqref{eq_area_reward}, the agent is discouraged from overlapping its previous path, as this reduces the reward in the short term. This might lead to holes or stripes in the coverage, which we observed in our experiments (see Appendix \ref{supp_sec_tv_impact}), where the agent would leave a small gap when driving adjacent to a previously covered part. These leftover parts can be costly to cover later on for reaching complete coverage. Covering the thin stripes afterward only yields a minor reward, resulting in slow convergence towards an optimal path. To reduce the leftover parts, we propose a reward term based on minimizing the total variation (TV) of the coverage map. Minimizing the TV corresponds to reducing the boundary of the coverage map, and thus leads to fewer holes. Given a 2D signal $x$, the discrete isotropic total variation, which has been used for image denoising \citep{rudin1992nonlinear}, is expressed as
\begin{equation}
    V(x) = \sum_{i,j} \sqrt{|x_{i+1,j} - x_{i,j}|^2 + |x_{i,j+1} - x_{i,j}|^2}.
\end{equation}
We consider two variants of the TV reward term, a global and an incremental. For the global TV reward $R_\mathrm{TV}^\mathrm{G}$, the agent is given a reward based on the global coverage map $C_t$ at time step $t$. To avoid an unbounded TV for large environments, it is scaled by the square root of the covered area $A_\mathrm{covered}$, as this results in a constant TV for a given shape of the coverage map, independent of scale. The incremental TV reward $R_\mathrm{TV}^\mathrm{I}$ is based on the difference in TV between the current and the previous time step. A positive reward is given if the TV is decreased, and vice versa. The incremental reward is scaled by the maximum possible increase in TV in a time step, which is twice the traveled distance. The global and incremental rewards are respectively given by
\begin{align}
    R_\mathrm{TV}^\mathrm{G}(t) & = - \lambda_\mathrm{TV}^\mathrm{G} \frac{V(C_t)}{\sqrt{A_\mathrm{covered}}}, \\
    R_\mathrm{TV}^\mathrm{I}(t) & = - \lambda_\mathrm{TV}^\mathrm{I} \frac{V(C_t) - V(C_{t-1})}{2 v_\mathrm{max} \Delta t},
\end{align}
where $\lambda_\mathrm{TV}^\mathrm{G}$ and $\lambda_\mathrm{TV}^\mathrm{I}$ are reward scaling parameters to make sure that $|R_\mathrm{TV}| < |R_\mathrm{area}|$ on average. Otherwise, the optimal behaviour is simply to stand still.

To avoid obstacle collisions, a negative reward $R_\mathrm{coll}$ is given each time the agent collides with an obstacle. Finally, a small constant negative reward $R_\mathrm{const}$ is given in each time step to encourage fast execution. Thus, our final reward function reads
\begin{equation}
    R = R_{\mathrm{area}} + R_\mathrm{TV}^\mathrm{G} + R_\mathrm{TV}^\mathrm{I} + R_{\mathrm{coll}} + R_{\mathrm{const}}.
\end{equation}
During training, each episode is terminated when the agent reaches a pre-defined goal coverage, or when it has not covered any new space in $\tau$ consecutive time steps.

\begin{figure}[t]
    \vskip 0.05in
    \centering
    \includegraphics[width=0.99\linewidth]{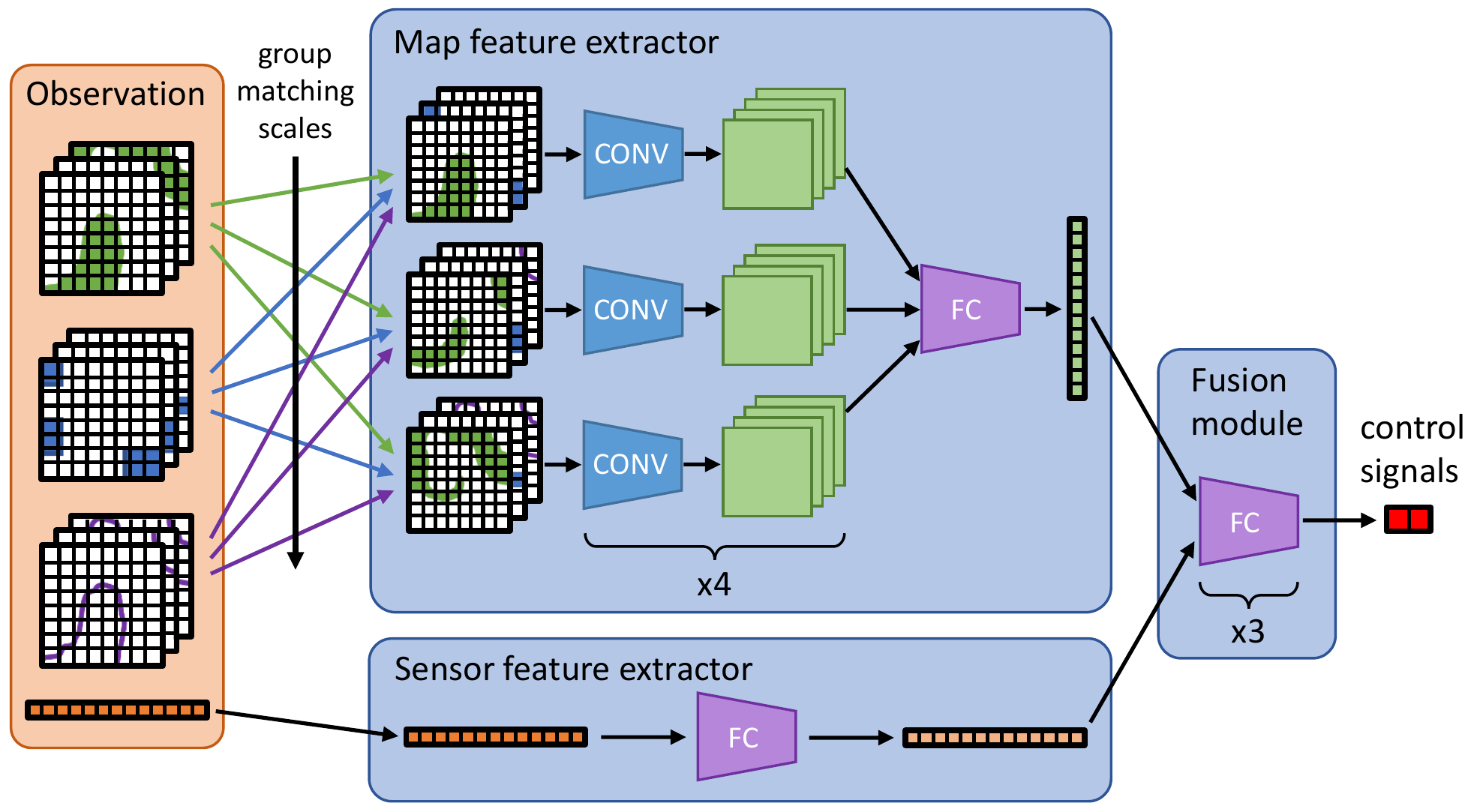}
    \vspace{-6pt}
    \caption{Our proposed SGCNN architecture consists of convolution (CONV) and fully connected (FC) layers. The scales of the multi-scale maps are convolved separately as their spatial positions are not aligned in the grid. x3/x4 refer to the number of layers.}
    \label{fig_cnn_architecture}
    \vspace{-8pt}
\end{figure}

\subsection{Agent Architecture}
\label{sec_agent_architecture}

Due to the multi-modal nature of the observation space, we use a map feature extractor $g_m$, a sensor feature extractor $g_s$, and a fusing module $g_f$. The map and sensor features are fused, resulting in the control signal prediction
\begin{equation}
    (v, \omega) = g_f ( g_m(M_c, M_o, M_f) , g_s(S) ).
\end{equation}

We consider three network architectures, a simple multilayer perceptron (MLP), a standard convolutional neural network (CNN), and our proposed scale-grouped convolutional neural network (SGCNN) that independently processes the different scales in the maps, see Figure \ref{fig_cnn_architecture}. The MLP is mainly used as a benchmark to evaluate the inductive priors in the CNN-based architectures. For the MLP, the feature extractors, $g_m$ and $g_s$, are identity functions that simply flatten the inputs, and the fusing module consists of three fully connected (FC) layers. The CNN-based architectures use convolutional layers in the map feature extractor followed by a single FC layer. In SGCNN, we group the maps in $M_c$, $M_o$ and $M_f$ by scale, as the pixel positions between different scales do not correspond to the same world coordinate. Each scale is convolved separately using grouped convolutions. This ensures that each convolution kernel is applied to grids where the spatial context is consistent across channels.
The sensor feature extractor is a single FC layer, and the fusing module consists of three FC layers. Full details can be found in Appendix~\ref{supp_sec_agent_architecture}.

\section{Experiments}
\label{sec_experiments}

In this section, we first describe the implementation details regarding training, environment setup, and evaluation in \cref{sec_implementation_details}. Subsequently, we evaluate our method in three settings; both \textit{omnidirectional} and \textit{non-omnidirectional exploration} in \cref{sec_exploration}, and on the \textit{lawn mowing} task in \cref{sec_lawn_mowing}. Finally, we evaluate the robustness to noise in \cref{sec_robustness_to_noise} and provide ablation studies in \cref{sec_ablation_study}.

\subsection{Implementation Details}
\label{sec_implementation_details}

\textbf{Agent and training details.} We use soft actor-critic (SAC) RL \citep{haarnoja2018soft} with automatic temperature tuning \citep{haarnoja2018soft_2}. The actor and critic networks share the network architecture, but not any weights. We train for $8$M iterations with learning rate $10^{-5}$, batch size $256$, replay buffer size $5 \cdot 10^5$, discount factor $\gamma = 0.99$, and a minimal noise level. The simulated lidar field-of-view was $360^\circ$ in omnidirectional exploration, and $180^\circ$ in the other two settings. The coverage radius was set to the lidar range in both exploration tasks, which was $7$~m and $3.5$~m in omnidirectional and non-omnidirectional exploration respectively. In the lawn mowing case, the coverage radius was $0.15$ m, same as the agent radius. For full details on the physical dimensions, see Appendix \ref{supp_sec_physical_dimensions}. The training time for one agent varied between $100$ to $150$ hours on a T4 GPU and a 6226R CPU.

\begin{figure*}[!t]
    \vskip 0.05in
    \centering
    \setlength{\tabcolsep}{0.5pt}
    \setlength{\fboxsep}{0pt}%
    \setlength{\fboxrule}{0.5pt}%
    \begin{tabular}{cccccccc}
        \fbox{\includegraphics[width=.11\linewidth]{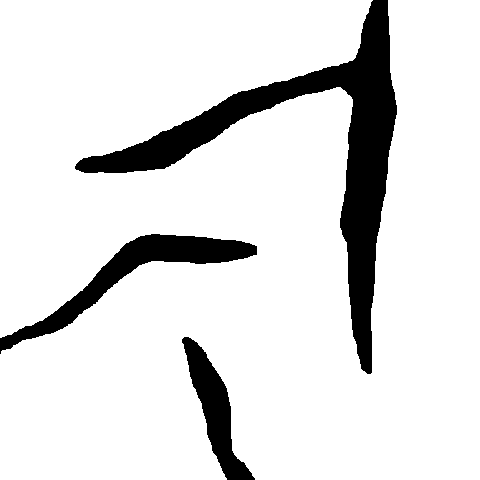}} &
        \fbox{\includegraphics[width=.11\linewidth]{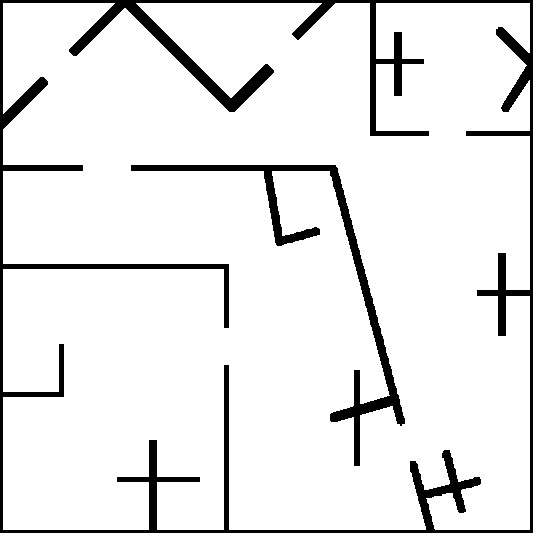}} &
        \fbox{\includegraphics[width=.11\linewidth]{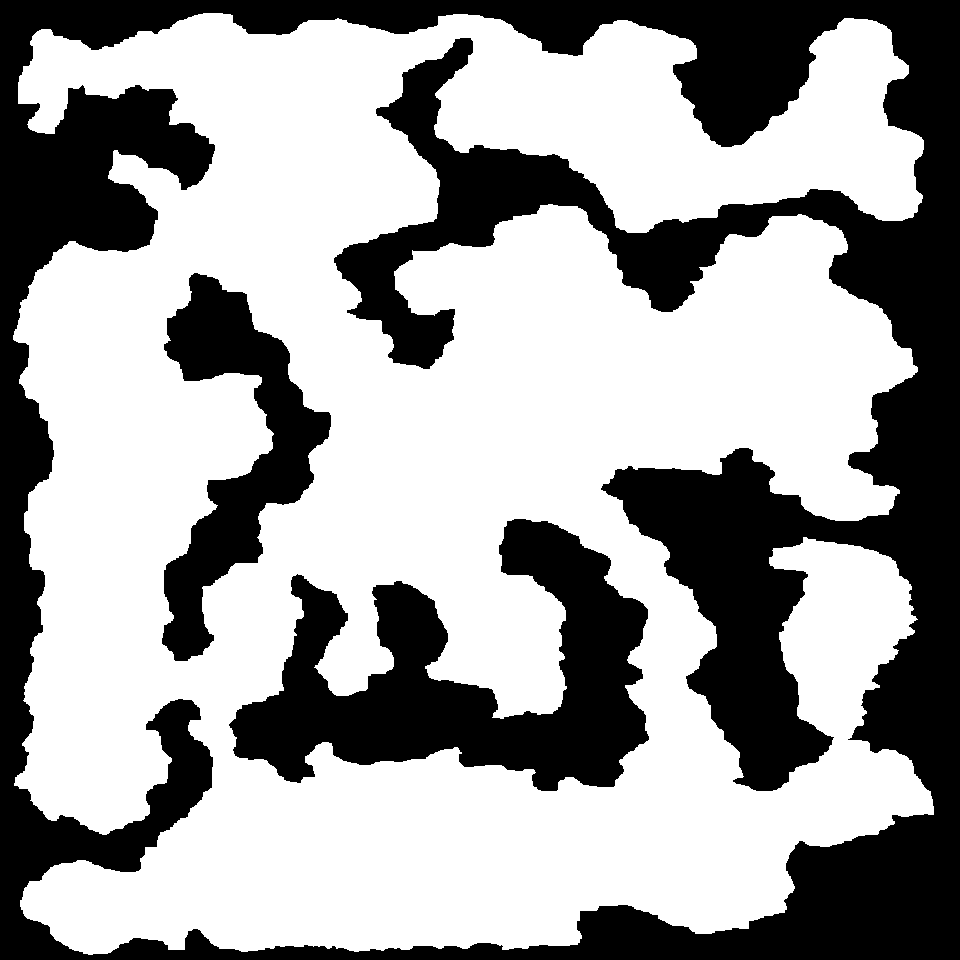}} &
        \fbox{\includegraphics[width=.11\linewidth]{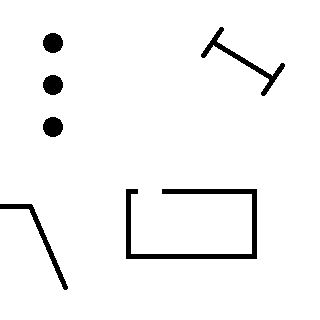}} &
        \fbox{\includegraphics[width=.11\linewidth]{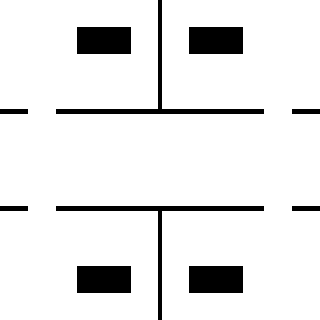}} &
        \fbox{\includegraphics[width=.11\linewidth]{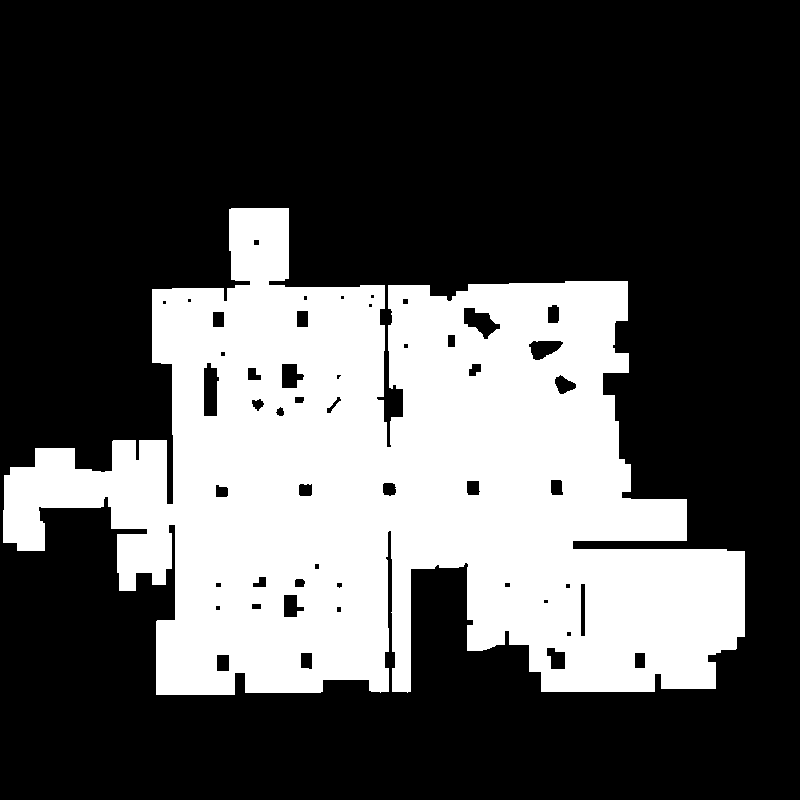}} &
        \fbox{\includegraphics[width=.11\linewidth]{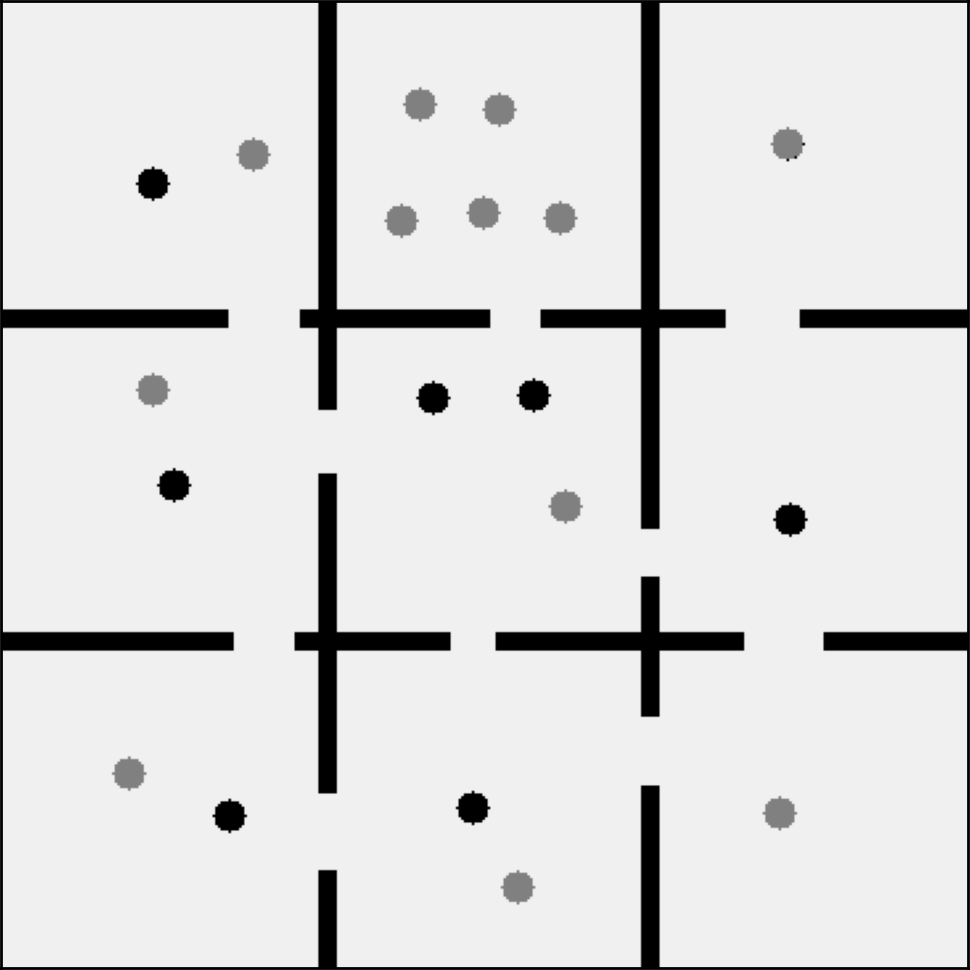}} &
        \fbox{\includegraphics[width=.11\linewidth]{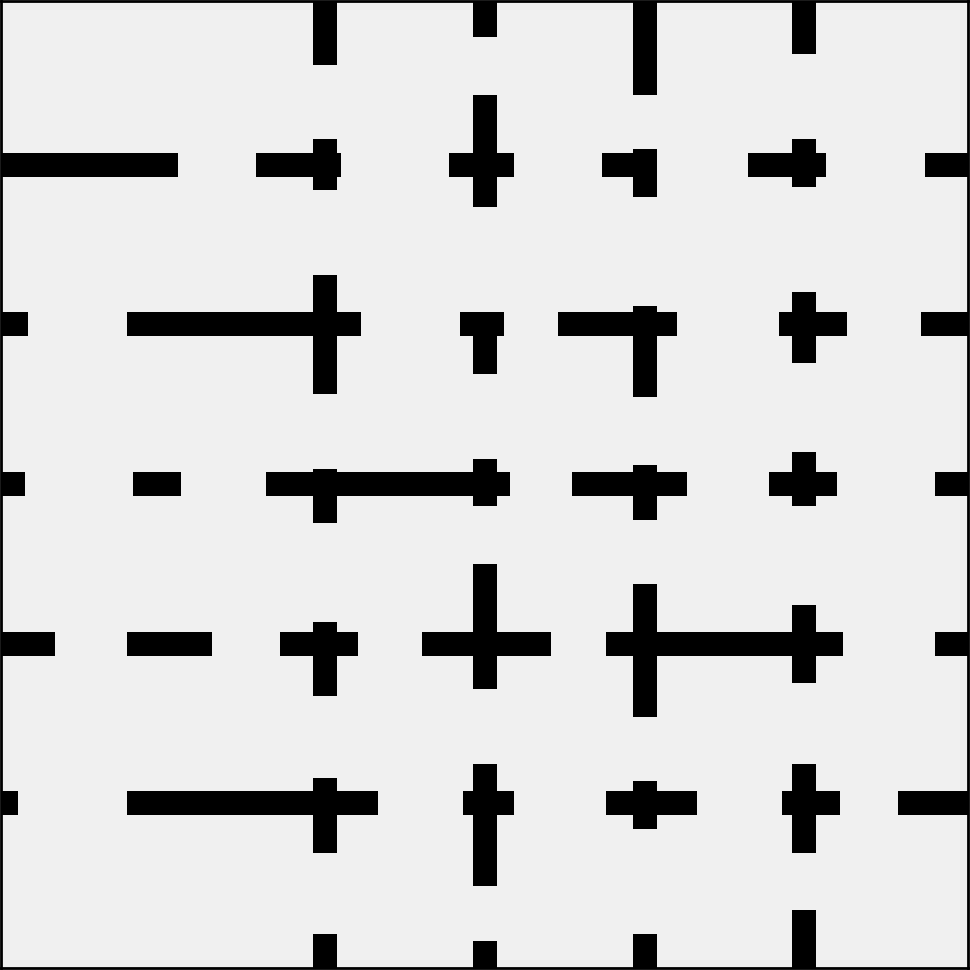}} \\
        (a) & (b) & (c) & (d) & (e) & (f) & (g) & (h) \\
    \end{tabular}
    \vspace{-6pt}
    \caption{Examples of exploration maps (a-c), lawn mowing maps (d-f), and randomly generated maps (g-h).}
    \label{fig_maps}
\end{figure*}

\begin{table*}[!t]
    \vspace{-6pt}
    \setlength{\tabcolsep}{5pt}
    \centering
    \caption{Time in seconds for reaching $90\%$ and $99\%$ coverage in Explore-Bench.}
    \vspace{6pt}
    \small
    \begin{tabular}{lcccccccccccccc}
        \toprule
        Map $\rightarrow$ & \multicolumn{2}{c}{Loop} & \multicolumn{2}{c}{Corridor} & \multicolumn{2}{c}{Corner} & \multicolumn{2}{c}{Rooms} & \multicolumn{2}{c}{Comb.\ 1} & \multicolumn{2}{c}{Comb.\ 2} & \multicolumn{2}{c}{Total} \\
        \cmidrule(l{\tabcolsep}r{\tabcolsep}){2-3}
        \cmidrule(l{\tabcolsep}r{\tabcolsep}){4-5}
        \cmidrule(l{\tabcolsep}r{\tabcolsep}){6-7}
        \cmidrule(l{\tabcolsep}r{\tabcolsep}){8-9}
        \cmidrule(l{\tabcolsep}r{\tabcolsep}){10-11}
        \cmidrule(l{\tabcolsep}r{\tabcolsep}){12-13}
        \cmidrule(l{\tabcolsep}r{\tabcolsep}){14-15}
        Method $\downarrow$ & $T_{90}$ & $T_{99}$ & $T_{90}$ & $T_{99}$ & $T_{90}$ & $T_{99}$ & $T_{90}$ & $T_{99}$ & $T_{90}$ & $T_{99}$ & $T_{90}$ & $T_{99}$ & $T_{90}$ & $T_{99}$ \\
        \midrule
        Distance frontier & 124 & 145 & 162 & 169 & 210 & 426 & 159 & 210 & 169 & 175 & 230 & 537 & 1054 & 1662 \\
        RRT frontier & 145 & 180 & 166 & 170 & 171 & 331 & 176 & 211 & 141 & 192 & 249 & 439 & 1048 & 1523 \\
        Potential field frontier & 131 & 152 & 158 & 162 & 133 & 324 & 156 & 191 & 165 & 183 & 224 & 547 & 967 & 1559 \\
        Active Neural SLAM & 190 & 214 & 160 & 266 & 324 & 381 & 270 & 315 & 249 & 297 & 588 & 755 & 1781 & 2228 \\
        \textbf{Ours} & \textbf{89} & \textbf{101} & \textbf{98} & \textbf{128} & \textbf{80} & \textbf{291} & \textbf{82} & \textbf{99} & \textbf{87} & \textbf{93} & \textbf{120} & \textbf{349} & \textbf{556} & \textbf{1061} \\
        \bottomrule
    \end{tabular}
    \label{table_explore_bench}
\end{table*}

\textbf{Environment details.} For the multi-scale maps, we use $m = 4$ scales with $32 \times 32$ pixel resolution, a scale factor of $s = 4$, and $0.0375$ meters per pixel for the finest scale, which corresponds to $4$ pixels per agent radius. Thus, the maps span a square with side length $76.8$ m. We progressively increase the difficulty of the environment during training, starting from simple maps and increase their complexity over time. The episodes are terminated when a goal coverage rate between $90-99\%$ is reached, depending on the training progression. For full details on the progressive training, see Appendix \ref{supp_sec_progressive_training}. Based on initial experiments, we find suitable reward parameters. The episodes are prematurely truncated if $\tau = 1000$ consecutive steps have passed without the agent covering any new space. We set the maximum coverage reward $\lambda_\mathrm{area} = 1$, the incremental TV reward scale $\lambda_\mathrm{TV}^\mathrm{I} = 0.2$ for exploration and $\lambda_\mathrm{TV}^\mathrm{I} = 1$ for lawn mowing, the collision reward $R_\mathrm{coll} = -10$, and the constant reward $R_\mathrm{const} = -0.1$. The global TV reward scale was set to $\lambda_\mathrm{TV}^\mathrm{G} = 0$ as it did not contribute to a performance gain in our ablations in Section \ref{sec_ablation_study}.

\textbf{Map geometry.} To increase the variation of encountered scenarios, we use both a fixed set of maps and randomly generated maps, see Figure \ref{fig_maps} for examples. The fixed maps provide both simple, and more challenging scenarios that a CPP method is expected to solve. The randomly generated maps provide orders of magnitude more variation during training, and avoid the problem of overfitting to the fixed maps. They are created by randomizing grid-like floor plans with obstacles. For more details, see Appendix \ref{supp_sec_random_map_generation}.

\textbf{Evaluation and comparison.} We measure the times $T_{90}$ and $T_{99}$ to reach $90\%$ and $99\%$ coverage, respectively. For inference times and collision statistics, see \cref{supp_sec_inference_time,supp_sec_collision}. We use separate maps for evaluation that are not seen during training, see Appendix \ref{supp_sec_evaluation_maps} for a full list. We compare our approach across three CPP variations, to a total of seven previous methods that are tailored for either exploration or the lawn mowing task. For exploration, we evaluate our method on Explore-Bench \citep{xu2022explore}, which is a recent benchmark that implements challenging environments, where four methods have been evaluated by the authors. These include three frontier-based methods, namely a distance-based frontier method \citep{yamauchi1997frontier}, an RRT-based frontier method \citep{umari2017autonomous}, and a potential field-based frontier method \citep{yu2021smmr}. The fourth method is an RL-based approach, where \citet{xu2022explore} train an RL model to determine a global goal based on active neural SLAM \citep{chaplot2020Learning}. Finally, we reimplement a fifth method by \citet{hu2020tovt} that uses RL to navigate to a chosen frontier node. For the lawn mowing task, we compare with the backtracking spiral algorithm (BSA) \citep{gonzalez2005icra}, which is a common benchmark for this CPP variation. Additionally, we implement a baseline that combines A* \citep{hart1968formal} with a travelling salesman problem (TSP) solver on nodes in a grid-like configuration. This has previously been proposed for the lawn mowing task in the offline setting for small environments \citep{bormann2018indoor}. To make it feasible for our larger environments, a heuristic was required for distant nodes, and periodic replanning was used for the online case. For full details, see Appendix \ref{supp_sec_implementation_compared_methods}. We provide videos showcasing the learned paths of our approach.\footnote{Videos: \url{https://drive.google.com/drive/folders/18s1pKaMHwABli4meCwUF6qMsJ4-jPZvU?usp=sharing}}

\begin{table*}[!t]
    \setlength{\tabcolsep}{5pt}
    \centering
    \caption{Time in minutes for reaching $90\%$ and $99\%$ coverage on the lawn mowing task.}
    \label{table_mowing}
    \vspace{6pt}
    \small
    \begin{tabular}{clcccccccccccccc}
        \toprule
        & Map $\rightarrow$ & \multicolumn{2}{c}{Map 1} & \multicolumn{2}{c}{Map 2} & \multicolumn{2}{c}{Map 3} & \multicolumn{2}{c}{Map 4} & \multicolumn{2}{c}{Map 5} & \multicolumn{2}{c}{Map 6} & \multicolumn{2}{c}{Total} \\
        \cmidrule(l{\tabcolsep}r{\tabcolsep}){3-4}
        \cmidrule(l{\tabcolsep}r{\tabcolsep}){5-6}
        \cmidrule(l{\tabcolsep}r{\tabcolsep}){7-8}
        \cmidrule(l{\tabcolsep}r{\tabcolsep}){9-10}
        \cmidrule(l{\tabcolsep}r{\tabcolsep}){11-12}
        \cmidrule(l{\tabcolsep}r{\tabcolsep}){13-14}
        \cmidrule(l{\tabcolsep}r{\tabcolsep}){15-16}
        Setting & Method $\downarrow$ & $T_{90}$ & $T_{99}$ & $T_{90}$ & $T_{99}$ & $T_{90}$ & $T_{99}$ & $T_{90}$ & $T_{99}$ & $T_{90}$ & $T_{99}$ & $T_{90}$ & $T_{99}$ & $T_{90}$ & $T_{99}$ \\
        \midrule
        \multirow{2}{*}{\textit{Offline}} & TSP & 45 & 49 & 45 & 51 & 43 & 48 & 55 & 61 & 27 & 30 & 110 & 122 & 325 & 361 \\
        & BSA & \textbf{30} & \textbf{35} & \textbf{29} & \textbf{35} & \textbf{31} & \textbf{36} & \textbf{34} & \textbf{41} & \textbf{17} & \textbf{23} & \textbf{88} & \textbf{100} & \textbf{229} & \textbf{270} \\
        \midrule
        \multirow{2}{*}{\textit{Online}} & TSP & 62 & 69 & 70 & 77 & 67 & 75 & 70 & 78 & 37 & 41 & 142 & 158 & 448 & 498 \\
        & \textbf{Ours} & \textbf{44} & \textbf{60} & \textbf{40} & \textbf{50} & \textbf{43} & \textbf{49} & \textbf{40} & \textbf{69} & \textbf{25} & \textbf{32} & \textbf{118} & \textbf{149} & \textbf{310} & \textbf{409} \\
        \bottomrule
    \end{tabular} 
    \vspace{6pt}
\end{table*}

\begin{table*}[t]
\begin{minipage}{0.35\linewidth}
    \setlength{\tabcolsep}{5pt}
    \centering
    \vspace{-10pt}
    \makeatletter\def\@captype{table}\makeatother
    \caption{Coverage time in minutes for different levels of noise in the lawn mowing task.}
    \vspace{10pt}
    \small
    \begin{tabular}{ccccc}
        \toprule
        \multicolumn{3}{c}{Noise standard deviation} & \multicolumn{2}{c}{Time} \\
        \cmidrule(l{\tabcolsep}r{\tabcolsep}){1-3}
        \cmidrule(l{\tabcolsep}r{\tabcolsep}){4-5}
        Position & Heading & Lidar & $T_{90}$ & $T_{99}$ \\
        \midrule
        0.01 m & 0.05 rad & 0.05 m & 310 & 409 \\
        0.02 m & 0.1 rad & 0.1 m & 338 & 486 \\
        0.05 m & 0.2 rad & 0.2 m & 317 & 434 \\
        \bottomrule
    \end{tabular}
    \label{table_noise}
\end{minipage}%
\hfill
\begin{minipage}{0.3\linewidth}
    \makeatletter\def\@captype{figure}\makeatother
    \centering
    \includegraphics[height=0.72\linewidth]{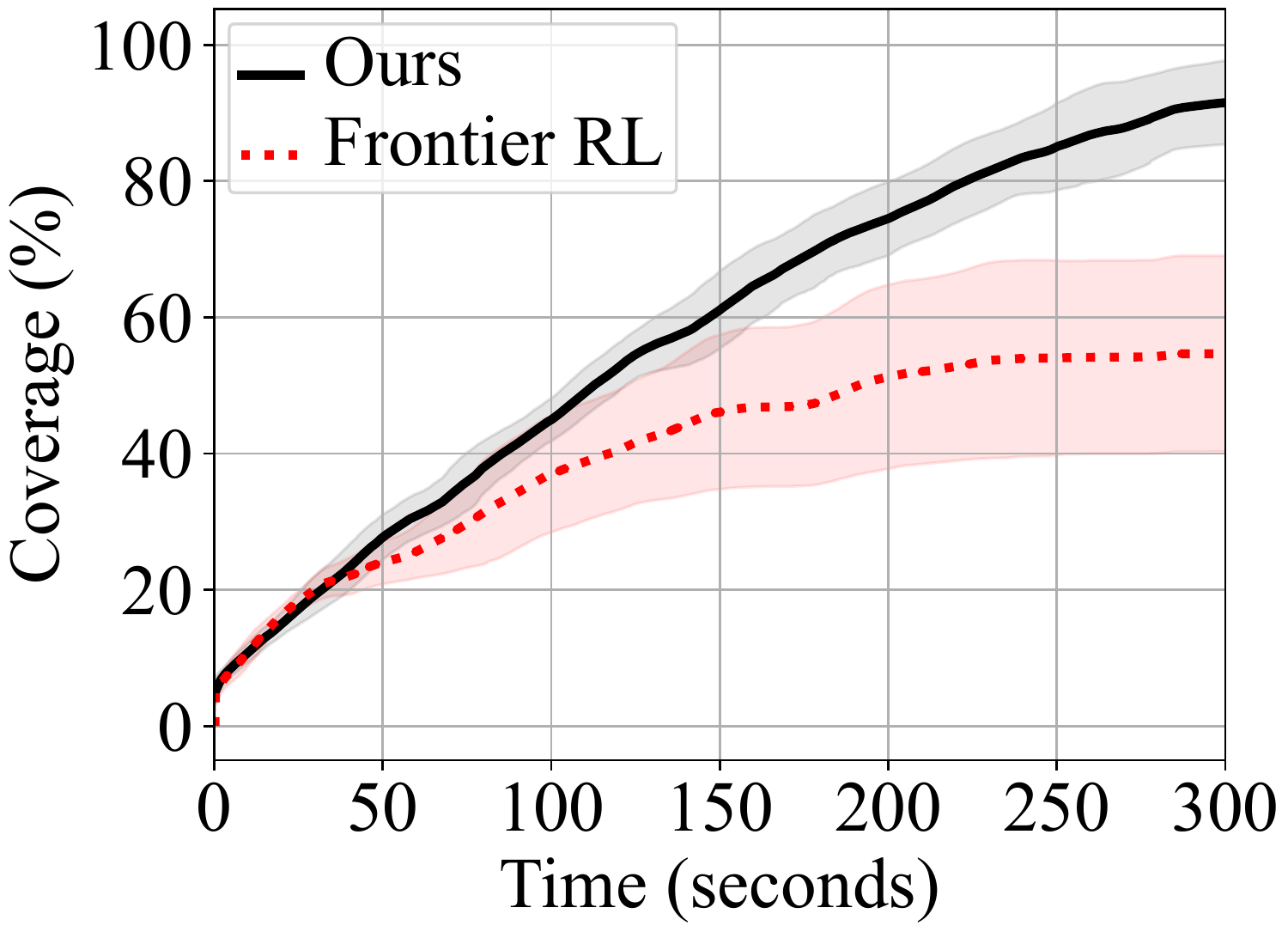}
    \vspace{-20pt}
    \caption{Coverage over time in non-omnidirectional exploration.}
    \label{fig_non_omni}
\end{minipage}%
\hfill
\begin{minipage}{0.3\linewidth}
    \makeatletter\def\@captype{figure}\makeatother
    \centering
    \includegraphics[height=0.72\linewidth]{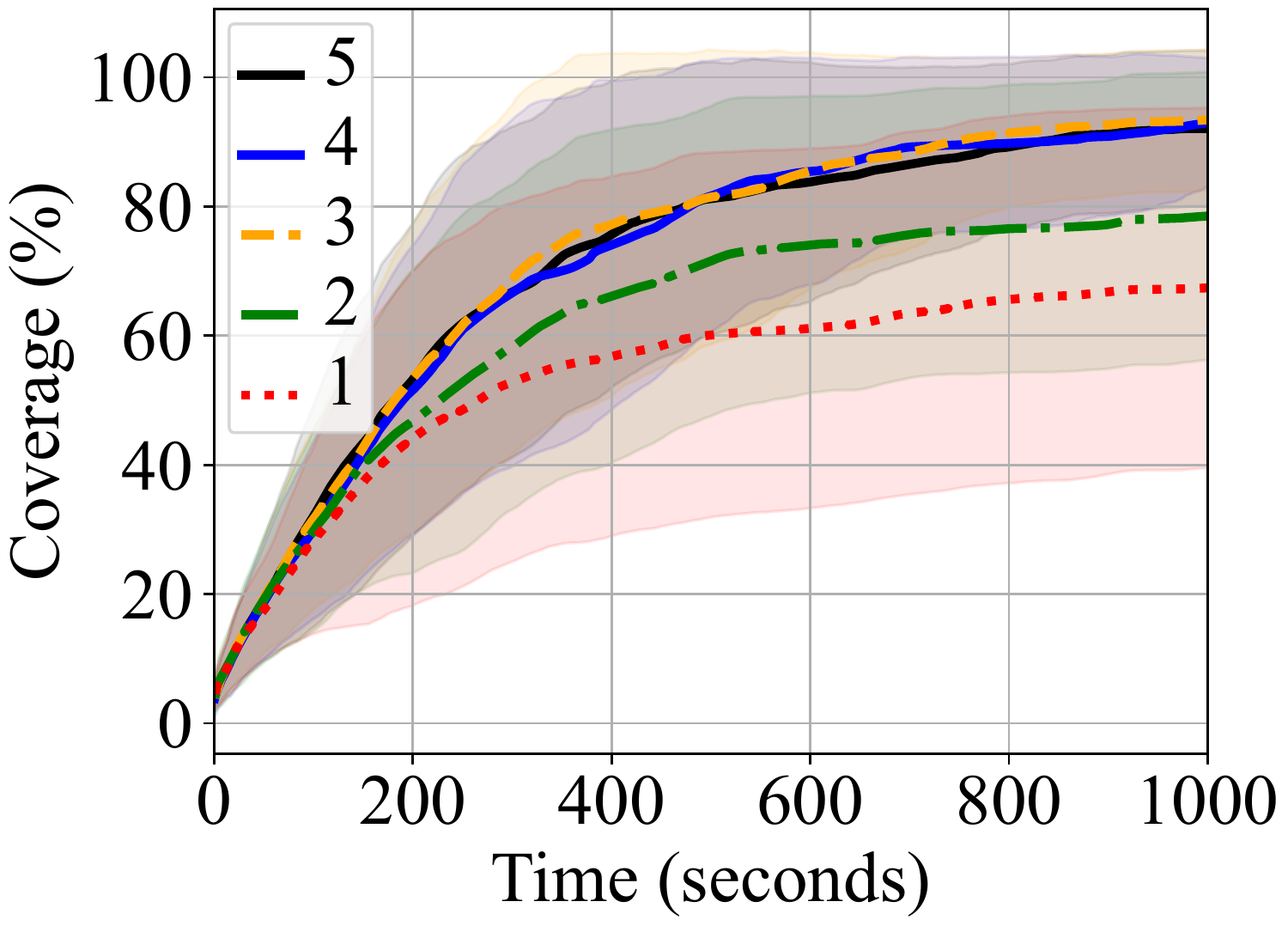}
    \vspace{-20pt}
    \caption{Exploration coverage for different number of scales.}
    \label{fig_num_maps}
\end{minipage}
\end{table*}

\subsection{Exploration}
\label{sec_exploration}

\textbf{Omnidirectional exploration.} In omnidirectional exploration, the agent observes its surroundings in all directions through a 360$^\circ$ field-of-view lidar sensor. To evaluate our method on this CPP variation, we use Explore-Bench \citep{xu2022explore}, which includes six environments; loop, corridor, corner, rooms, combination 1 (rooms with corridors), and combination 2 (complex rooms with tight spaces), which can be found in Appendix \ref{supp_sec_evaluation_maps}. The results are presented in Table \ref{table_explore_bench}. Our approach surpasses the performance of both the frontier-based methods and Active Neural SLAM. This shows that learning control signals end-to-end with RL is in fact a suitable approach to CPP. Furthermore, \cref{fig_qualitative_paths} shows that the agent has learned an efficient exploration path in a complex and obstacle-cluttered environment.

\textbf{Non-omnidirectional exploration.} We further evaluate our method on non-omnidirectional exploration, with a 180$^\circ$ lidar field-of-view. We compare with the recent frontier-based RL method of \citet{hu2020tovt}, which was trained specifically for this setting. The coverage over time is presented in Figure \ref{fig_non_omni}. Our method outperforms the frontier-based RL approach, demonstrating that end-to-end learning of control signals is superior to a multi-stage approach for adapting to a specific sensor setup.

\subsection{Lawn Mowing}
\label{sec_lawn_mowing}

For the lawn mowing problem, we compare with the backtracking spiral algorithm (BSA) \citep{gonzalez2005icra}. Note however that BSA is an offline method and does not solve the mapping problem. As such, we do not expect our approach to outperform it in this comparison. Instead, we use it to see how close we are to a good solution. We further compare with an offline and online version of the TSP-based solution with A* \citep{bormann2018indoor}. \Cref{table_mowing} shows $T_{90}$ and $T_{99}$ for six maps numbered 1-6, which can be found in Appendix \ref{supp_sec_evaluation_maps}. Compared to offline BSA, our method takes $35\%$ and $51\%$ more time to reach $90\%$ and $99\%$ coverage respectively. This is an impressive result considering the challenge of simultaneously mapping the environment. Moreover, our approach outperforms the online version of the TSP-based solution, and even surpasses the offline version for $90\%$ coverage. The limiting factors for TSP are likely the grid discretization and suboptimal replanning, which lead to overlap, see \cref{supp_sec_implementation_compared_methods}.

\subsection{Robustness to Noise}
\label{sec_robustness_to_noise}

As the real world is noisy, we evaluate how robust our method is to noise. We apply Gaussian noise to the position, heading, and lidar measurements during both training and evaluation. Thus, both the maps and the sensor data perceived by the agent contain noise. We consider three levels of noise and present $T_{90}$ and $T_{99}$ for the lawn mowing task in \cref{table_noise}. The result shows that, even under high levels of noise, our method still functions well. Our approach surpasses the TSP solution under all three noise levels.

\begin{table}[t]
    \vspace{-10pt}
    \setlength{\tabcolsep}{3pt}
    \centering
    \caption{Coverage ($\%$) at 1500 and 1000 seconds for mowing (Mow) and exploration (Exp) respectively, comparing agent architecture (NN), TV rewards, and frontier map observation ($M_f$).}
    \vspace{6pt}
    \small
    \begin{tabular}{ccccccc}
        \toprule
        \multicolumn{5}{c}{Settings} & \multicolumn{2}{c}{Coverage} \\
        \cmidrule(l{\tabcolsep}r{\tabcolsep}){1-5}
        \cmidrule(l{\tabcolsep}r{\tabcolsep}){6-7}
        &$R_\mathrm{TV}^{I}$ & $R_\mathrm{TV}^{G}$ & $M_f$ & NN & Mow & Exp \\
        \midrule
        \ding{172}&\checkmark & & \checkmark & MLP & 81.4 & 27.3 \\
        \ding{173}&\checkmark & & \checkmark & CNN & 93.2 & 88.4 \\
        \ding{174}&& & \checkmark & SGCNN & 85.1 & 91.5 \\
        \ding{175}&\checkmark & & & SGCNN & 72.6 & 80.6 \\
        \ding{176}&\checkmark & \checkmark & \checkmark & SGCNN & 96.6 & 89.0 \\
        \ding{177}&\checkmark & & \checkmark & SGCNN & \textbf{97.8} & \textbf{93.0} \\
        \bottomrule
    \end{tabular}
    \label{table_ablation}
    \vspace{-8pt}
\end{table}

\begin{figure*}[!t]
    \vskip 0.05in
    \centering
    \setlength{\tabcolsep}{0.5pt}
    \setlength{\fboxsep}{0pt}%
    \setlength{\fboxrule}{0.5pt}%
    \begin{tabular}{cccccc}
        \includegraphics[width=.18\linewidth]{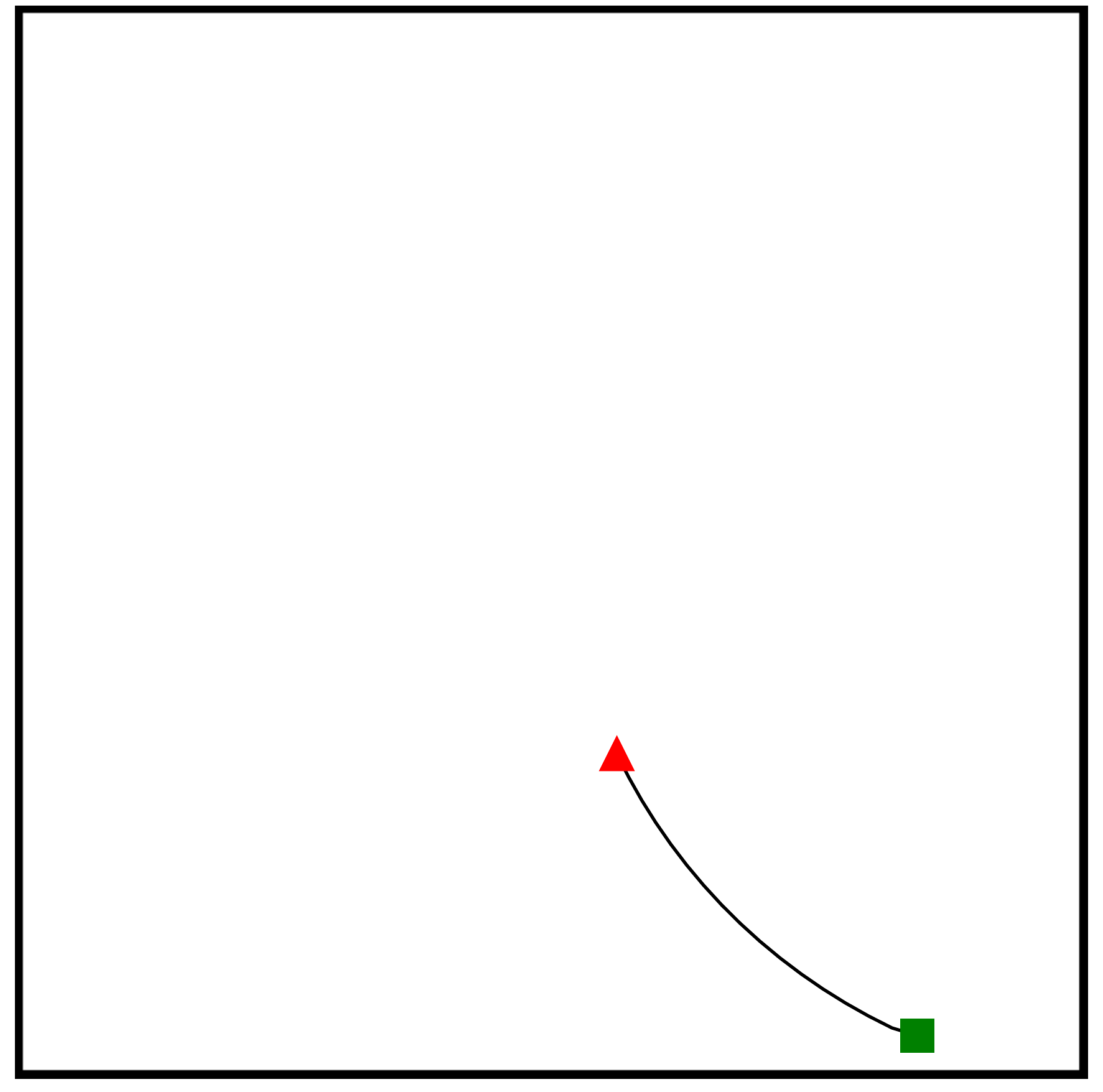} &
        \includegraphics[width=.18\linewidth]{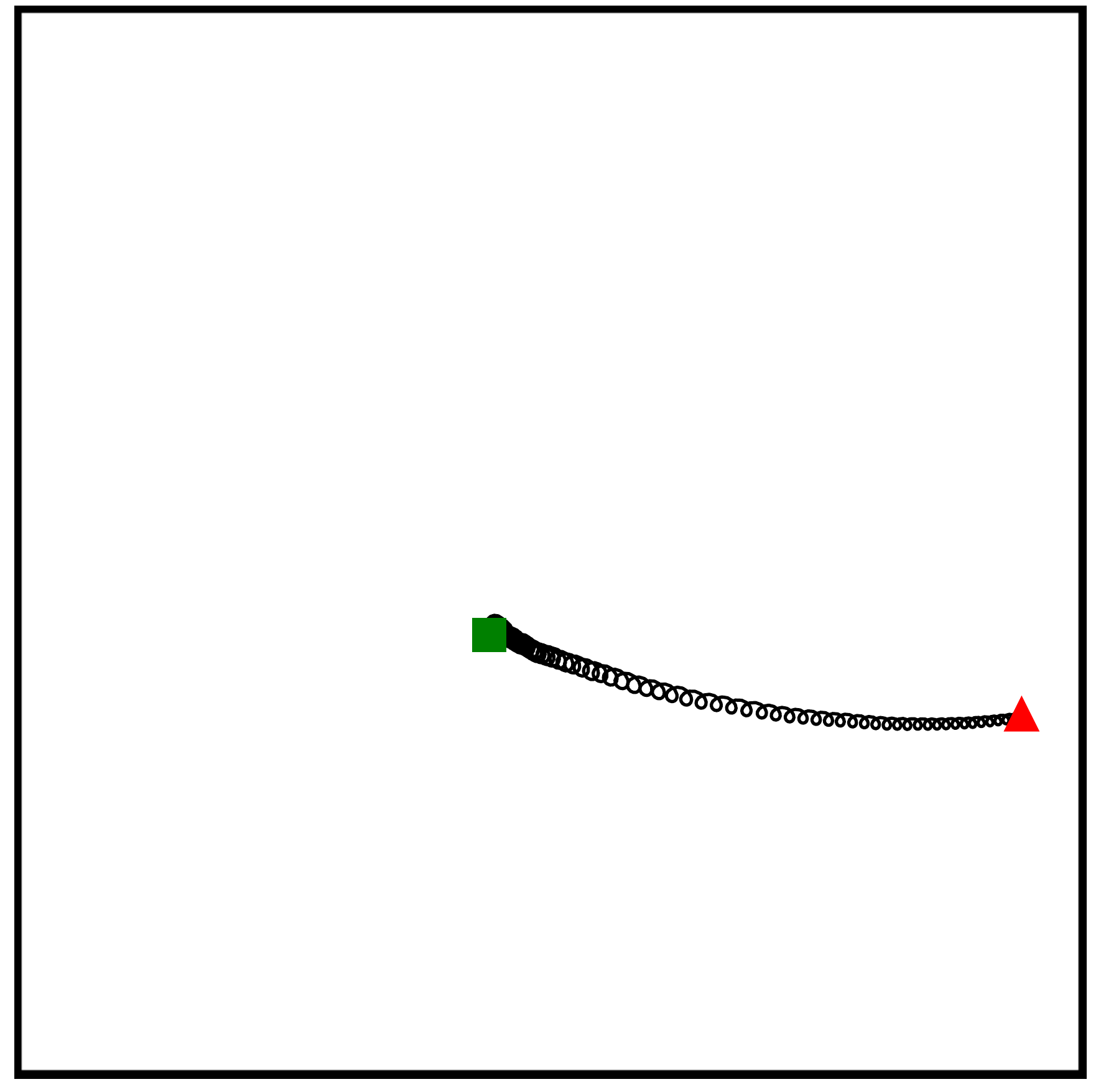} &
        \includegraphics[width=.18\linewidth]{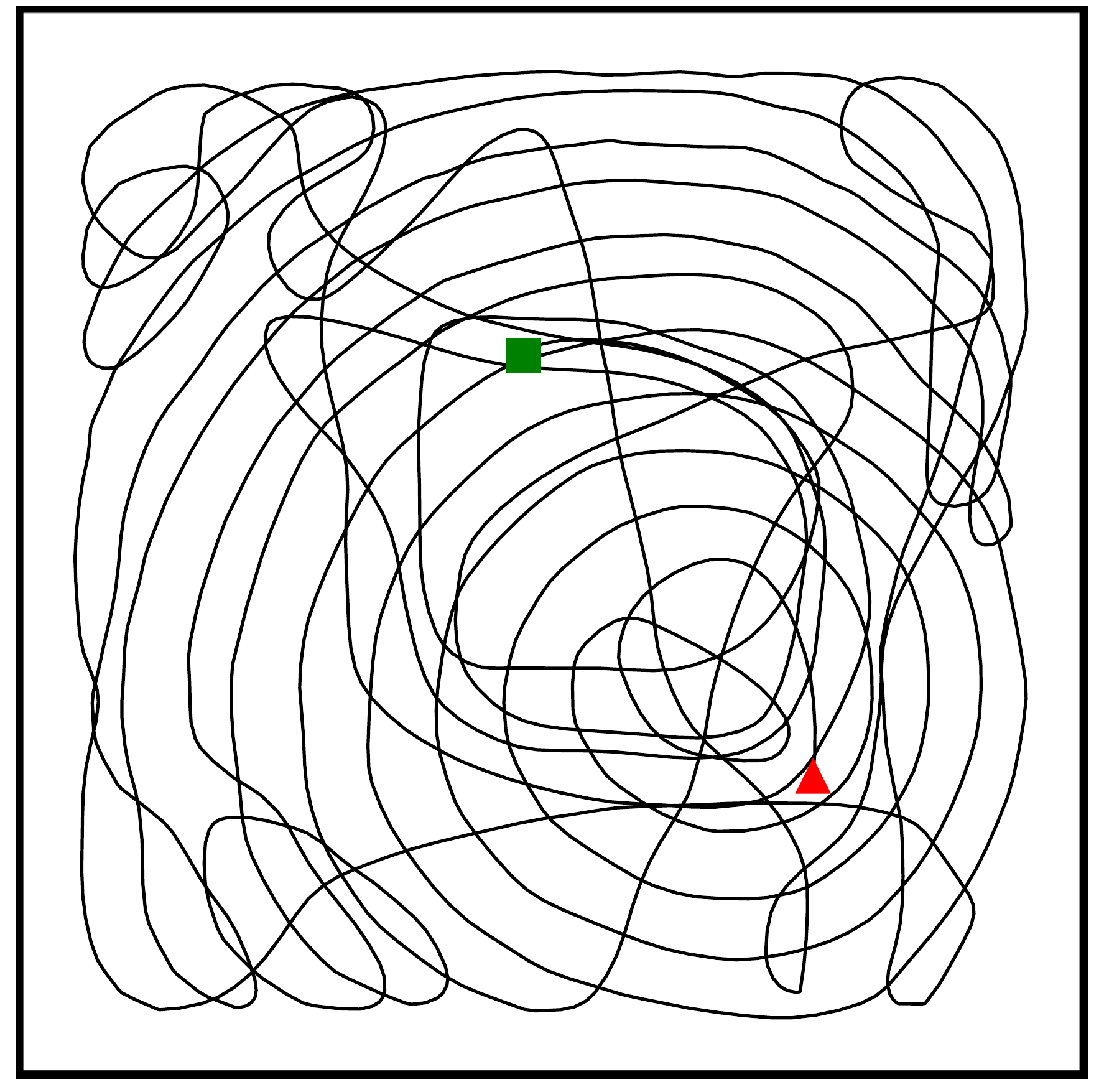} &
        \includegraphics[width=.18\linewidth]{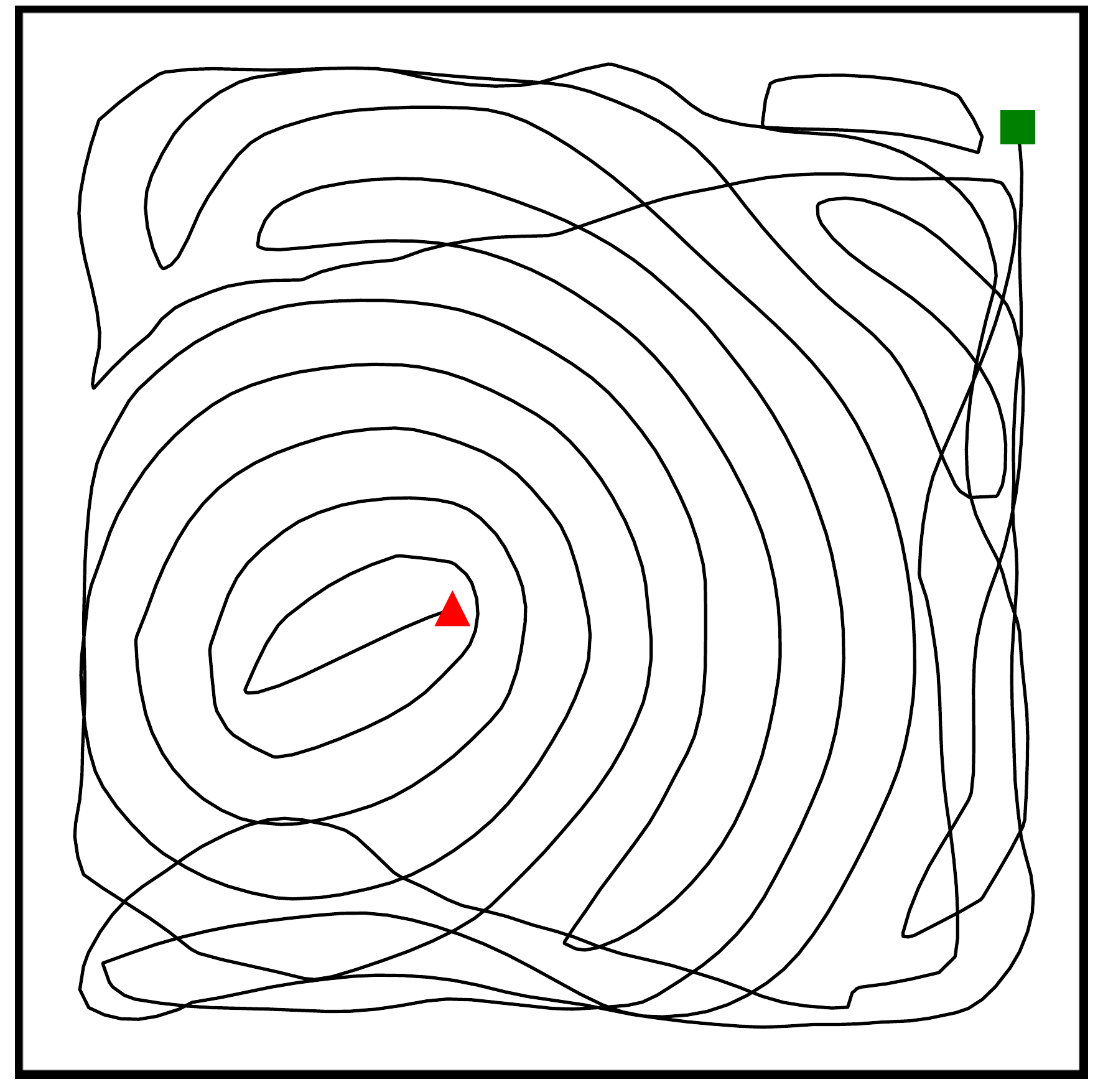} &
        \includegraphics[width=.18\linewidth]{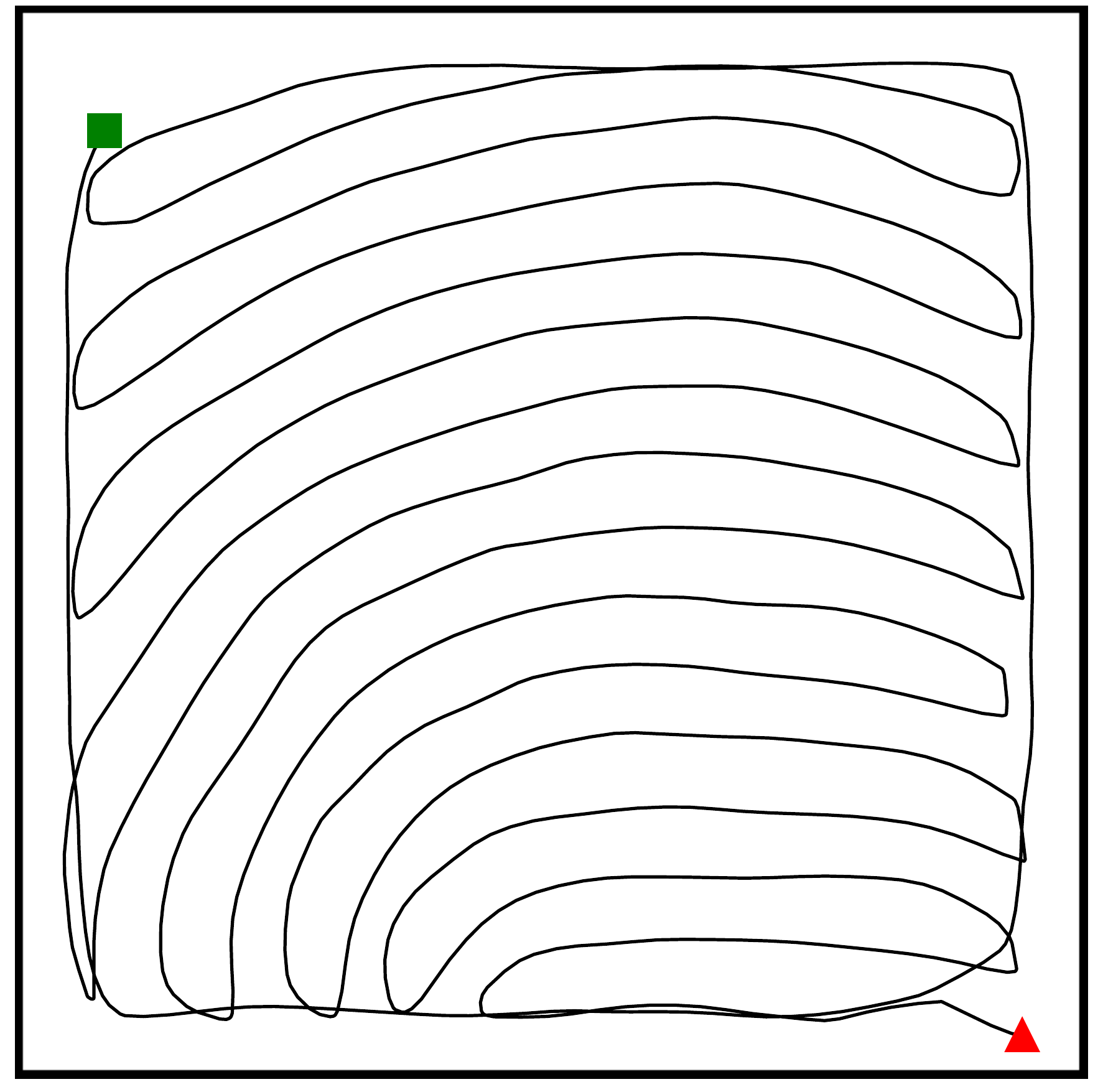} \\
        1k iterations & 10k iterations & 100k iterations & 1M iterations & 8M iterations \\
    \end{tabular}
    \vspace{-6pt}
    \caption{Learned paths on the lawn mowing task after different number of training iterations.}
    \label{fig_paths_training_steps}
\end{figure*}

\begin{figure*}[!t]
    \vskip 0.05in
    \centering
    \setlength{\tabcolsep}{0.5pt}
    \setlength{\fboxsep}{0pt}%
    \setlength{\fboxrule}{0.5pt}%
    \begin{tabular}{cccccc}
        \includegraphics[width=.18\linewidth]{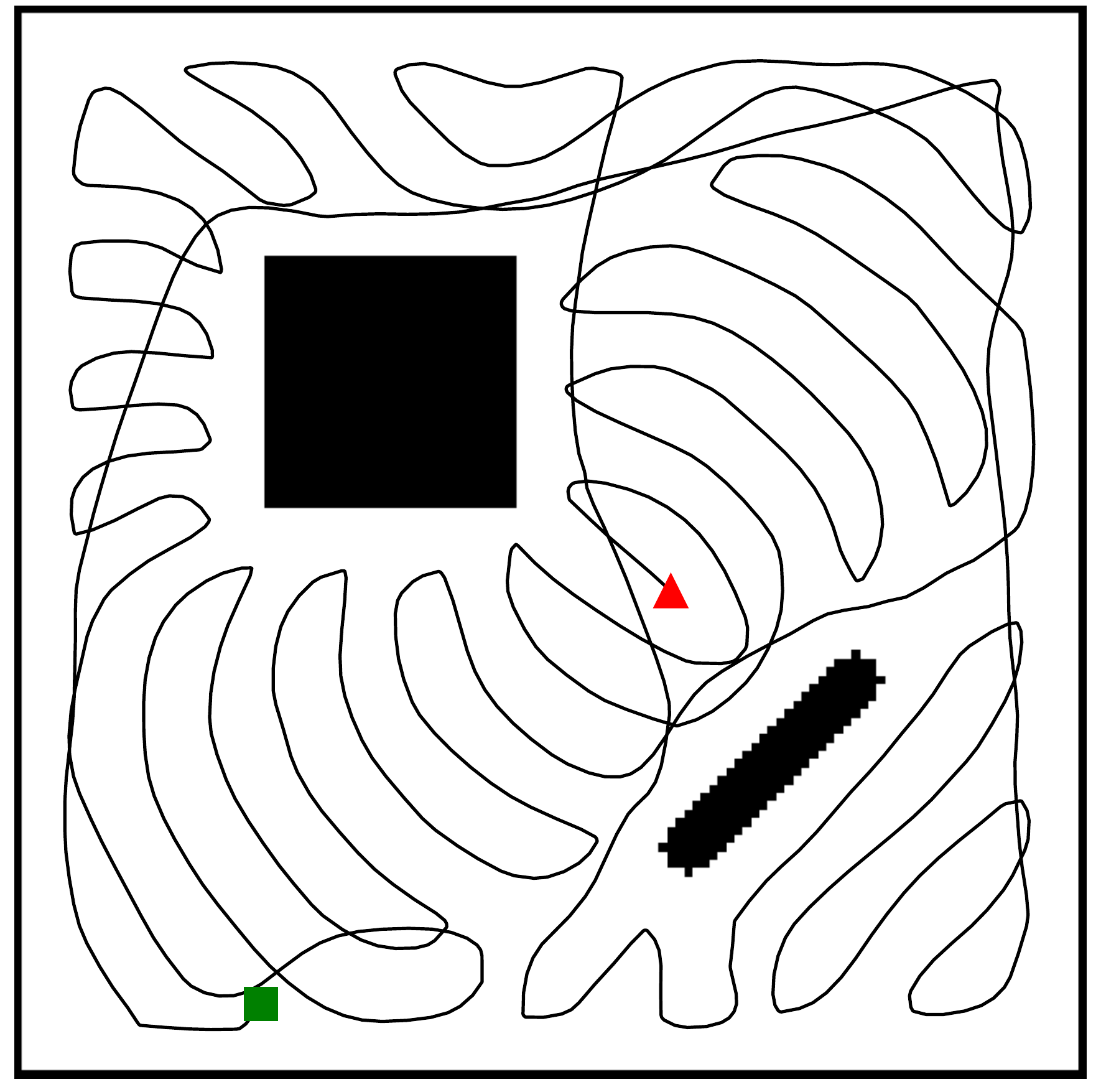} &
        \includegraphics[width=.18\linewidth]{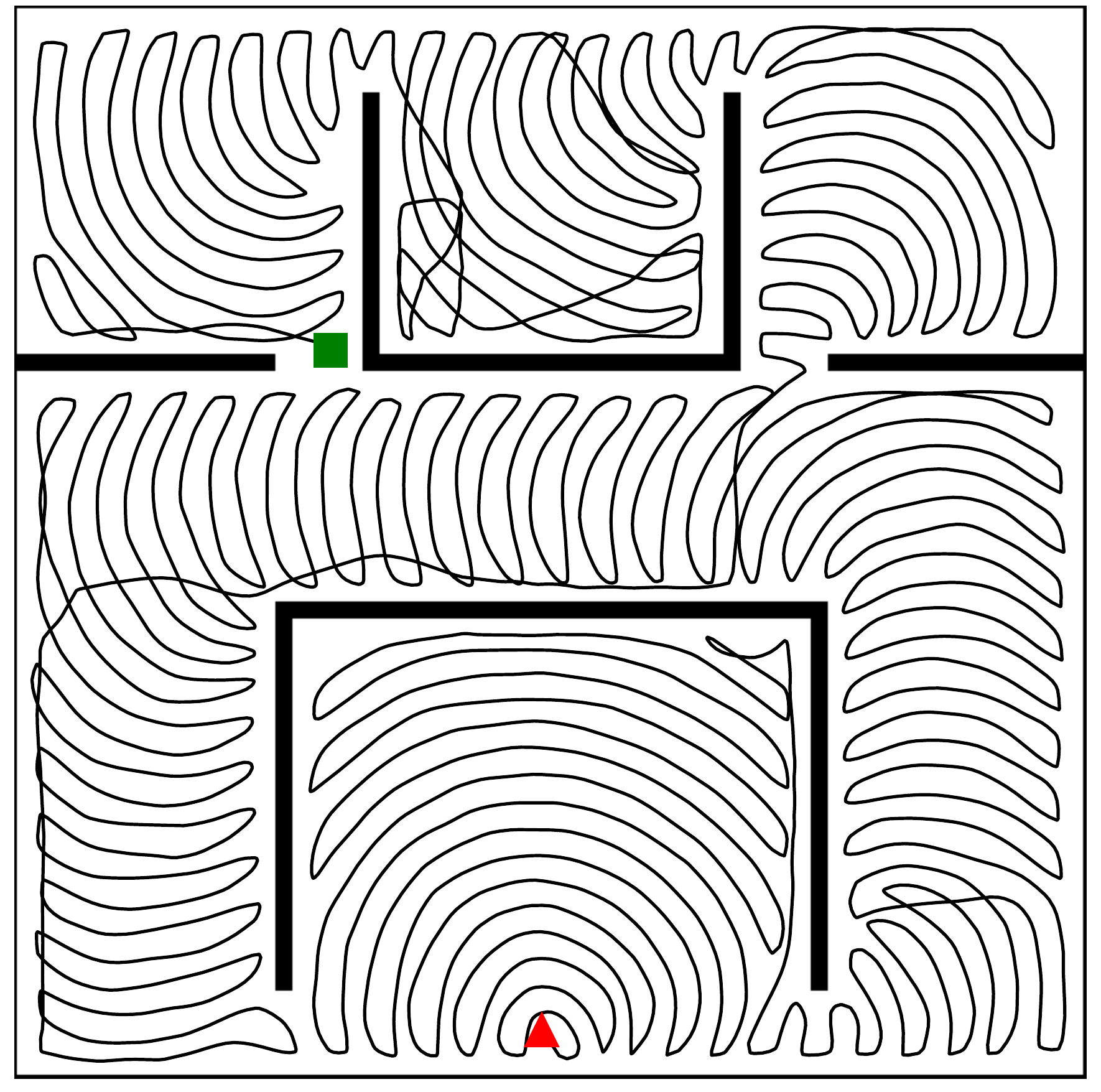} &
        \includegraphics[width=.18\linewidth]{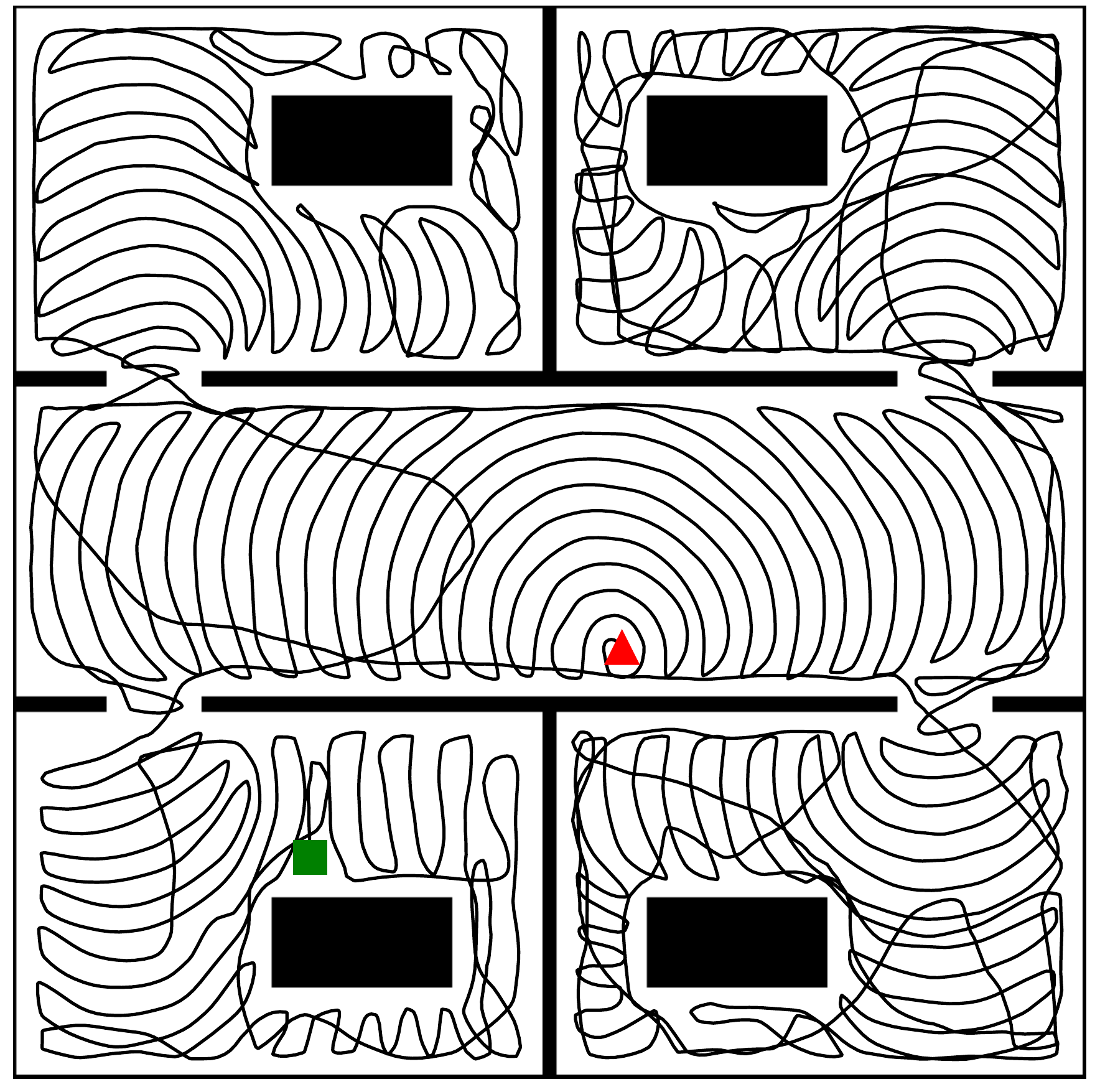} &
        \includegraphics[width=.18\linewidth]{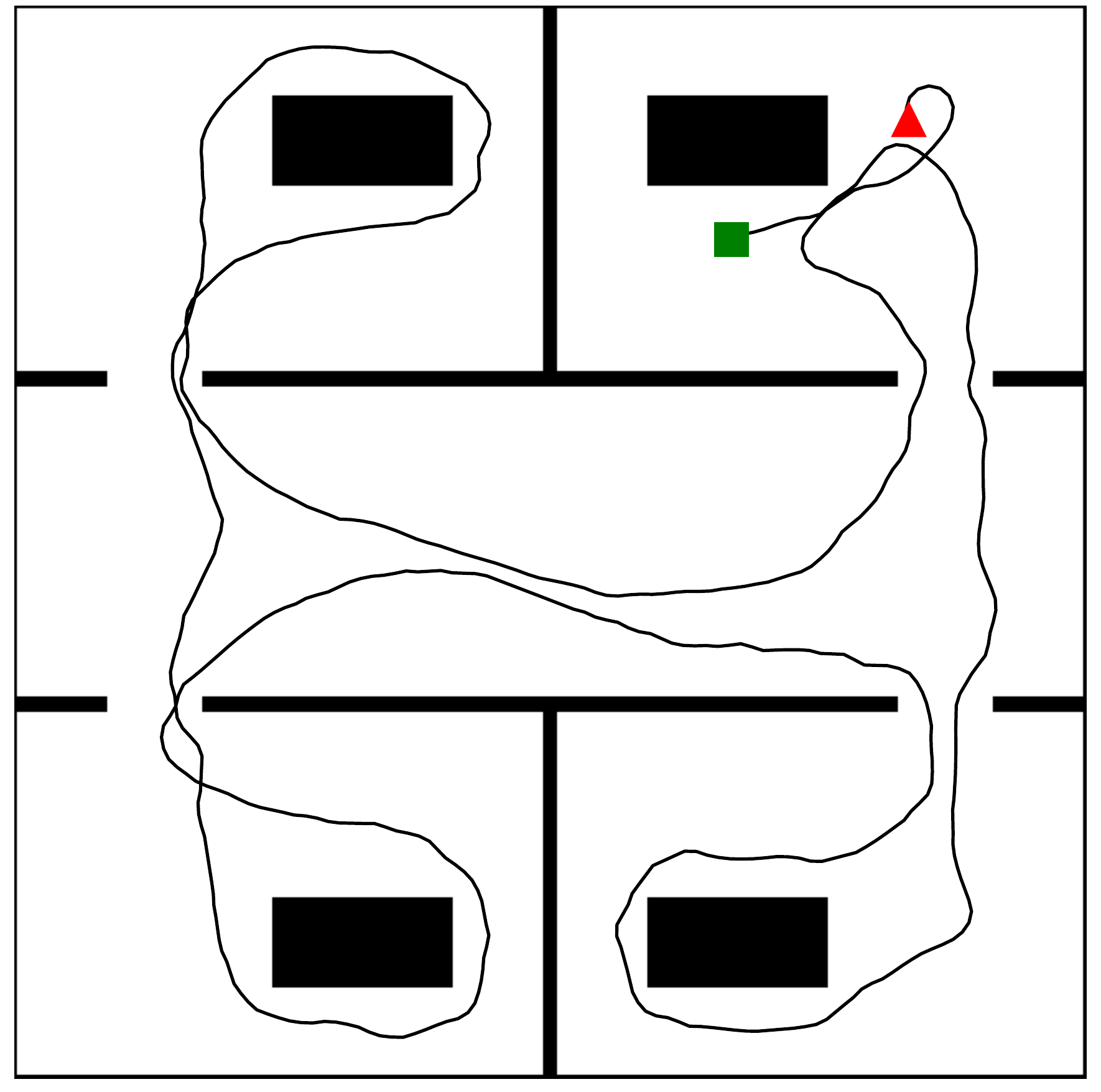} &
        \includegraphics[width=.18\linewidth]{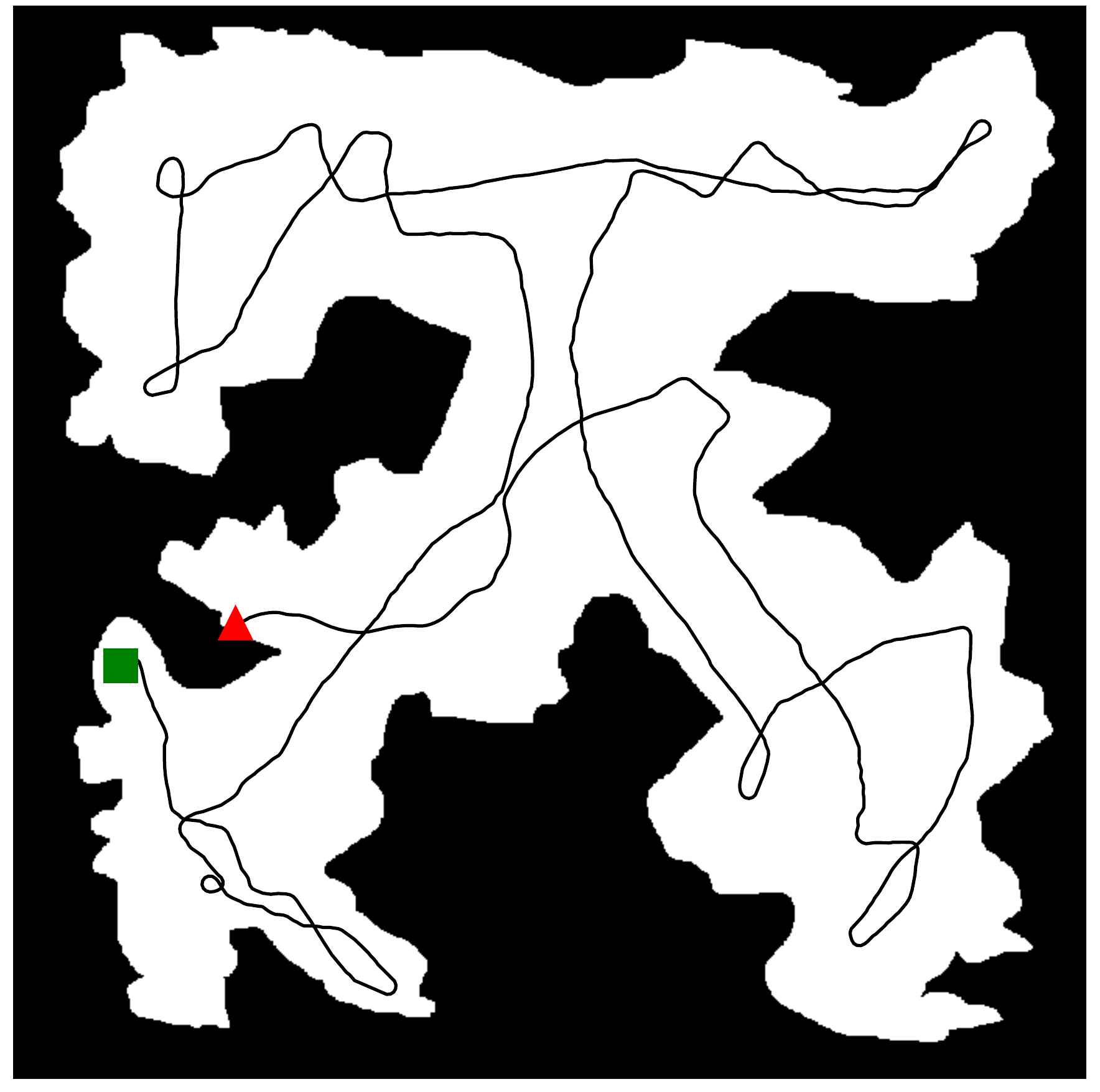} \\
        (a) & (b) & (c) & (d) & (e) \\
    \end{tabular}
    \vspace{-6pt}
    \caption{Learned paths for a fully trained agent on the lawn mowing task (a-c) and exploration (d-e).}
    \label{fig_more_qualitative}
\end{figure*}

\subsection{Ablation Study}
\label{sec_ablation_study}

In Table \ref{table_ablation}, we explore the impact of different components of our approach via a series of ablations. Since some baselines struggled to reach $90\%$ or $99\%$ coverage, we provide the coverage at fixed times, instead of $T_{90}$ and $T_{99}$.

\textbf{TV rewards.} We find that the incremental TV reward has a major impact in the lawn mowing problem, increasing the coverage from $85.1\%$ to $97.8\%$ (\ding{174} vs.\ \ding{177}). It also affects exploration, but not to the same extent. This is reasonable, as the total variation in the coverage map is lower in this case. Moreover, as global TV has less temporal variation, and behaves more like a constant reward, it turns out not to be beneficial for CPP (\ding{176} vs.\ \ding{177}).

\textbf{Agent architectures.} The higher model capacity of the CNN architecture enables a better understanding of the environment, and outperforms the MLP baseline (\ding{173} vs.\ \ding{172}). Our proposed architecture (SGCNN, \ding{177}), which groups the different scales and convolves them separately, further improves the performance compared to a naive CNN, which treats all scales as the same spatial structure.

\textbf{Multi-scale frontier maps.} We find that our proposed frontier map representation is crucial for achieving a high performance with the multi-scale map approach (\ding{175} vs.\ \ding{177}). Without this input feature, too much information is lost in the coarser scales, thus hindering long-term planning.

\textbf{Number of scales.} In Figure \ref{fig_num_maps}, we compare the coverage for different number of scales on the exploration task. It shows that only using one or two scales is not sufficient, where at least three scales are required to represent a large enough environment to enable long-term planning. The discrepancy was not as large for the lawn mowing problem, as it is more local in nature with a lower coverage radius.

\subsection{Qualitative Results}

In Figure \ref{fig_paths_training_steps}, we show learned paths during different stages of the training process. Initially, the agent performs constant actions, first with a low steering input, and later with a high steering input, presumably to avoid collisions. After $100$k iterations, the agent can already cover an empty environment, but it leaves small patches that it struggles to cover efficiently. After $1$M iterations, it leaves fewer small patches, but still struggles to efficiently cover the final few percentages of the area. In the later stages, the agent perfects its abilities of not leaving any small patches uncovered, to quickly cover the final remaining regions, and to reduce overlap. Figure \ref{fig_more_qualitative} shows learned paths for a fully trained agent, both on the lawn mowing task and on exploration.

\section{Conclusions}

We present a method for online coverage path planning in unknown environments, based on a continuous end-to-end deep reinforcement learning approach. The agent implicitly solves three tasks simultaneously, namely \textit{mapping} of the environment, \textit{planning} a coverage path, and \textit{navigating} the planned path while avoiding obstacles. For a scalable solution, we propose to use multi-scale maps, which is made viable through frontier maps that preserve the existence of non-covered space in coarse scales. We propose a novel reward term based on total variation, which significantly improves coverage time when applied locally, but not globally. The result is a task- and robot-agnostic CPP method that can be applied to different applications, which has been demonstrated on the exploration and lawn mowing tasks.

\section*{Acknowledgements}

This work was partially supported by the Wallenberg AI, Autonomous Systems and Software Program (WASP), funded by the Knut and Alice Wallenberg (KAW) Foundation. The work was funded in part by the Vinnova project, human-centered autonomous regional airport, Dnr 2022-02678. The computational resources were provided by the National Academic Infrastructure for Supercomputing in Sweden (NAISS), partially funded by the Swedish Research Council through grant agreement no.\ 2022-06725, and by the Berzelius resource, provided by the KAW Foundation at the National Supercomputer Centre (NSC).

\section*{Impact Statement}

This paper presents work whose goal is to advance the field of Machine Learning. There are many potential societal consequences of our work, none which we feel must be specifically highlighted here.

\bibliography{egbib}
\bibliographystyle{icml2024}

\newpage
\onecolumn
\section*{Appendix}

This appendix contains an extended analysis in Section \ref{supp_sec_extended_analysis} and additional implementation details in Section \ref{supp_sec_implementation_details}.

\appendix

\section{Extended Analysis}
\label{supp_sec_extended_analysis}

\subsection{The Necessity of Multi-scale Maps}
\label{supp_sec_multi_scale_maps}

While the multi-scale maps can represent large regions efficiently, they are even necessary for large-scale environments. In large environments, using a single map is not feasible as the computational cost is $\mathcal{O}(n^2)$. For example, with our current setup with 4 maps that span an area of size $76.8 \times 76.8$ m$^2$, the training step time is 40 ms. With a single map spanning only $19.2 \times 19.2$ m$^2$ using the same pixel resolution as the finest scale, the training step time is 2500 ms. This would increase the total training time by roughly two orders of magnitude. Going beyond this size resulted in memory problems on a T4 16GB GPU. Thus, a multi-scale map is necessary for large-scale environments.

\subsection{Inference Time Analysis}
\label{supp_sec_inference_time}

We evaluate the inference time of our method on two different systems in Table \ref{supp_table_inference_time}. System 1 is a high-performance computing cluster, with 16-core 6226R CPUs, and T4 GPUs. However, only 4 CPU cores and one GPU were allocated for the analysis. System 2 is a laptop computer, with a 2-core i5-520M CPU, without a GPU. The times are compared between the mowing and exploration tasks, where the inference time is lower for the mowing task since a smaller local neighborhood is used for updating the global coverage map. As our network is fairly lightweight it can run in real time, i.e.\ $\geq 20$ frames per second, even on a low-performance laptop without a GPU. All map updates are performed locally, resulting in high scalability, reaching real-time performance on the cluster node, and close to real-time performance on the laptop.

\begin{table}[h]
    \setlength{\tabcolsep}{8pt}
    \begin{center}
        \caption{Inference time in milliseconds, on a high-performance computing cluster and a laptop, for the model forward pass, other components (map updates, observation creation etc.), and in total.}
        \vspace{8pt}
        \label{supp_table_inference_time}
        \begin{tabular}{lcccccc}
        \toprule
            & \multicolumn{3}{c}{\textbf{Cluster node}} & \multicolumn{3}{c}{\textbf{Laptop}} \\
            & \multicolumn{3}{c}{6226R CPU, T4 GPU} & \multicolumn{3}{c}{i5-520M CPU, No GPU} \\
            \cmidrule(l{\tabcolsep}r{\tabcolsep}){2-4}
            \cmidrule(l{\tabcolsep}r{\tabcolsep}){5-7}
            Task & model & other & total & model & other & total \\
            \midrule
            Exploration & 2 & 19 & 21 & 5 & 57 & 62 \\
            Mowing & 2 & 3 & 5 & 5 & 9 & 14 \\
            \bottomrule
        \end{tabular}
    \end{center}
\end{table}

\subsection{Collision Frequency}
\label{supp_sec_collision}

In most real-world applications, it is vital to minimize the collision frequency, as collisions can inflict harm on humans, animals, the environment, and the robot. To gain insights into the collision characteristics of our trained agent, we tracked the collision frequency during evaluation. For exploration, it varied between once every 100-1000 seconds, which is roughly once every 50-500 meters. For lawn mowing, it varied between once every 150-250 seconds, which is roughly once every 30-50 meters. These include all forms of collisions, even low-speed side collisions. The vast majority of collisions were near-parallel, and we did not observe any head on collisions. The practical implications are very different between these cases.

\subsection{Impact of the TV Reward on Learned Paths}
\label{supp_sec_tv_impact}

\begin{figure}[h]
    \centering
    \setlength{\tabcolsep}{5pt}
    \begin{tabular}{cccccccc}
        \includegraphics[height=.35\linewidth]{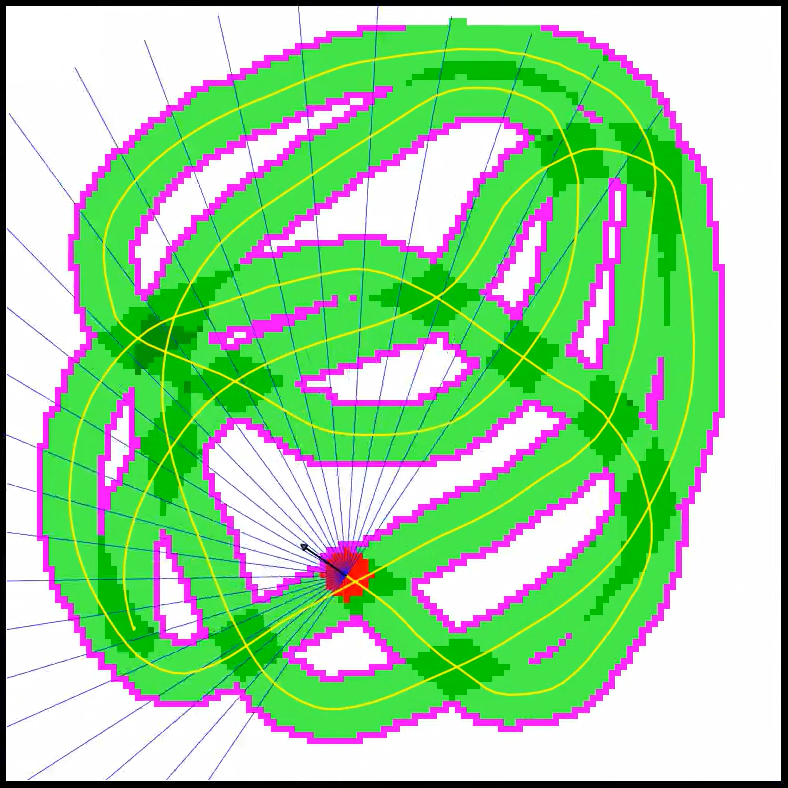} &
        \includegraphics[height=.35\linewidth]{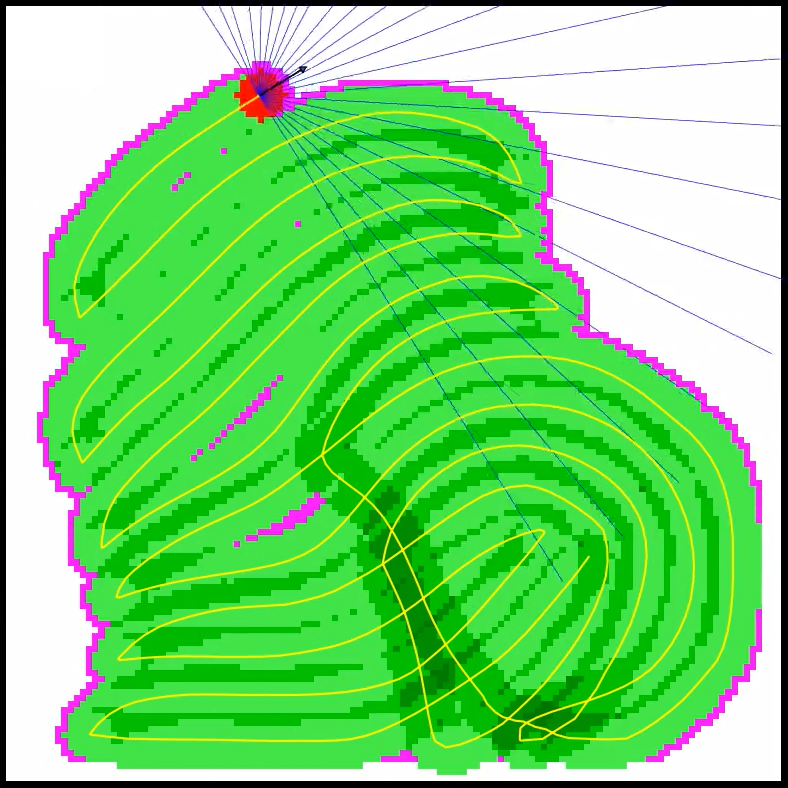} \\
    \end{tabular}
    \vspace{-8pt}
    \caption{The total variation reward term improves path quality. The figure shows the path (yellow) without TV reward (left), and with TV reward (right), including the covered region (green), lidar rays (blue lines) and frontier points (magenta).}
    \label{supp_fig_tv_paths}
\end{figure}

Figure \ref{supp_fig_tv_paths} shows learned paths both with and without the incremental TV reward. The TV reward reduces small leftover regions which are costly to cover later on.

\subsection{Additional Rewards that Did Not Work}
\label{supp_sec_additional_rewards}

In addition to the reward terms described in Section \ref{sec_reward_function}, we experimented with a goal coverage reward, and an episode truncation reward. Both were added at the end of every episode. The goal coverage reward was a positive reward that was given once the agent reached the predefined goal coverage. The truncation reward was a negative reward that was given in case the episode was prematurely truncated due to the agent not covering new ground in $\tau$ consecutive steps. However, neither of these rewards affected the end result, likely due to the fact that they are sparse compared to the other rewards, and do not provide a strong reward signal. The coverage reward already encourages complete coverage, and the constant negative reward heavily penalizes not reaching full coverage. Since both of these are dense, adding sparse rewards for the same purpose does not provide any benefit.

\subsection{Generalization of the Lidar Feature Representation}
\label{supp_sec_normalizing_distance}

Since we represent the lidar distance measurements as a vector of a fixed size and normalize the distances to $[0, 1]$, the sensor observation is tied to a specific sensor setup. Due to the normalization, the sensor observation is tied to a specific sensor range. This is important to keep in mind when transferring to another sensor setup. If the maximum range differs, the normalization constant needs to be considered, such that the same distance corresponds to the same observed value, and the maximum observable value should be limited to 1. If the number of lidar rays differ, the representation might not generalize as easily, as the dimensionality of the sensor vector would change. One option is to interpolate the observed distances at the ray angles that were used during training. Although, inaccuracies may occur if the angles are not well aligned. However, the map representation is more flexible, and does in theory also contain the information of the raw lidar data. So excluding the sensor feature extractor would in theory generalize to any sensor setup.

\section{Additional Implementation Details}
\label{supp_sec_implementation_details}

\subsection{Physical Dimensions}
\label{supp_sec_physical_dimensions}

Table \ref{supp_table_physical_dimensions} shows the physical dimensions in the three different variations of the CPP problem that were considered. We match the dimensions of the methods under comparison to get comparable performance values.

\begin{table}[h]
    \begin{center}
        \caption{Physical dimensions for the three different CPP variations.}
        \vspace{8pt}
        \label{supp_table_physical_dimensions}
        \begin{tabular}{lccc}
            \toprule
            & Omnidirectional & Non-omnidirectional & \multirow{2}{*}{Lawn mowing} \\
            & exploration & exploration & \\
            \midrule
            Coverage radius & $7$ m & $3.5$ m & $0.15$ m \\
            Agent radius & $0.08$ m & $0.15$ m & $0.15$ m \\
            Maximum linear velocity & $0.5$ m/s & $0.26$ m/s & $0.26$ m/s \\
            Maximum angular velocity & $1$ rad/s & $1$ rad/s & $1$ rad/s \\
            Simulation step size & $0.5$ s & $0.5$ s & $0.5$ s \\
            Lidar rays & $20$ & $24$ & $24$ \\
            Lidar range & $7$ m & $3.5$ m & $3.5$ m \\
            Lidar field-of-view & $360^\circ$ & $180^\circ$ & $180^\circ$ \\
            \bottomrule
        \end{tabular}
    \end{center}
\end{table}

\subsection{Agent Architecture Details}
\label{supp_sec_agent_architecture}

As proposed for soft actor-critic (SAC) \citep{haarnoja2018soft_2}, we use two Q-functions that are trained independently, together with accompanying target Q-functions whose weights are exponentially moving averages. No separate state-value network is used. All networks, including actor network, Q-functions, and target Q-functions share the same network architecture, which is either a multilayer perceptron (MLP), a naive convolutional neural network (CNN), or our proposed scale-grouped convolutional neural network (SGCNN). These are described in the following paragraphs. The MLP contains a total of $3.2$M parameters, where most are part of the initial fully connected layer, which takes the entire observation vector as input. Meanwhile, the CNN-based architectures only contain $0.8$M parameters, as the convolution layers process the maps with fewer parameters.

\textbf{MLP.} As mentioned in the main paper, the map and sensor feature extractors for MLP simply consist of flattening layers that restructure the inputs into vectors. The flattened multi-scale coverage maps, obstacle maps, frontier maps, and lidar detections are concatenated and fed to the fusion module. For the Q-functions, the predicted action is appended to the input of the fusion module. The fusion module consists of two fully connected layers with ReLU and 256 units each. A final linear prediction head is appended, which predicts the mean and standard deviation for sampling the action in the actor network, or the Q-value in the Q-functions.

\textbf{Naive CNN.} This architecture is identical to SGCNN, see below, except that it convolves all maps simultaneously in one go, without utilizing grouped convolutions.

\textbf{SGCNN.} Our proposed SGCNN architecture uses the same fusion module as MLP, but uses additional layers for the feature extractors, including convolutional layers for the image-like maps. Due to the simple nature of the distance readings for the lidar sensor, the lidar feature extractor only uses a single fully connected layer with ReLU. It has the same number of hidden neurons as there are lidar rays. The map feature extractor consists of a $2 \times 2$ convolution with stride $2$ to reduce the spatial resolution, followed by three $3 \times 3$ convolutions with stride 1 and without padding. We use $24$ total channels in the convolution layers. Note that the maps are grouped by scale, and convolved separately by their own set of filters. The map features are flattened into a vector and fed to a fully connected layer of $256$ units, which is the final output of the map feature extractor. ReLU is used as the activation function throughout.

\subsection{Choice of Hyperparameters}
\label{supp_sec_hyperparameters}

\textbf{Multi-scale map parameters.} We started by considering the size and resolution in the finest scale, such that sufficiently small details could be represented in the nearest vicinity. We concluded that a resolution of $0.0375$ meters per pixel was a good choice as it corresponds to $8$ pixels for a robot with a diameter of $30$ cm. Next, we chose a size of $32 \times 32$ pixels, as this results in the finest scale spanning $1.2 \times 1.2$ meters. Next, we chose the scale factor $s = 4$, as this allows only a few maps to represent a relatively large region, while maintaining sufficient detail in the second finest scale. Finally, the training needs to be done with a fixed number of scales. The most practical way is to simply train with sufficiently many scales to cover the maximum size for a particular use case. Thus, we chose $m = 4$ scales, which in total spans a $76.8 \times 76.8$ m region, and can contain any of the considered training and evaluation environments. If the model is deployed in a small area, the larger scales would not contain any frontier points in the far distance, but the agent can still cover the area by utilizing the smaller scales. For a larger use case, increasing the size of the represented area by adding more scales is fairly cheap, as the computational cost is $\mathcal{O}(\log n)$.

\textbf{Reward parameters.} Due to the normalization of the area and TV rewards, they are given approximately the same weight for $\lambda_\mathrm{area} = 1$ and  $\lambda_\mathrm{TV} = 1$. Thus, we used this as a starting point, which seemed to work well for the lawn mowing problem, while a lower weight for the TV rewards was advantageous for exploration.

\textbf{Episode truncation.} For large environments, we found it important not to truncate the episodes too early, as this would hinder learning. If the truncation parameter $\tau$ was set too low, the agent would not be forced to learn to cover an area completely, as the episode would simply truncate whenever the agent could not progress, without a large penalty. With the chosen value of $\tau = 1000$, the agent would be greatly penalized by the constant reward $R_\mathrm{const}$ for not reaching the goal coverage, and would be forced to learn to cover the complete area in order to maximize the reward.

\subsection{Random Map Generation}
\label{supp_sec_random_map_generation}

Inspired by procedural environment generation for reinforcement learning \citep{justesen2018illuminating}, we use randomly generated maps during training to increase the variation in the training data and to improve generalization. We consider random floor plans and randomly placed circular obstacles. First, the side length of a square area is chosen uniformly at random, from the interval $[2.4, 7.5]$ meters for mowing and $[9.6, 15]$ meters for exploration. Subsequently, a random floor plan is created with a $70\%$ probability. Finally, with $70\%$ probability, a number of obstacles are placed such that they are far enough apart to guarantee that the agent can visit the entire free space, i.e.\ that they are not blocking the path between different parts of the free space. An empty map can be generated if neither a floor plan is created nor obstacles placed. In the following paragraphs, we describe the floor plan generation and obstacle placement in more detail.

\textbf{Random floor plans.} The random floor plans contain square rooms of equal size in a grid-like configuration, where neighboring rooms can be accessed through door openings. On occasion, a wall is removed completely or some openings are closed off to increase the variation further. First, floor plan parameters are chosen uniformly at random, such as the side length of the rooms from $[1.5, 4.8]$ meters, wall thickness from $[0.075, 0.3]$ meters, and door opening size from $[0.6, 1.2]$ meters. Subsequently, each vertical and horizontal wall spanning the whole map either top-to-bottom or left-to-right is placed with a $90\%$ probability. After that, door openings are created at random positions between each neighboring room. Finally, one opening is closed off at random for either each top-to-bottom spanning vertical wall or each left-to-right spanning horizontal wall, not both. This is to ensure that each part of the free space can be reached by the agent.

\textbf{Random circular obstacles.} Circular obstacles with radius $0.25$ meters are randomly scattered across the map, where one obstacle is placed for every four square meters. If the closest distance between a new obstacle and any previous obstacle or wall is less than $0.6$ meters, the new obstacle is removed to ensure that large enough gaps are left for the agent to navigate through and that it can reach every part of the free space.

\subsection{Progressive Training}
\label{supp_sec_progressive_training}

\begin{figure}[h]
    \setlength{\tabcolsep}{0.5pt}
    \setlength{\fboxsep}{0pt}%
    \setlength{\fboxrule}{0.5pt}%
    \begin{tabular}{p{1cm}cccccccc}
        \makecell{\scriptsize Tier 0 \\[2pt] \scriptsize \textcolor{white}{line}} &
        \fbox{\includegraphics[width=.09\linewidth]{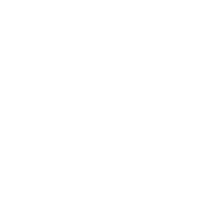}} &
        \fbox{\includegraphics[width=.0745\linewidth]{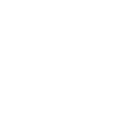}}
    \end{tabular} \\
    \begin{tabular}{p{1cm}cccccccc}
        \makecell{\scriptsize Tier 1 \\[2pt] \scriptsize \textcolor{white}{line}} &
        \fbox{\includegraphics[width=.0745\linewidth]{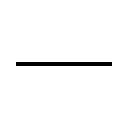}} &
        \fbox{\includegraphics[width=.0745\linewidth]{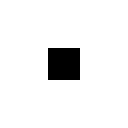}} &
        \fbox{\includegraphics[width=.0745\linewidth]{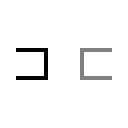}} &
        \fbox{\includegraphics[width=.0745\linewidth]{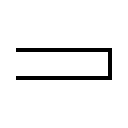}}
    \end{tabular} \\
    \begin{tabular}{p{1cm}cccccccc}
        \makecell{\scriptsize Tier 2 \\[2pt] \scriptsize \textcolor{white}{line}} &
        \fbox{\includegraphics[width=.1\linewidth]{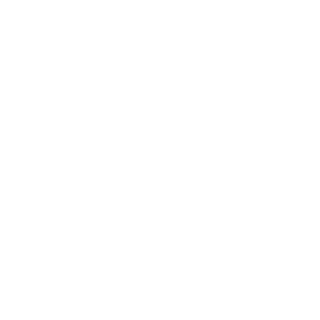}} &
        \fbox{\includegraphics[width=.09\linewidth]{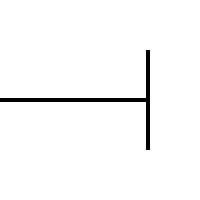}} &
        \fbox{\includegraphics[width=.09\linewidth]{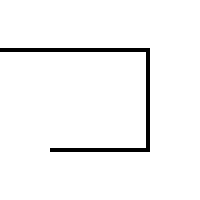}} &
        \fbox{\includegraphics[width=.09\linewidth]{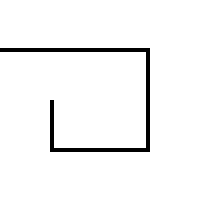}} &
        \fbox{\includegraphics[width=.09\linewidth]{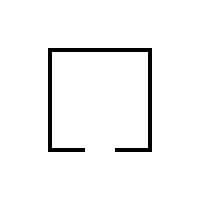}} &
        \fbox{\includegraphics[width=.09\linewidth]{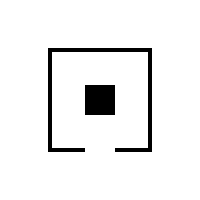}} &
        \fbox{\includegraphics[width=.09\linewidth]{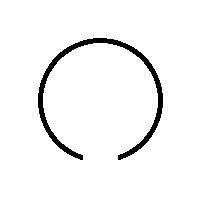}} &
        \fbox{\includegraphics[width=.09\linewidth]{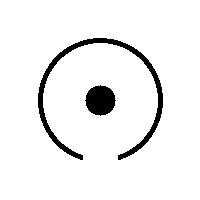}}
    \end{tabular} \\
    \begin{tabular}{p{1cm}cccccccc}
        \makecell{\scriptsize Tier 3 \\[8pt] \scriptsize \textcolor{white}{line}} &
        \fbox{\includegraphics[width=.1\linewidth]{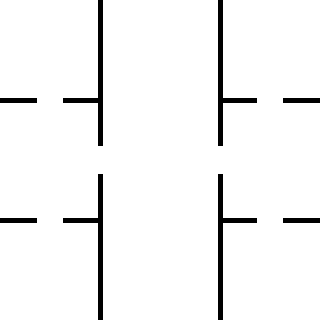}} &
        \fbox{\includegraphics[width=.1\linewidth]{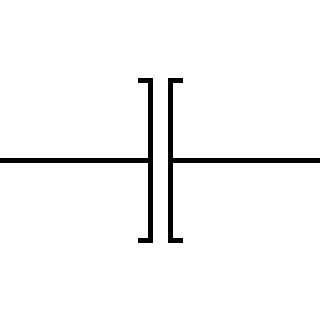}} &
        \fbox{\includegraphics[width=.1\linewidth]{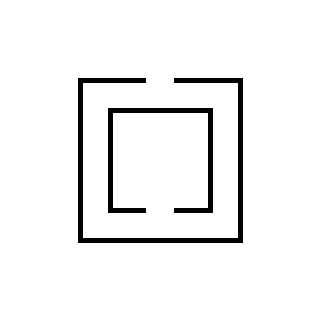}} &
        \fbox{\includegraphics[width=.1\linewidth]{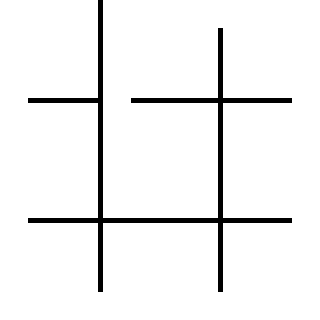}} &
        \fbox{\includegraphics[width=.1\linewidth]{figures/maps/train_3_5.png}} &
        \fbox{\includegraphics[width=.1\linewidth]{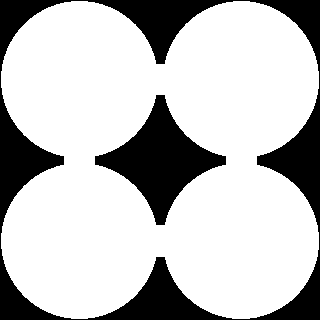}}
    \end{tabular} \\
    \begin{tabular}{p{1cm}cccccccc}
        \makecell{\scriptsize Tier 4 \\[14pt] \scriptsize \textcolor{white}{line}} &
        \fbox{\includegraphics[width=.14\linewidth]{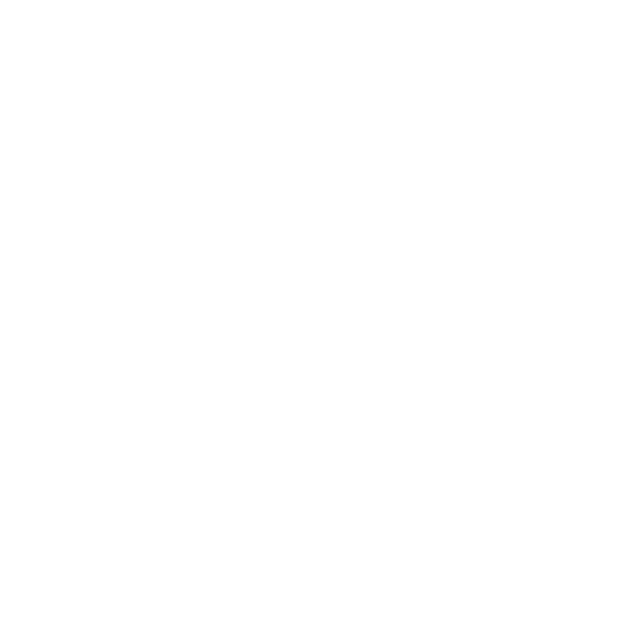}} &
        \fbox{\includegraphics[width=.12\linewidth]{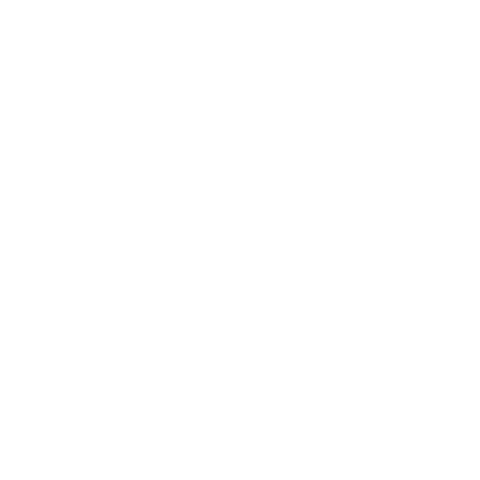}} &
        \fbox{\includegraphics[width=.12\linewidth]{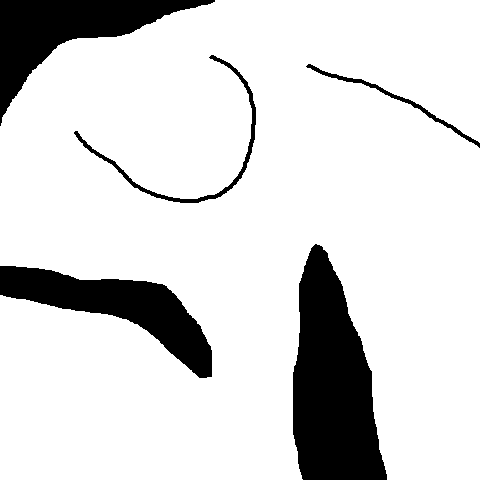}} &
        \fbox{\includegraphics[width=.12\linewidth]{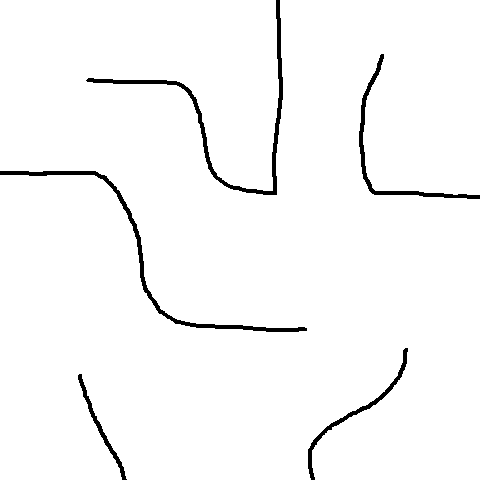}} \\[-12pt]
        &
        \fbox{\includegraphics[width=.12\linewidth]{figures/maps/train_4_3.png}} &
        \fbox{\includegraphics[width=.12\linewidth]{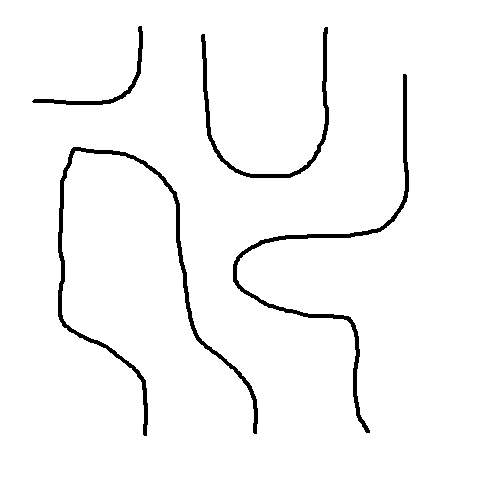}} &
        \fbox{\includegraphics[width=.12\linewidth]{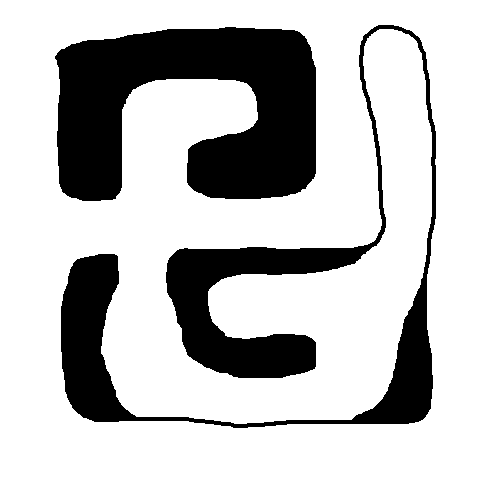}} &
        \fbox{\includegraphics[width=.12\linewidth]{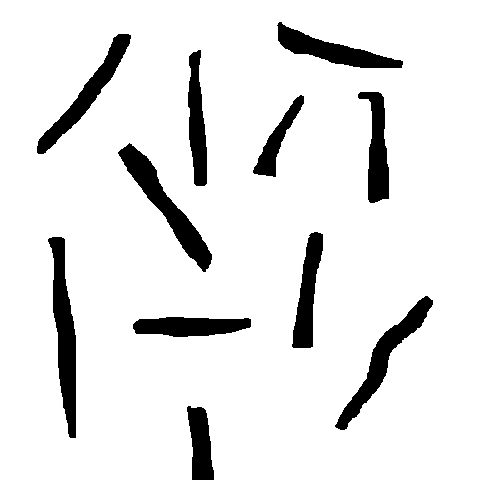}}
    \end{tabular} \\[12pt]
    \begin{tabular}{p{1cm}cccccccc}
        \makecell{\scriptsize Tier 5 \\[14pt] \scriptsize \textcolor{white}{line}} &
        \fbox{\includegraphics[width=.14\linewidth]{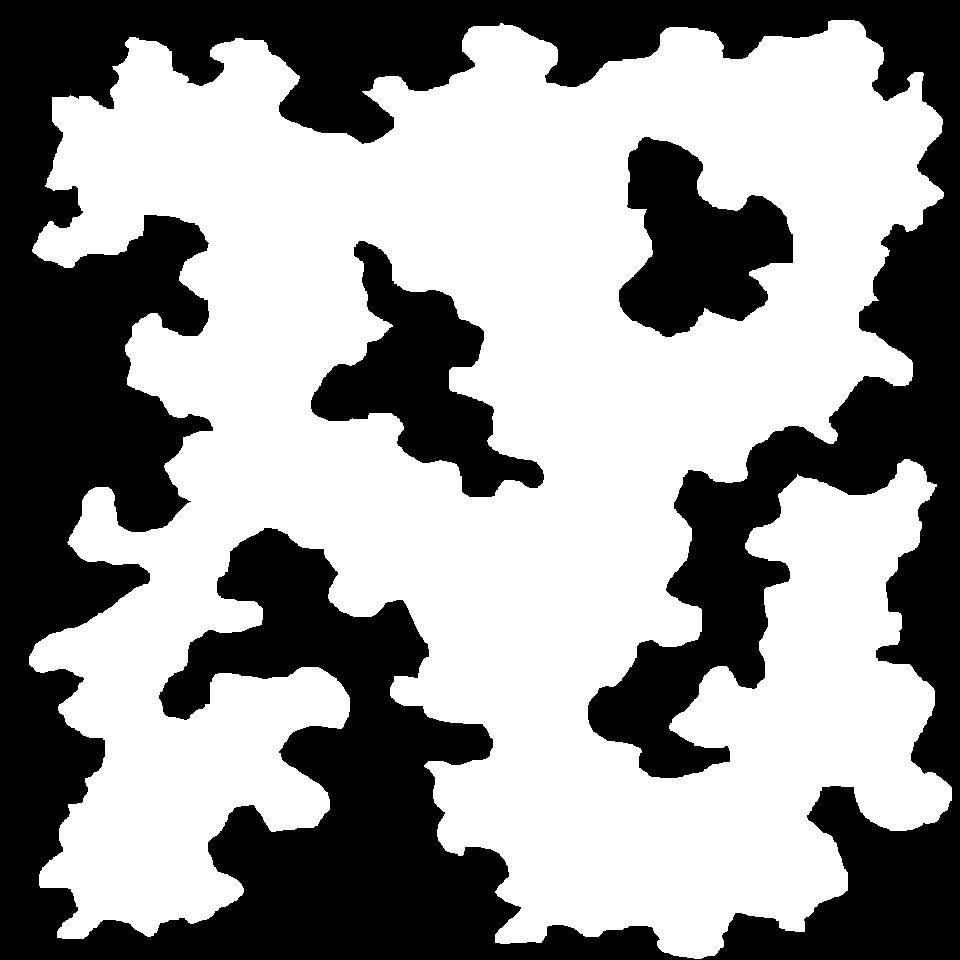}} &
        \fbox{\includegraphics[width=.14\linewidth]{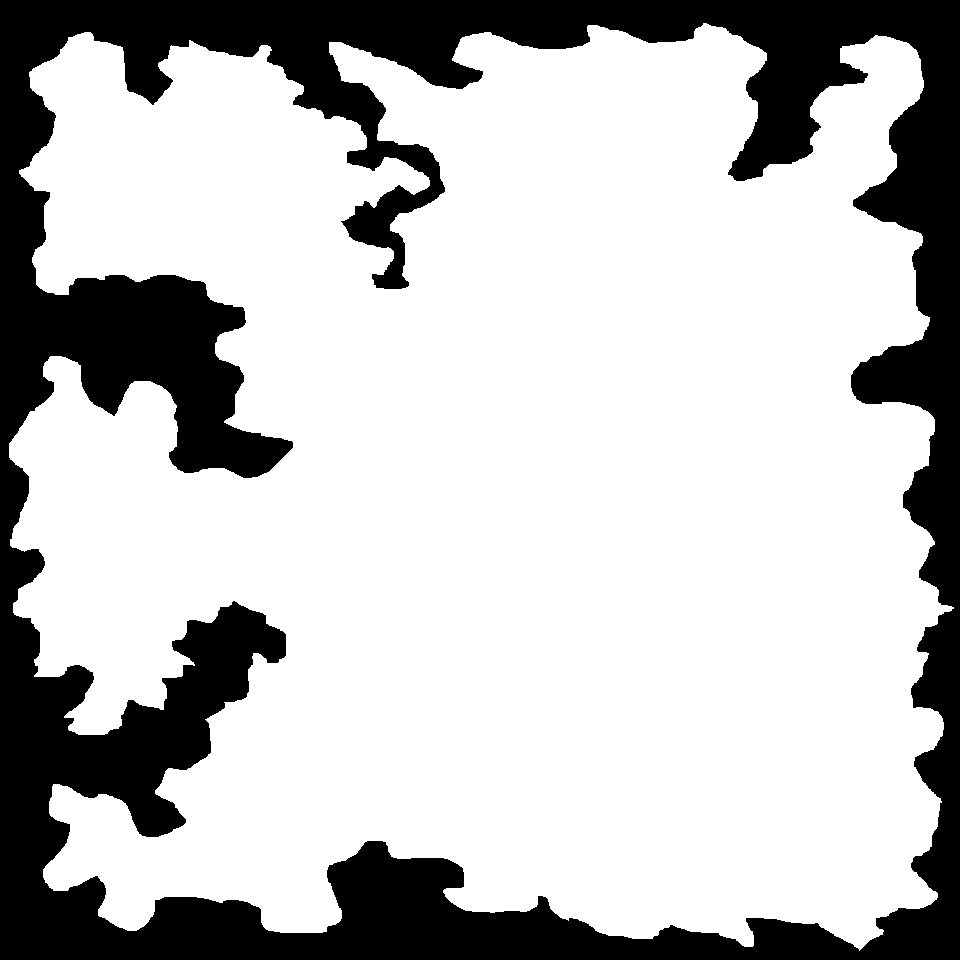}} &
        \fbox{\includegraphics[width=.14\linewidth]{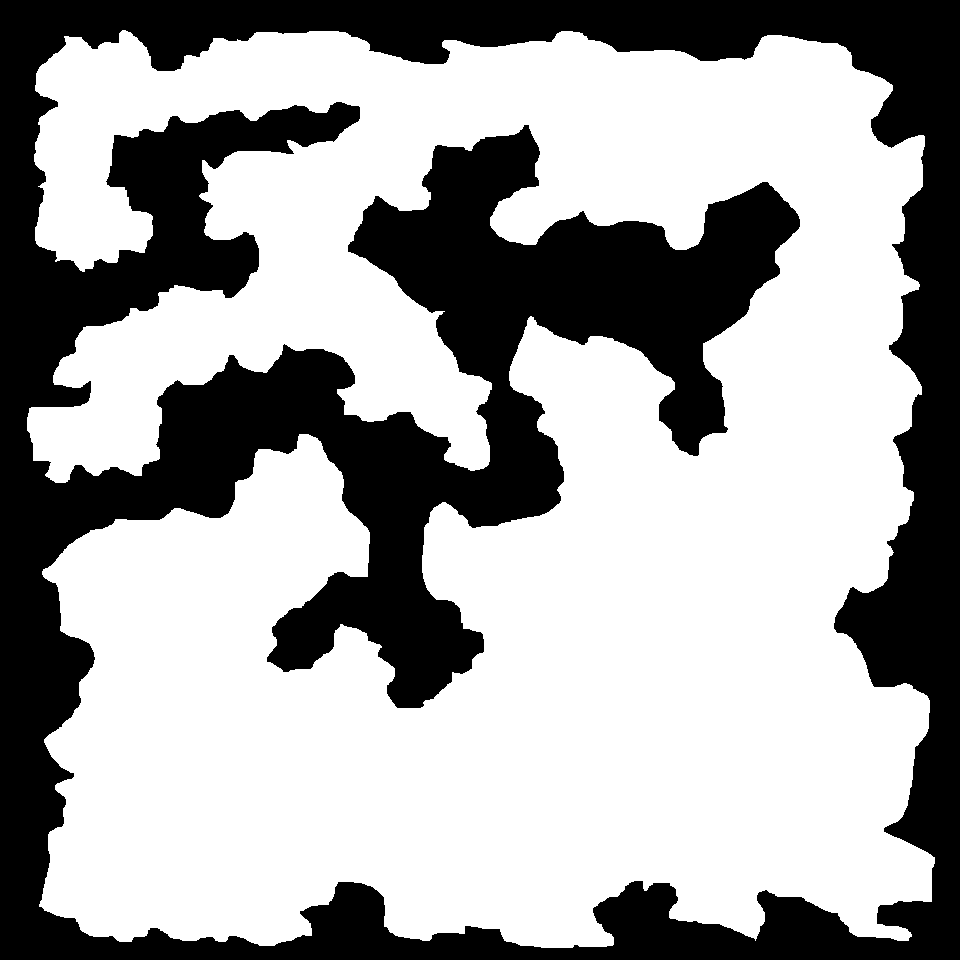}} &
        \fbox{\includegraphics[width=.14\linewidth]{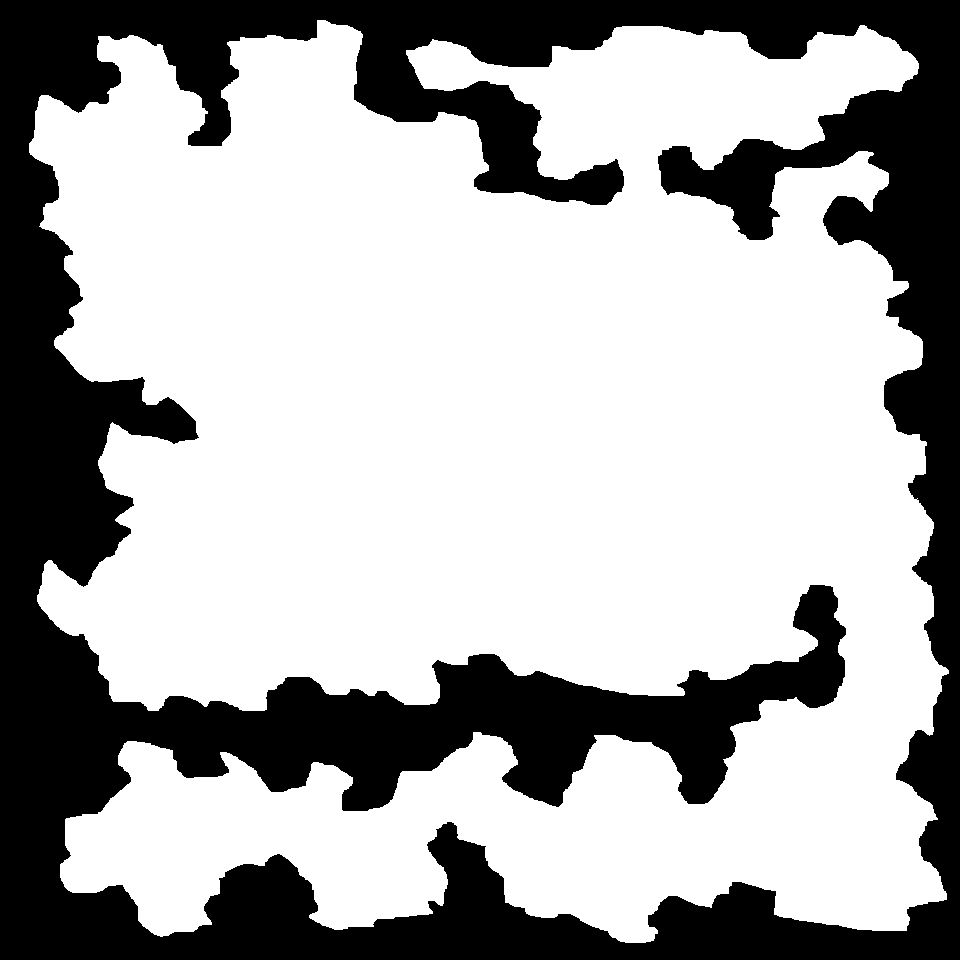}} &
        \fbox{\includegraphics[width=.14\linewidth]{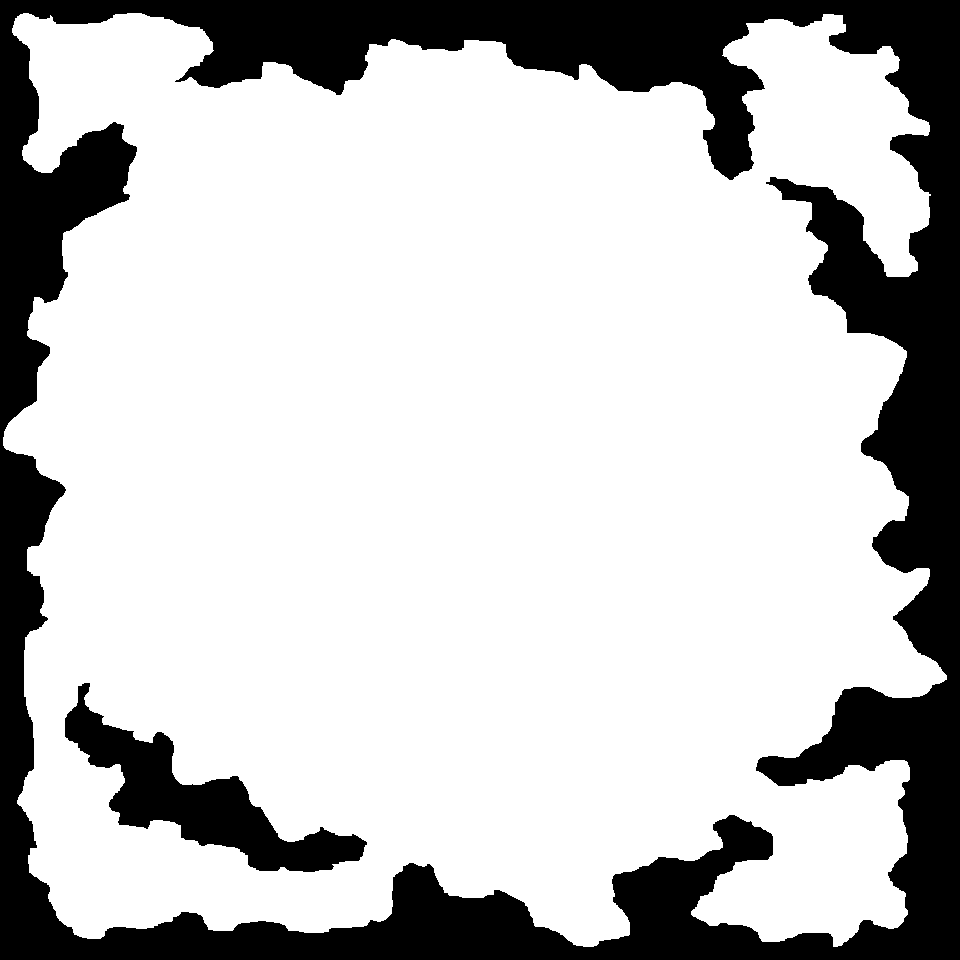}} &
        \fbox{\includegraphics[width=.14\linewidth]{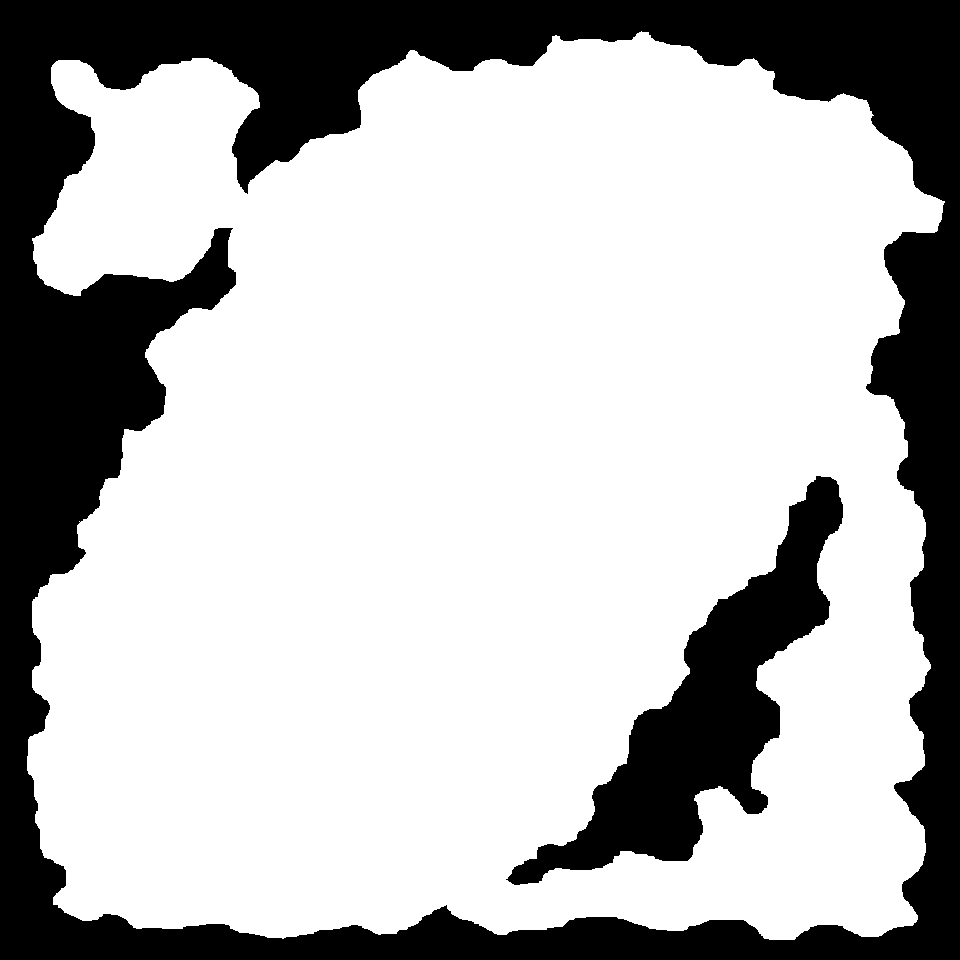}} \\[-12pt]
        &
        \fbox{\includegraphics[width=.14\linewidth]{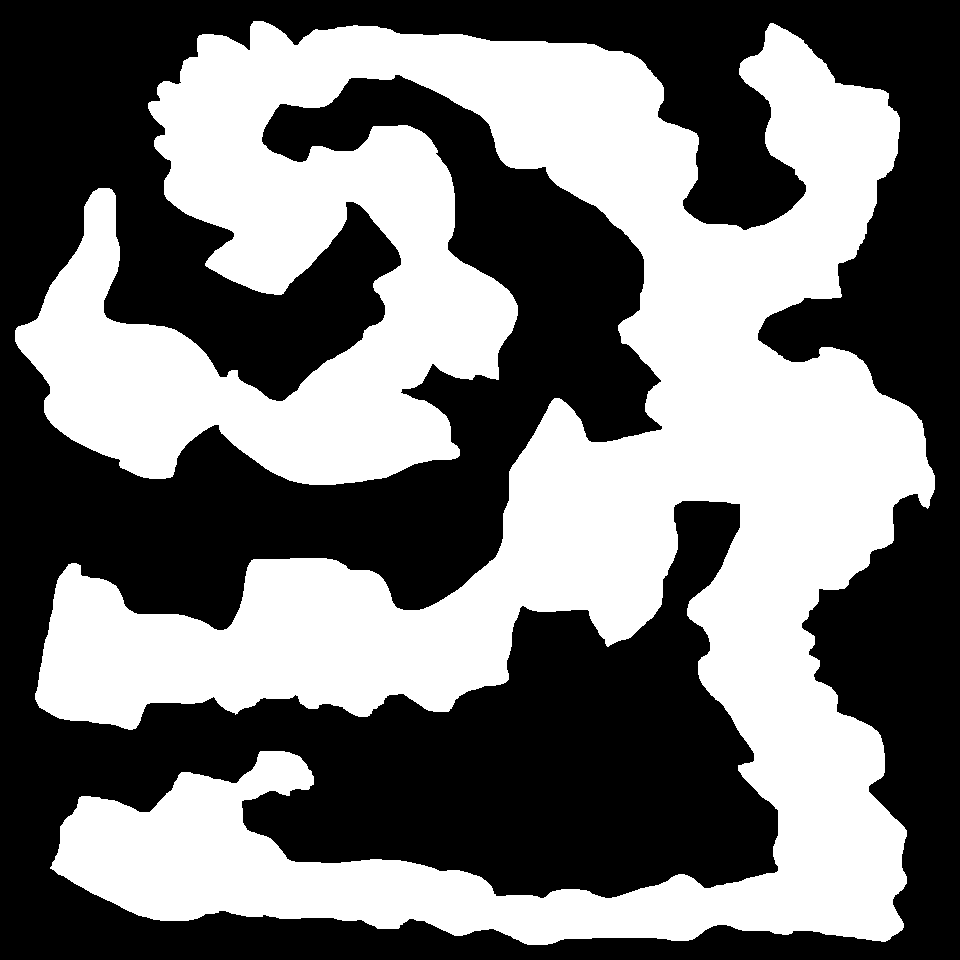}} &
        \fbox{\includegraphics[width=.14\linewidth]{figures/maps/train_5_8.png}} &
        \fbox{\includegraphics[width=.14\linewidth]{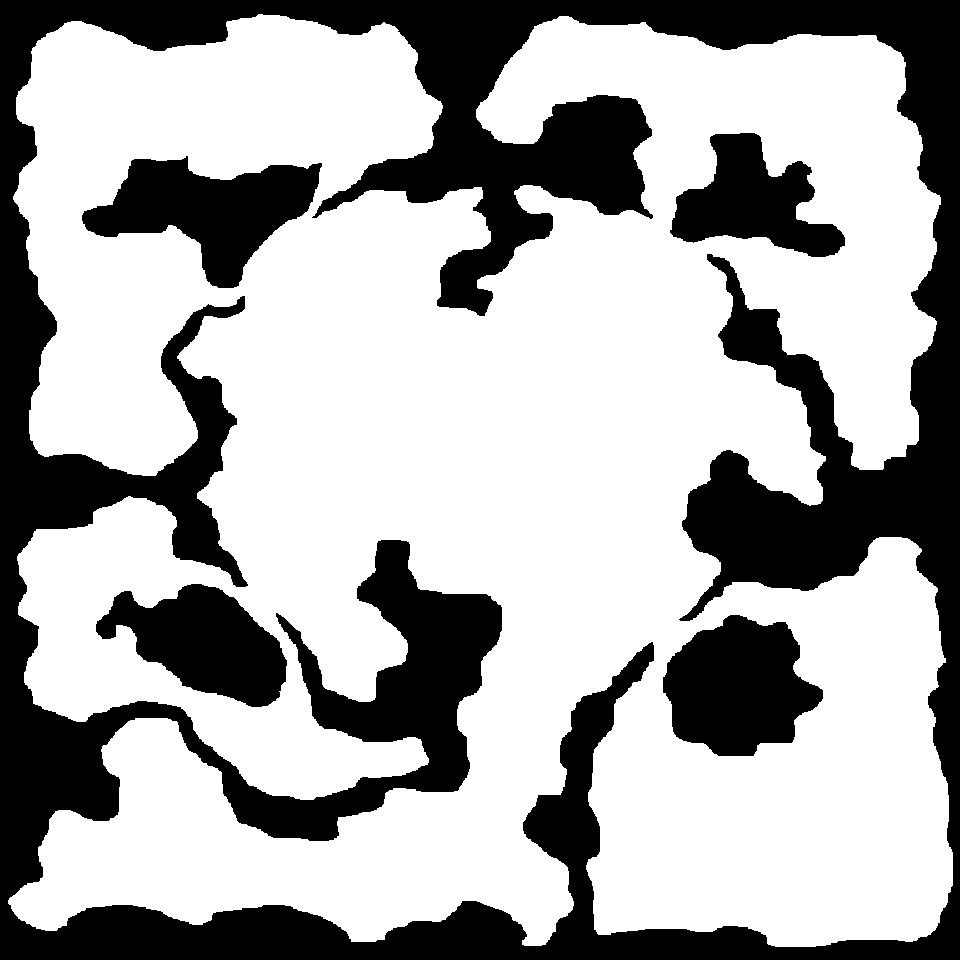}} &
        \fbox{\includegraphics[width=.14\linewidth]{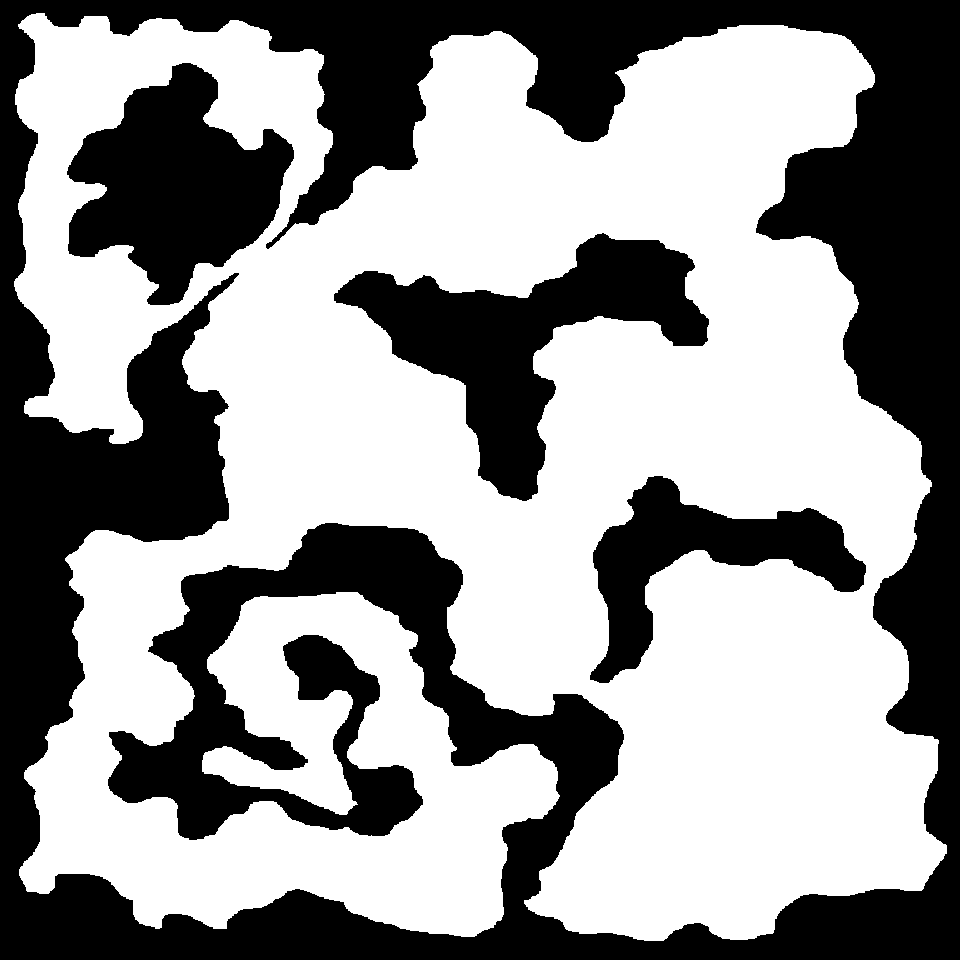}} &
        \fbox{\includegraphics[width=.14\linewidth]{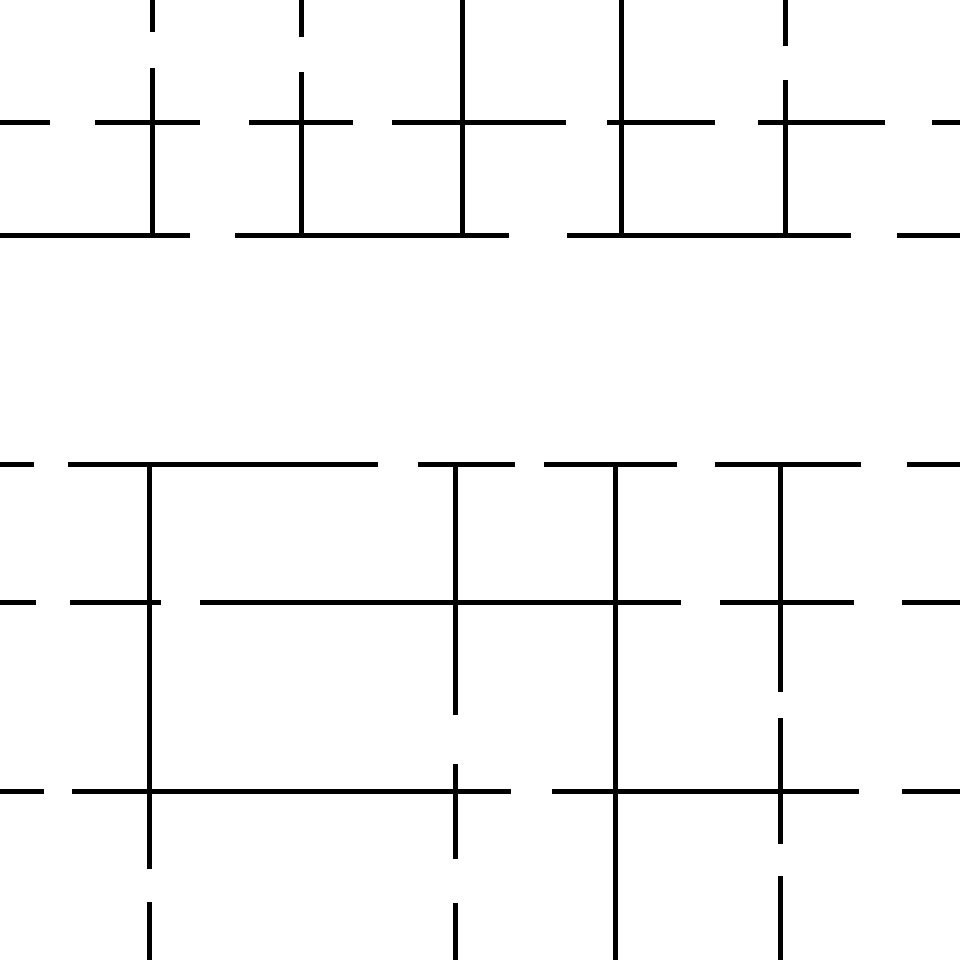}} &
        \fbox{\includegraphics[width=.14\linewidth]{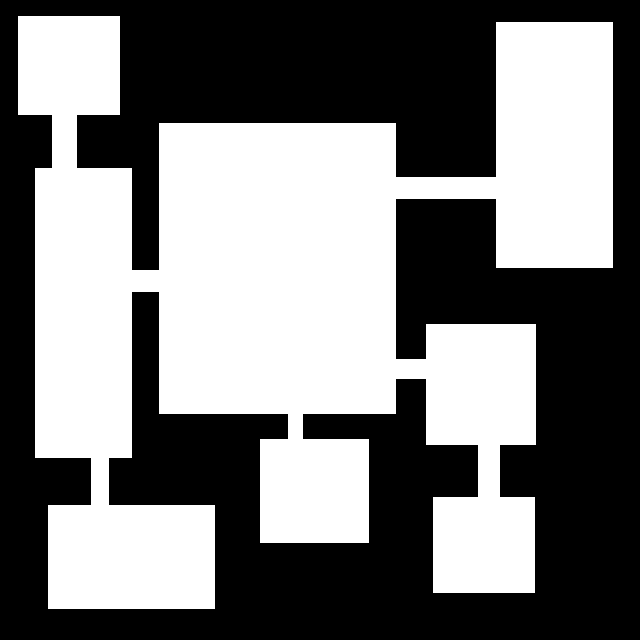}} \\
        &
        \fbox{\includegraphics[width=.14\linewidth]{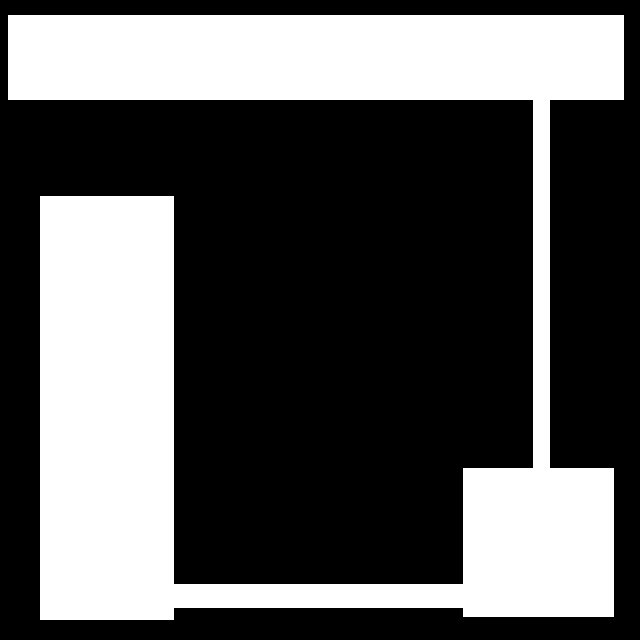}} &
        \fbox{\includegraphics[width=.14\linewidth]{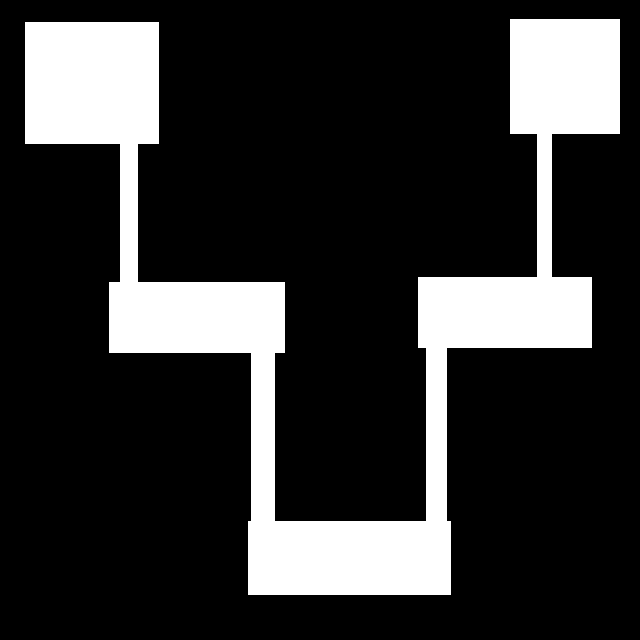}} &
        \fbox{\includegraphics[width=.14\linewidth]{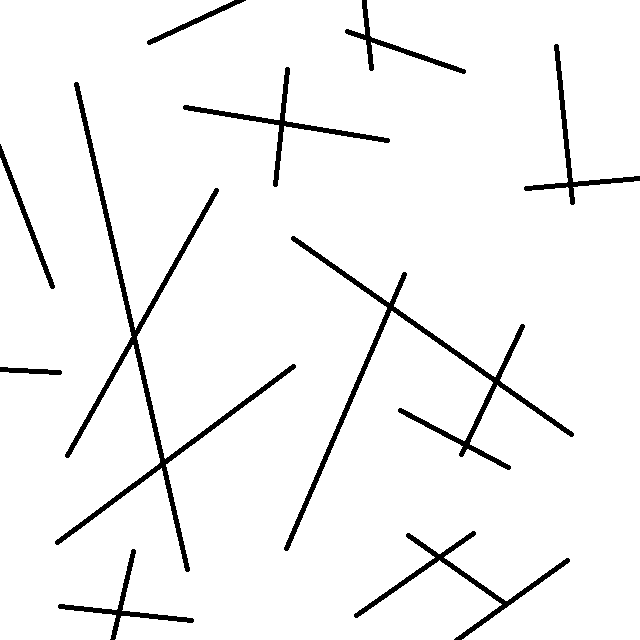}} &
        \fbox{\includegraphics[width=.14\linewidth]{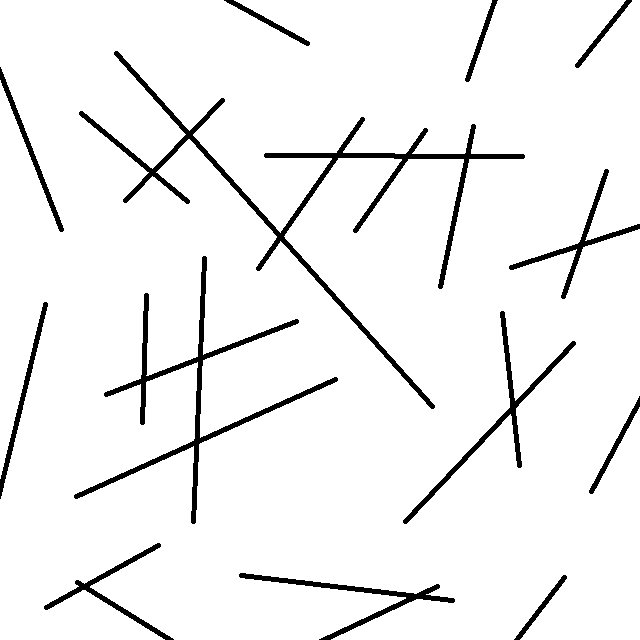}} &
        \fbox{\includegraphics[width=.14\linewidth]{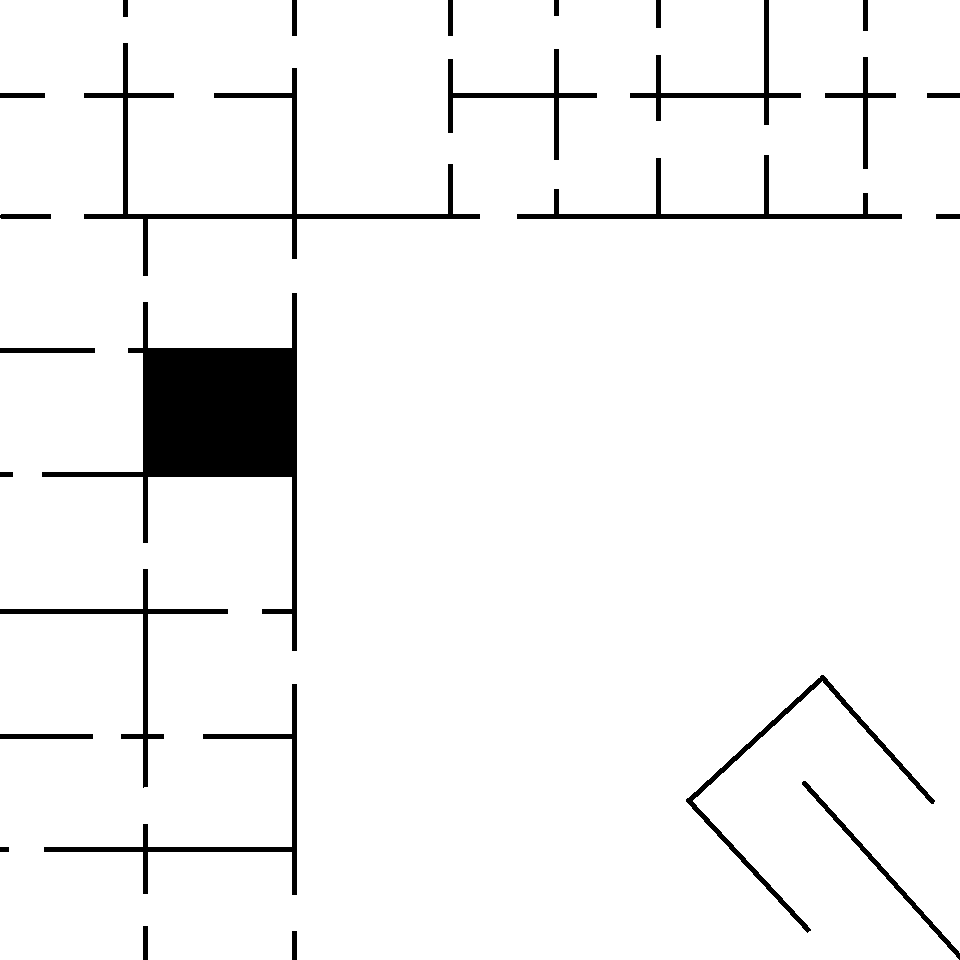}} &
        \fbox{\includegraphics[width=.14\linewidth]{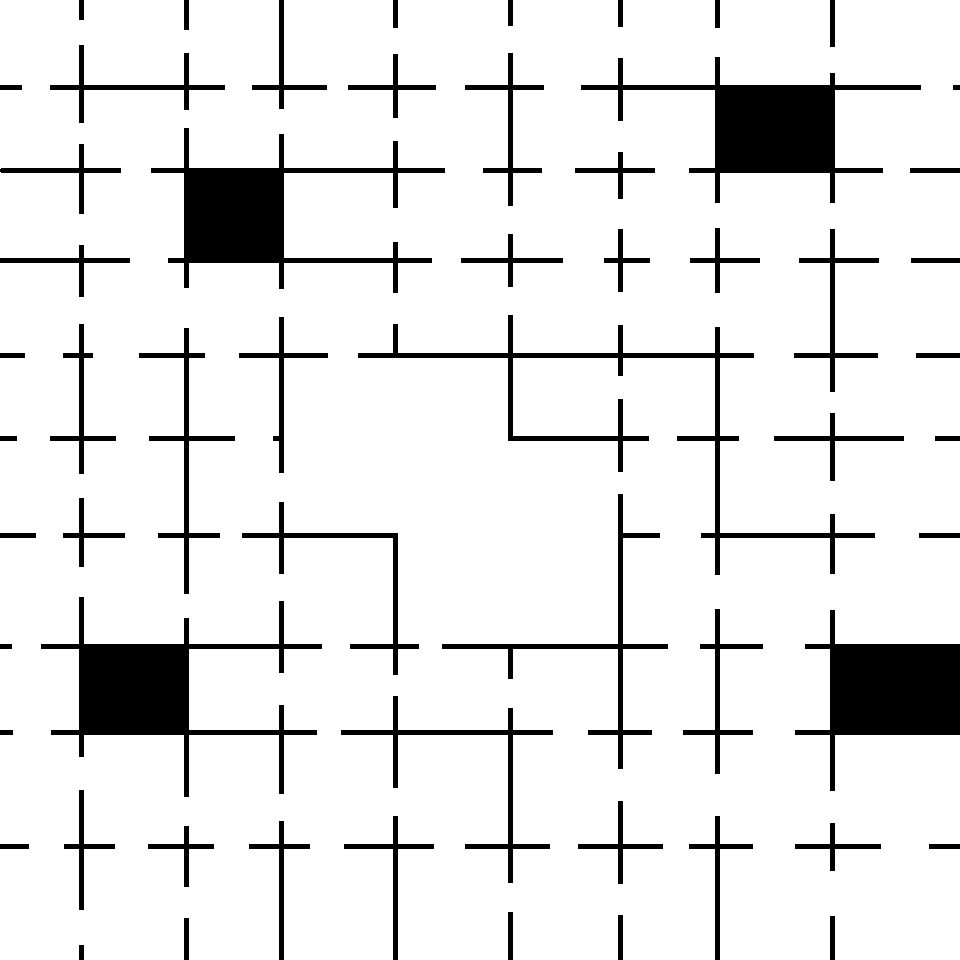}}
    \end{tabular}
    \caption{The fixed training maps are grouped into tiers by difficulty, depending on their size and complexity of the obstacles.}
    \label{supp_fig_map_tiers}
\end{figure}

To improve convergence in the early parts of training, we apply curriculum learning \citep{bengio2009curriculum}, which has shown to be effective in reinforcement learning \citep{narvekar2020curriculum}. We use simple maps and increase their difficulty progressively. To this end, we rank the fixed training maps by difficulty, and assign them into tiers, see Figure \ref{supp_fig_map_tiers}. The maps in the lower tiers have smaller sizes and simpler obstacles. For the higher tiers, the map size is increased together with the complexity of the obstacles.

Furthermore, we define levels containing certain sets of maps and specific goal coverage percentages, see Table \ref{supp_table_levels}. The agent starts at level 1, and progresses through the levels during training. To progress to the next level, the agent has to complete each map in the current level, by reaching the specified goal coverage. Note that randomly generated maps are also used in the higher levels, in which case the agent has to additionally complete a map with random floor plans, and a map with randomly placed circular obstacles to progress. Whenever random maps are used in a level, either a fixed map or a random map is chosen with a $50 \%$ probability each at the start of every episode.

\begin{table}[h]
    \setlength{\tabcolsep}{5pt}
    \begin{center}
        \caption{The progressive training levels contain maps of increasingly higher tiers and goal coverage percentages. At the highest levels, generated maps with random floor plans and obstacles are also used.}
        \vspace{8pt}
        \label{supp_table_levels}
        \begin{tabular}{ccccccc}
        \toprule
            & \multicolumn{3}{c}{Exploration} & \multicolumn{3}{c}{Lawn mowing} \\
            \cmidrule(l{\tabcolsep}r{\tabcolsep}){2-4}
            \cmidrule(l{\tabcolsep}r{\tabcolsep}){5-7}
            & Map & Random & Goal & Map & Random & Goal \\
            Level & tiers & maps & coverage & tiers & maps & coverage \\
            \midrule
            1 & 1,2   & & $90\%$ & 0    & & $90\%$ \\
            2 & 1,2,4 & & $90\%$ & 0,1  & & $90\%$ \\
            3 & 1,2,4 & & $95\%$ & 0,1  & & $95\%$ \\
            4 & 1,2,4 & & $97\%$ & 0--2 & & $95\%$ \\
            5 & 1,2,4 & & $99\%$ & 0--2 & & $97\%$ \\
            6 & 1--4  & & $99\%$ & 0--2 & & $99\%$ \\
            7 & 1--4  & \checkmark & $99\%$ & 0--3 & & $99\%$ \\
            8 & 1--5  & \checkmark & $99\%$ & 0--3 & \checkmark & $99\%$ \\
            \bottomrule
        \end{tabular}
    \end{center}
\end{table}

\subsection{Details on the Compared Methods}
\label{supp_sec_implementation_compared_methods}

\textbf{TSP-based solution with A*.} For the TSP baseline we subdivide the environment into square grid cells of comparable size to the agent, and apply a TSP solver to find the shortest path to visit the center of each cell, which are the nodes. In order to guarantee that each cell is fully covered by a circular agent, the side length for each grid cell is set to $\sqrt{2}r$, where $r$ is the agent radius. Note that this leads to overlap and thus increases the time to reach full coverage. In the offline case, we compute the weight between each pair of nodes using the shortest path algorithm, A* \citep{hart1968formal}. However, due to the size of our environments, the computation time was infeasible. Thus, we implemented a supremum heuristic for distant nodes, where they would be assigned the largest possible path length instead of running A*. This improved the runtime considerably, while not affecting the path length to a noticeable extent. For the online case we applied the TSP solver on all visible nodes, executed the path while observing new nodes, and repeated until all nodes were covered. In this case, the TSP execution time on an Epyc 7742 64-core CPU was included in the coverage time, as the path would need to be replanned online in a realistic setting. However, the execution time was relatively small compared to the time to navigate the path. We also tried replanning in shorter intervals, but the gain in time due to a decreased path length was smaller than the increase in computation time.

\textbf{Code repositories.} Table \ref{supp_table_github_links} lists the code implementations used for the methods under comparison. Note that for omnidirectional exploration, we evaluate our method under the same settings as in Explore-Bench, for which \citet{xu2022explore} report the performance of the compared methods. We report these results in Table \ref{table_explore_bench}.

\begin{table}[h]
    \begin{center}
        \caption{Links to code implementations used for the methods under comparison.}
        \vspace{8pt}
        \label{supp_table_github_links}
        \begin{tabular}{l}
            \toprule
            \textbf{Omnidirectional exploration} \\
            Official Explore-Bench implementation \citep{xu2022explore}: \\
            \scriptsize{\url{https://github.com/efc-robot/Explore-Bench}} \\
            \midrule
            \textbf{Non-omnidirectional exploration} \\
            Official Frontier RL implementation \citep{hu2020tovt}: \\
            \scriptsize{\url{https://github.com/hanlinniu/turtlebot3_ddpg_collision_avoidance}} \\
            Frontier RL implementation in simulation: \\
            \tiny{\url{https://github.com/Peace1997/Voronoi_Based_Multi_Robot_Collaborate_Exploration_Unknow_Enviroment}} \\
            \midrule
            \textbf{Lawn mowing} \\
            Implementation of BSA \citep{gonzalez2005icra}: \\
            \scriptsize{\url{https://github.com/nobleo/full_coverage_path_planner}} \\
            \bottomrule
        \end{tabular}
    \end{center}
\end{table}

\subsection{Evaluation Maps}
\label{supp_sec_evaluation_maps}

\begin{figure}[h]
    \setlength{\tabcolsep}{0.5pt}
    \setlength{\fboxsep}{0pt}%
    \setlength{\fboxrule}{0.5pt}%
    \begin{tabular}{p{1cm}cccccccc}
        \makecell{(a) \\[2pt] \scriptsize \textcolor{white}{line}} &
        \fbox{\includegraphics[width=.14\linewidth]{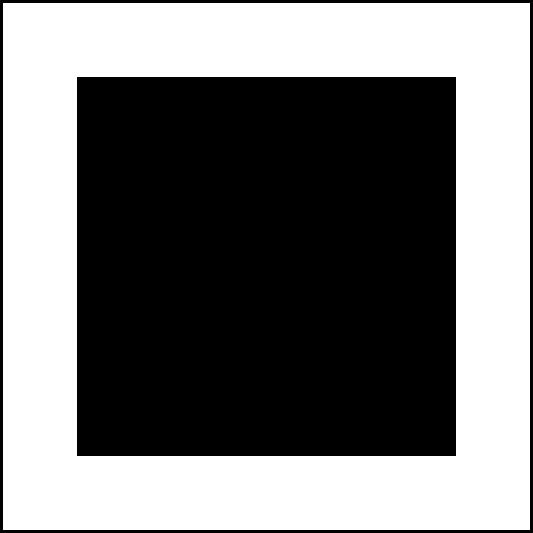}} &
        \fbox{\includegraphics[width=.14\linewidth]{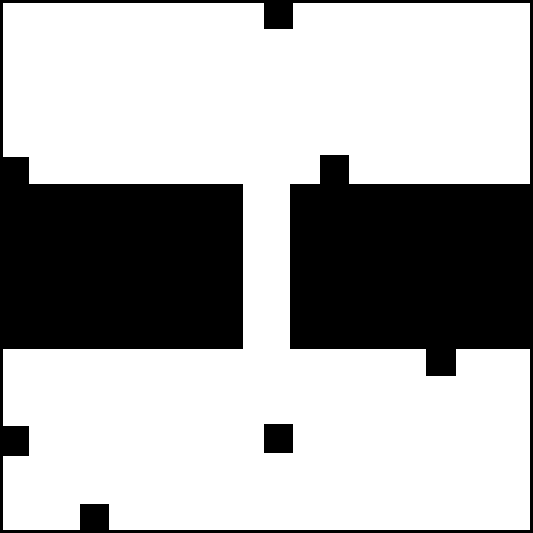}} &
        \fbox{\includegraphics[width=.14\linewidth]{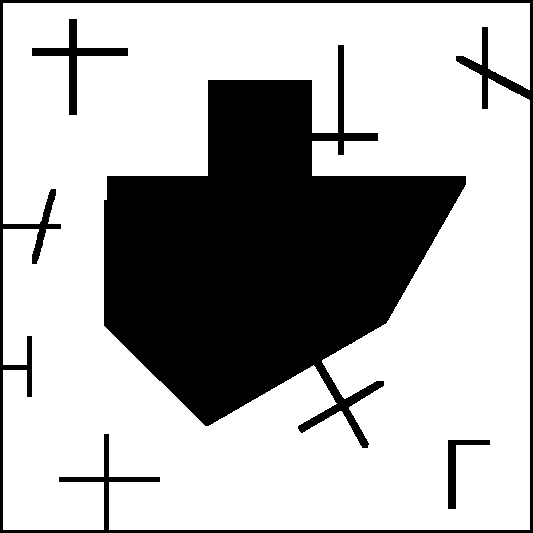}} &
        \fbox{\includegraphics[width=.14\linewidth]{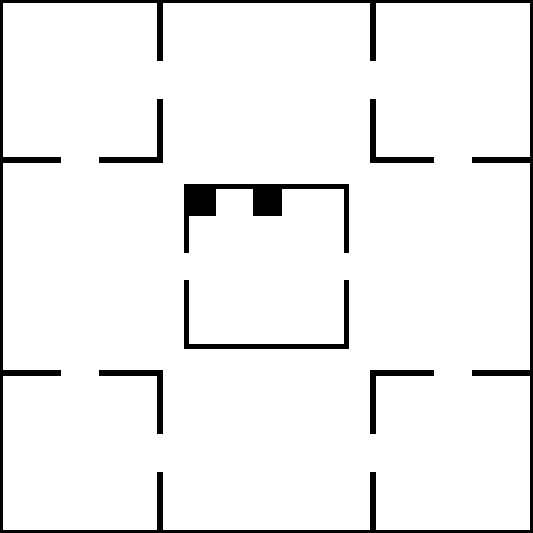}} &
        \fbox{\includegraphics[width=.14\linewidth]{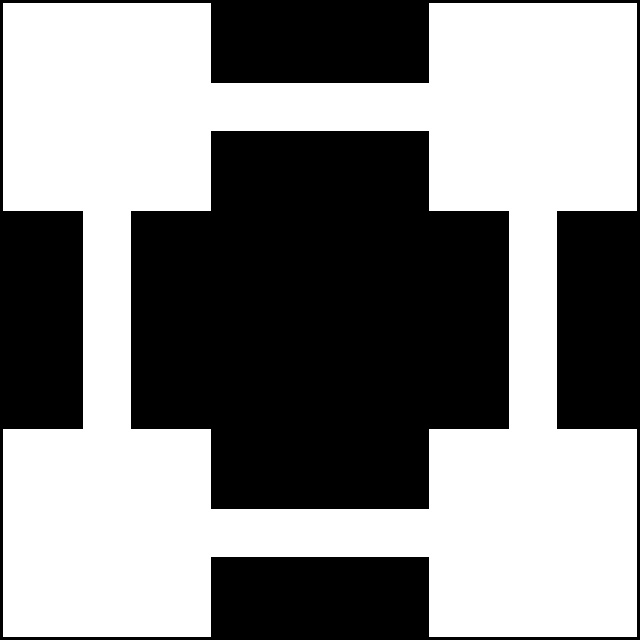}} &
        \fbox{\includegraphics[width=.14\linewidth]{figures/maps/eval_exploration_21.png}}
    \end{tabular} \\
    \begin{tabular}{p{1cm}cccccccc}
        \makecell{(b) \\[2pt] \scriptsize \textcolor{white}{line}} &
        \fbox{\includegraphics[width=.14\linewidth]{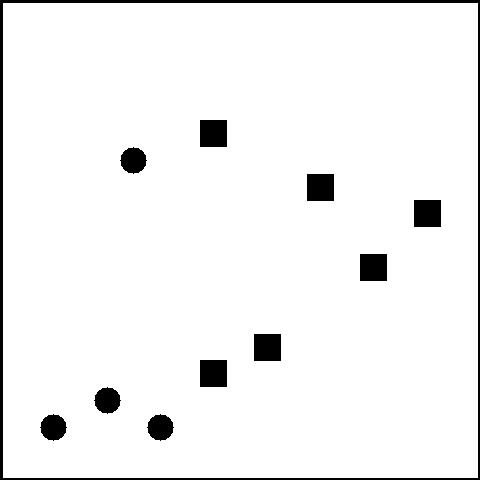}} &
        \fbox{\includegraphics[width=.14\linewidth]{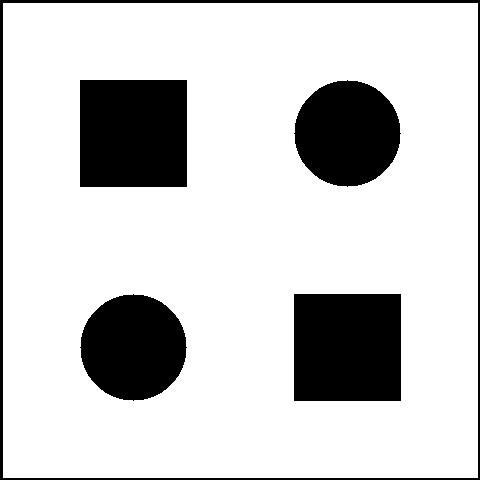}} &
        \fbox{\includegraphics[width=.14\linewidth]{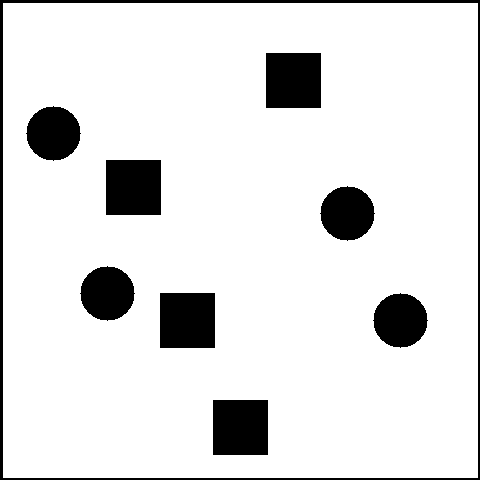}} &
        \fbox{\includegraphics[width=.14\linewidth]{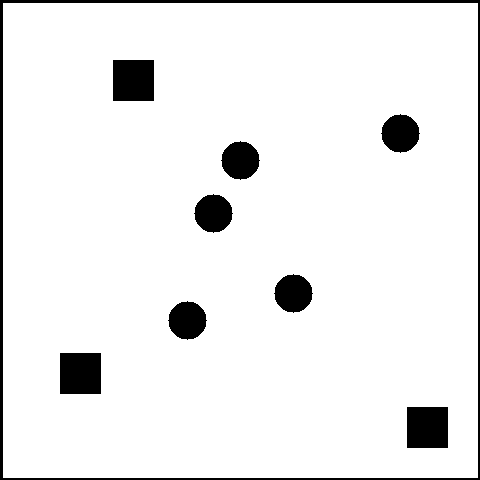}} &
        \fbox{\includegraphics[width=.14\linewidth]{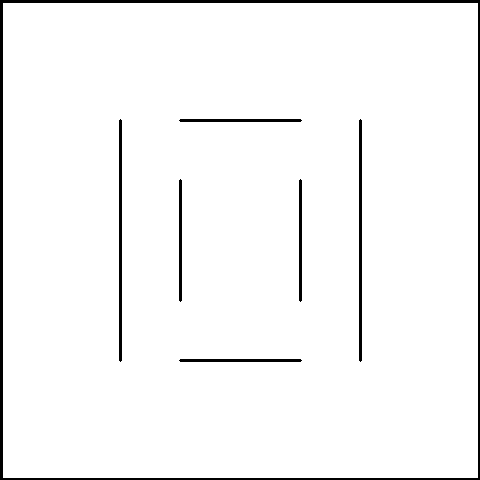}} &
        \fbox{\includegraphics[width=.14\linewidth]{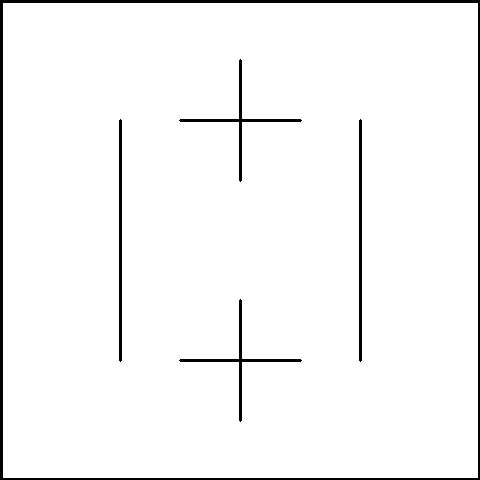}}
    \end{tabular} \\
    \begin{tabular}{p{1cm}cccccccc}
        \makecell{(c) \\[2pt] \scriptsize \textcolor{white}{line}} &
        \fbox{\includegraphics[width=.14\linewidth]{figures/maps/eval_mowing_9.png}} &
        \fbox{\includegraphics[width=.14\linewidth]{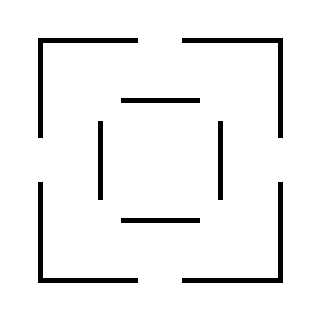}} &
        \fbox{\includegraphics[width=.14\linewidth]{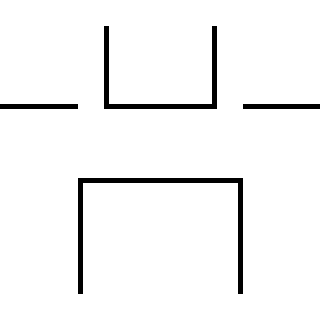}} &
        \fbox{\includegraphics[width=.14\linewidth]{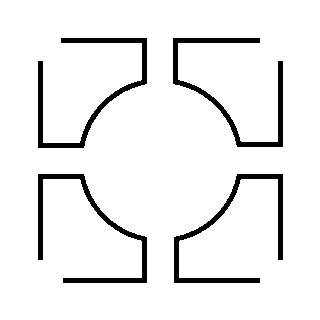}} &
        \fbox{\includegraphics[width=.14\linewidth]{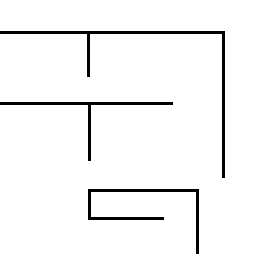}} &
        \fbox{\includegraphics[width=.14\linewidth]{figures/maps/eval_mowing_14.png}}
    \end{tabular}
    \begin{tabular}{p{1cm}cccccccccc}
        \makecell{(d) \\[2pt] \scriptsize \textcolor{white}{line}} &
        \fbox{\includegraphics[width=.085\linewidth]{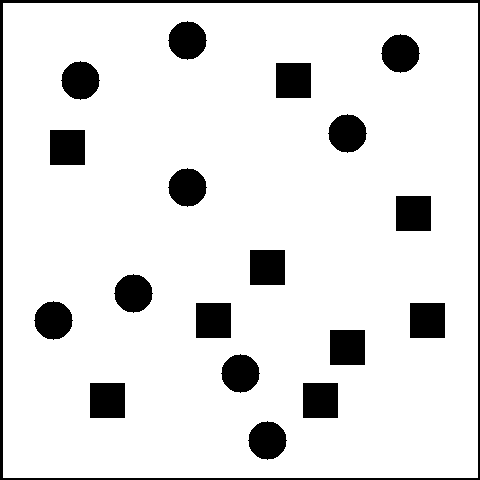}} &
        \fbox{\includegraphics[width=.085\linewidth]{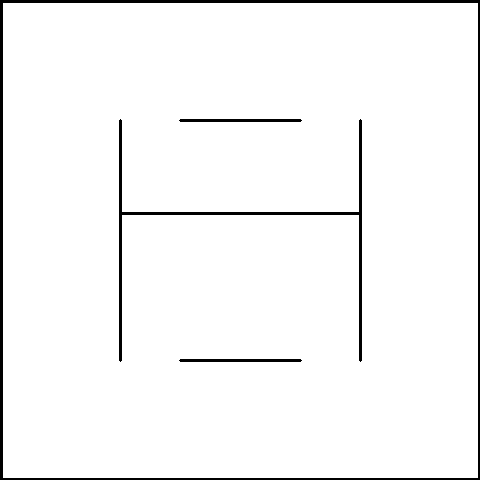}} &
        \fbox{\includegraphics[width=.085\linewidth]{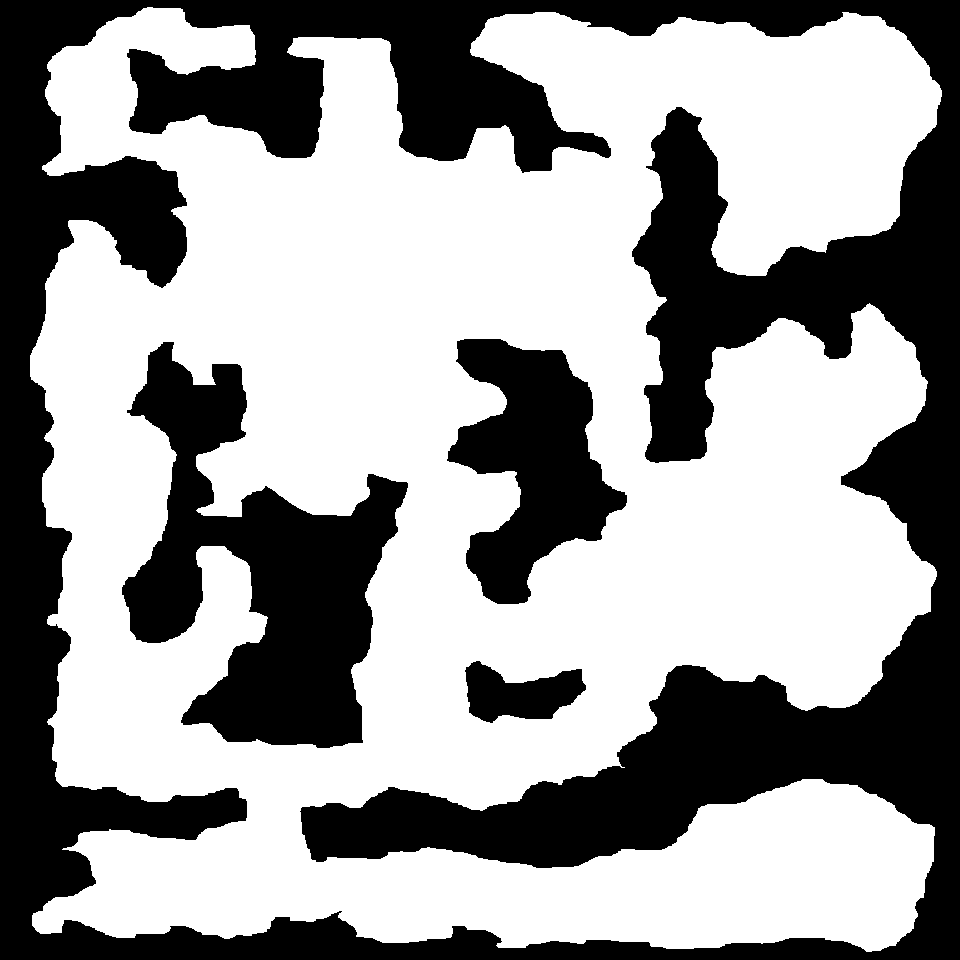}} &
        \fbox{\includegraphics[width=.085\linewidth]{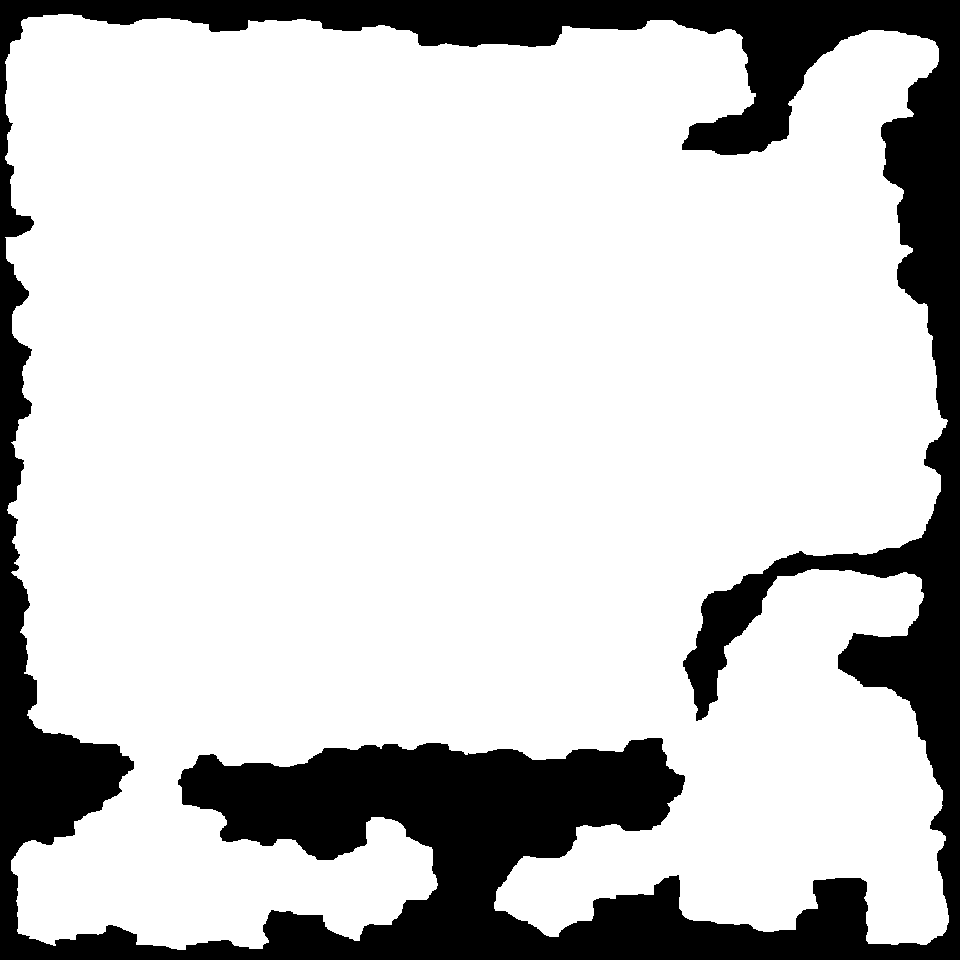}} &
        \fbox{\includegraphics[width=.085\linewidth]{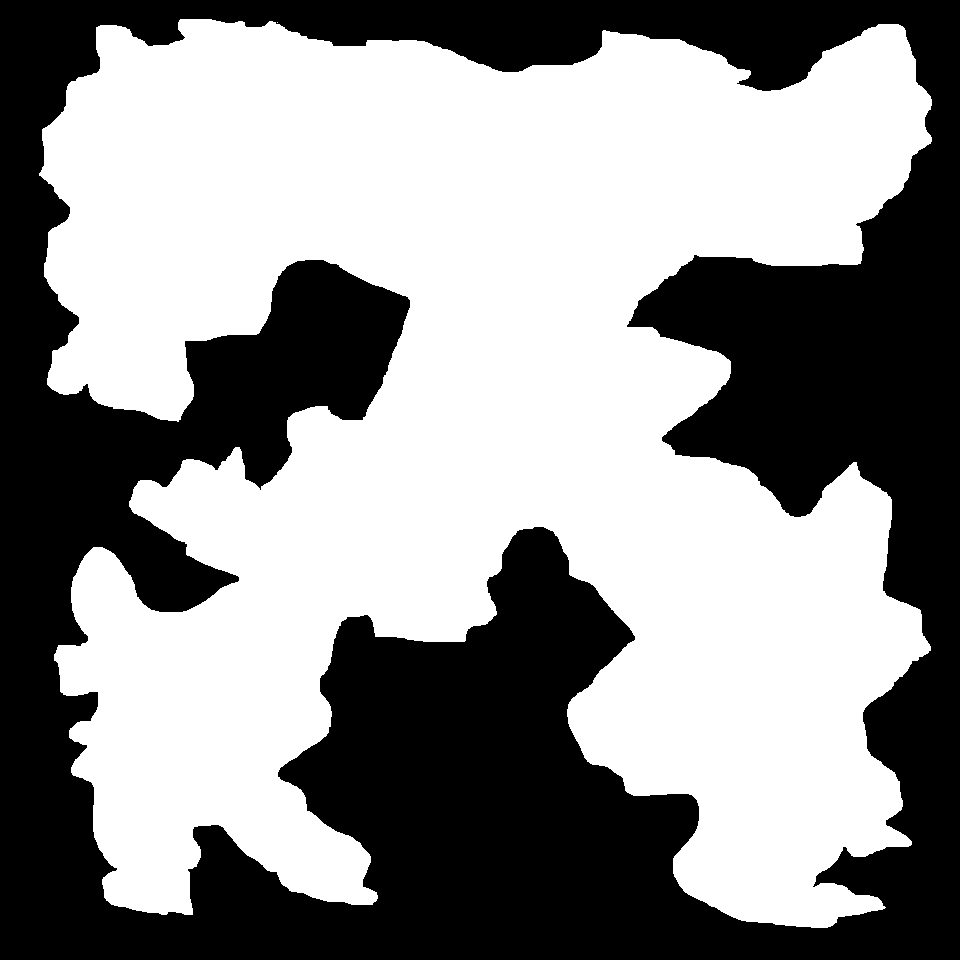}} &
        \fbox{\includegraphics[width=.085\linewidth]{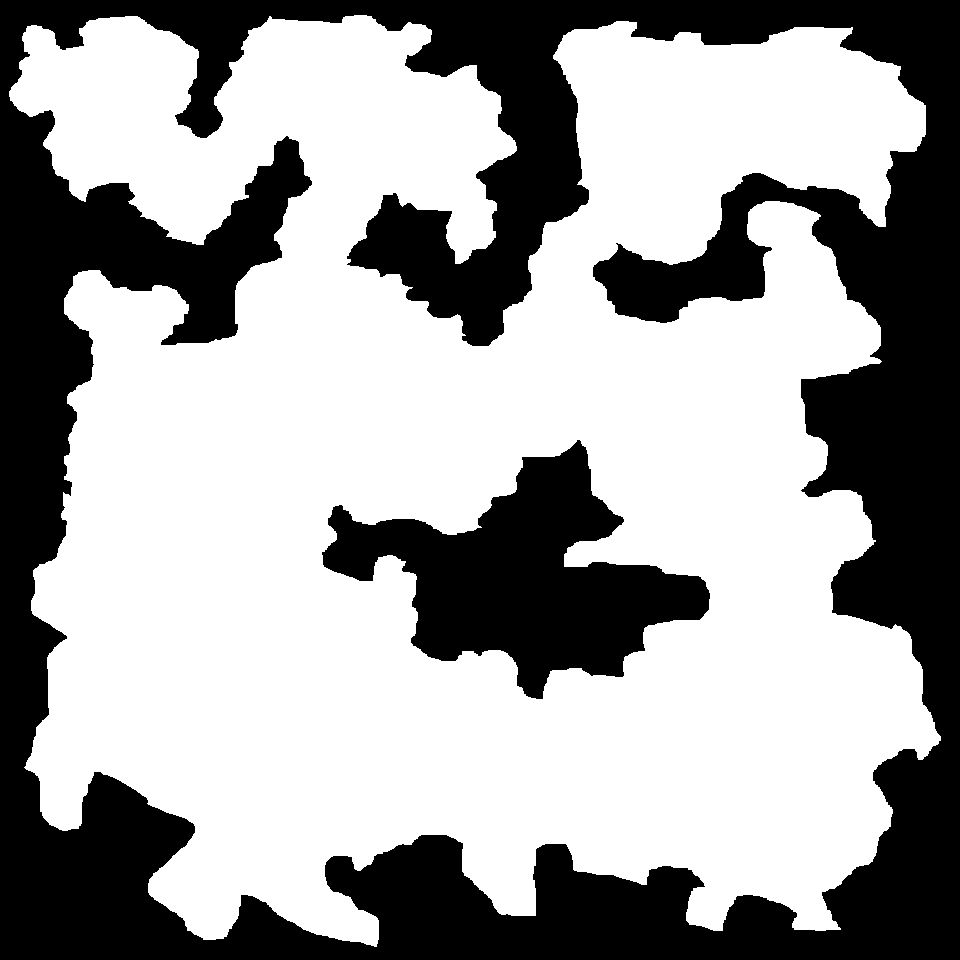}} &
        \fbox{\includegraphics[width=.085\linewidth]{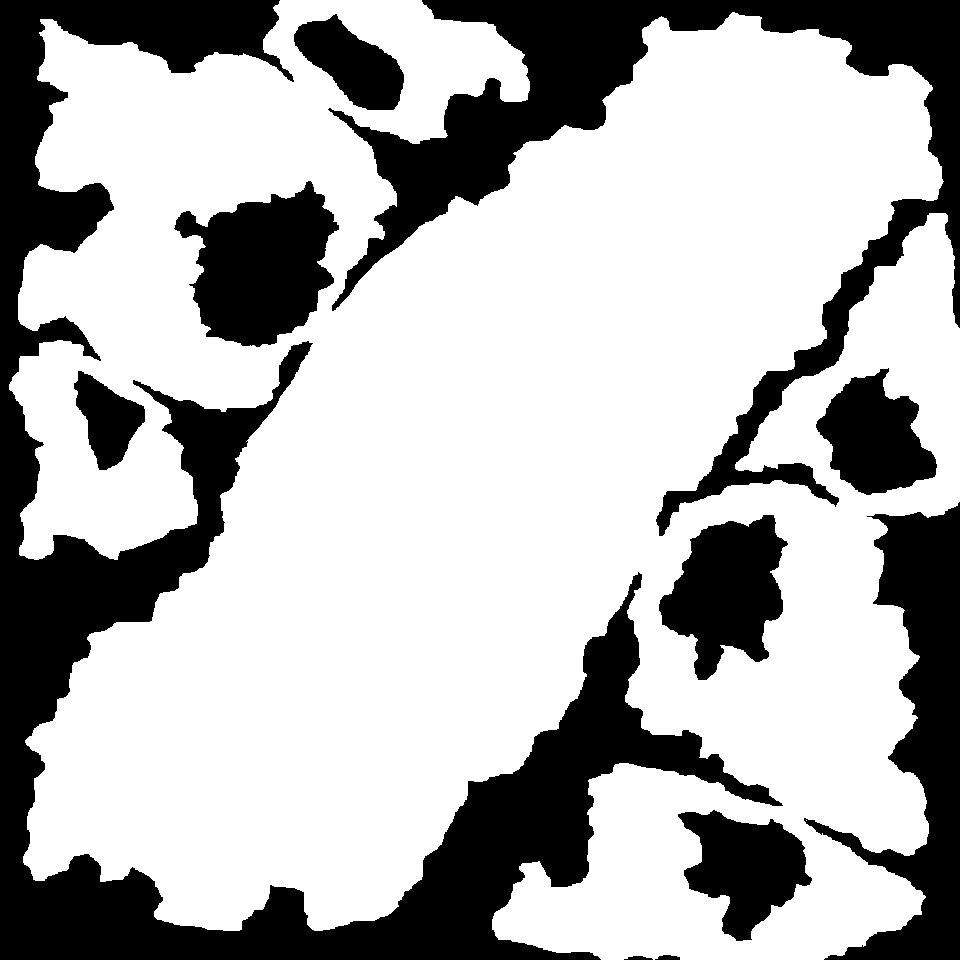}} &
        \fbox{\includegraphics[width=.085\linewidth]{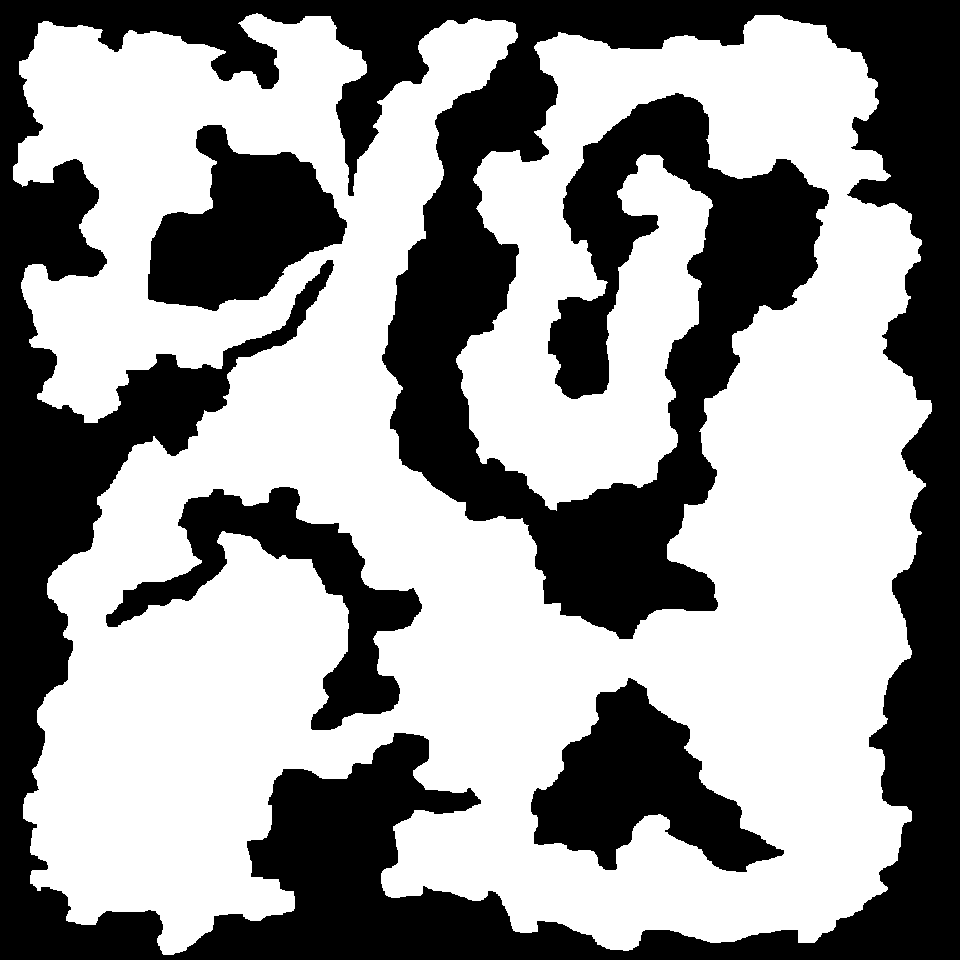}} &
        \fbox{\includegraphics[width=.085\linewidth]{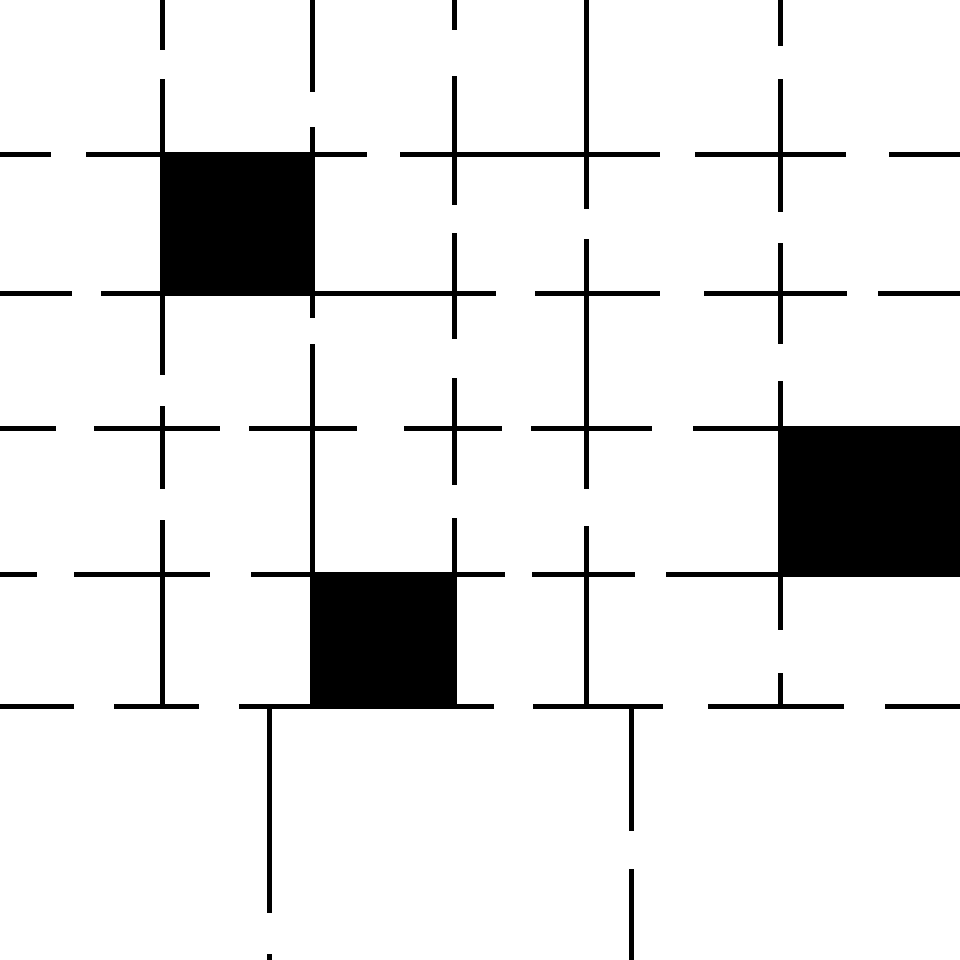}} &
        \fbox{\includegraphics[width=.085\linewidth]{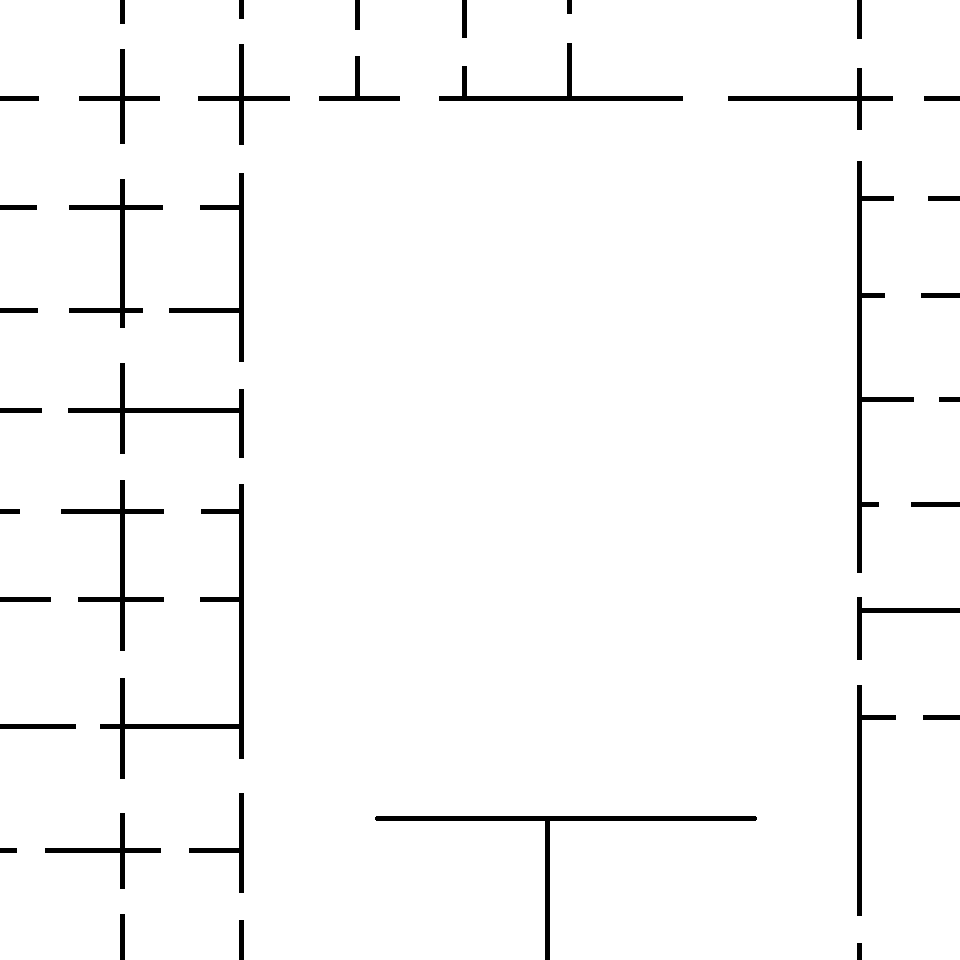}}
    \end{tabular} \\
    \begin{tabular}{p{1cm}cccccccc}
        \makecell{(e) \\[2pt] \scriptsize \textcolor{white}{line}} &
        \fbox{\includegraphics[width=.1\linewidth]{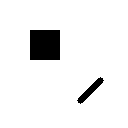}} &
        \fbox{\includegraphics[width=.1\linewidth]{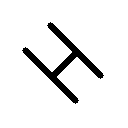}} &
        \fbox{\includegraphics[width=.1\linewidth]{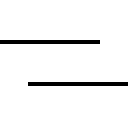}} &
        \fbox{\includegraphics[width=.1\linewidth]{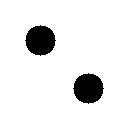}} &
        \fbox{\includegraphics[width=.1\linewidth]{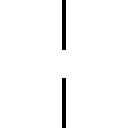}} &
        \fbox{\includegraphics[width=.1\linewidth]{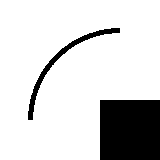}} &
        \fbox{\includegraphics[width=.1\linewidth]{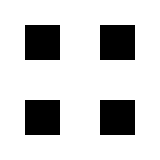}}
    \end{tabular}
    \caption{Evaluation maps used for (a) omnidirectional exploration, (b) non-omnidirectional exploration, and (c) lawn mowing. The maps in the last two rows were used additionally for ablations in (d) exploration and (e) lawn mowing.}
    \label{supp_fig_eval_maps}
\end{figure}

Figure \ref{supp_fig_eval_maps} shows the maps used for evaluating our method on the different CPP variations, as well as additional maps for the ablation study. Figures \ref{supp_fig_eval_maps}(a) and \ref{supp_fig_eval_maps}(c) show the maps in order from left to right as they appear in Tables \ref{table_explore_bench} and \ref{table_mowing} respectively.

\end{document}